\pdfoutput=1
\documentclass[manuscript, acmsmall, nonacm]{acmart}

\AtBeginDocument{%
  \providecommand\BibTeX{{%
    \normalfont B\kern-0.5em{\scshape i\kern-0.25em b}\kern-0.8em\TeX}}}

\usepackage[linesnumbered, ruled]{algorithm2e}

\usepackage{natbib}
\usepackage{xcolor} 
\usepackage{multirow}
\usepackage{wrapfig}
\usepackage{lipsum}
\usepackage{booktabs}
\usepackage{pifont}% http://ctan.org/pkg/pifont
\newcommand{\cmark}{\ding{51}}%
\newcommand{\xmark}{\ding{55}}%
\newcommand{\blue}[1]{{\color{black} #1}}
\newcommand{\teal}[1]{{\color{black} #1}}
\newcommand{\purple}[1]{{\color{black} #1}}
\newcommand{\rev}[1]{{\color{black} #1}} 
\setcopyright{acmcopyright}
\copyrightyear{2021}
\acmYear{2021}
\acmDOI{nnnn/nnnn.nnnn}

\acmJournal{JACM}
\acmVolume{37}
\acmNumber{4}
\acmArticle{111}
\acmMonth{8}

\begin{document}

% \title{Cadence: A Practical Time-series Partitioning Algorithm for Unlabeled Sensor Streams}
\title{Cadence: A Practical Time-series Partitioning Algorithm for Unlabeled IoT Sensor Streams}

%%\authornote{Both authors contributed equally to this research.}
%%\email{trovato@corporation.com}
%%\orcid{1234-5678-9012}
%%\author{G.K.M. Tobin}
%%\authornotemark[1]
%%\email{webmaster@marysville-ohio.com}
%%\affiliation{%
  %%\institution{Institute for Clarity in Documentation}
  %%\streetaddress{P.O. Box 1212}
  %%\city{Dublin}
  %%\state{Ohio}
  %%\country{USA}
  %%\postcode{43017-6221}
%%}

%
% The "author" command and its associated commands are used to define the authors and their affiliations.
% Of note is the shared affiliation of the first two authors, and the "authornote" and "authornotemark" commands
% used to denote shared contribution to the research.

%author information

% author 1
\author{Tahiya Chowdhury}
\email{tahiya.chowdhury@rutgers.edu}
 \affiliation
 {
 \institution{Rutgers University, New Brunswick, New Jersey}
 %\country{USA}
 }

% %author 2
\author{Murtadha Aldeer} 
\email{maldeer@winlab.rutgers.edu}
\affiliation{
  \institution{Rutgers University, New Brunswick, New Jersey}
  %\country{USA}
}

% %author 3
 \author{Shantanu Laghate} 
 \email{shantanu.laghate@rutgers.edu}
\affiliation{
  \institution{Rutgers University, New Brunswick, New Jersey}
  %\country{USA}
}

% %author 4
\author{Jorge Ortiz}
\email{jorge.ortiz@rutgers.edu}
\affiliation{
\institution{Rutgers University, New Brunswick, New Jersey}
%\country{USA}
 }
% 

%
% By default, the full list of authors will be used in the page headers. Often, this list is too long, and will overlap
% other information printed in the page headers. This command allows the author to define a more concise list
% of authors' names for this purpose.


%\begin{abstract}
  
%\end{abstract}
\begin{abstract}

Timeseries partitioning is an essential step in most machine-learning driven, sensor-based IoT applications. This paper introduces a sample-efficient, robust, time-series segmentation model and algorithm. We show that by learning a representation specifically with the segmentation objective based on maximum mean discrepancy (MMD), our algorithm can robustly detect time-series events across different applications. Our loss function allows us to infer whether consecutive sequences of samples are drawn from the same distribution (null hypothesis) and determines the change-point between pairs that reject the null hypothesis (i.e., come from different distributions). We demonstrate its applicability in a real-world IoT deployment for ambient-sensing based activity recognition. Moreover, while many works on change-point detection exist in the literature, our model is significantly simpler 
and can be fully trained in $9$--$93$ seconds on average with little variation in hyperparameters for data across different applications. We empirically evaluate Cadence on four popular change point detection (CPD) datasets where Cadence matches or outperforms existing CPD techniques.

%and matches or outperforms state-of-the-art methods. %~\cite{LSTNet2017, chang2018kernel}.%
%We can fully train our model in $9$--$93$ seconds on average with little variation in hyperparameters for data across different applications. 

\end{abstract}





\maketitle
% \section{Outline}
% Sensys CFP: [\href{http://sensys.acm.org/2020/cfp/}{Link}]

\section{Introduction}\label{sec:1}

\begin{figure*}[ht]
  \centering
  \includegraphics[width=\linewidth]{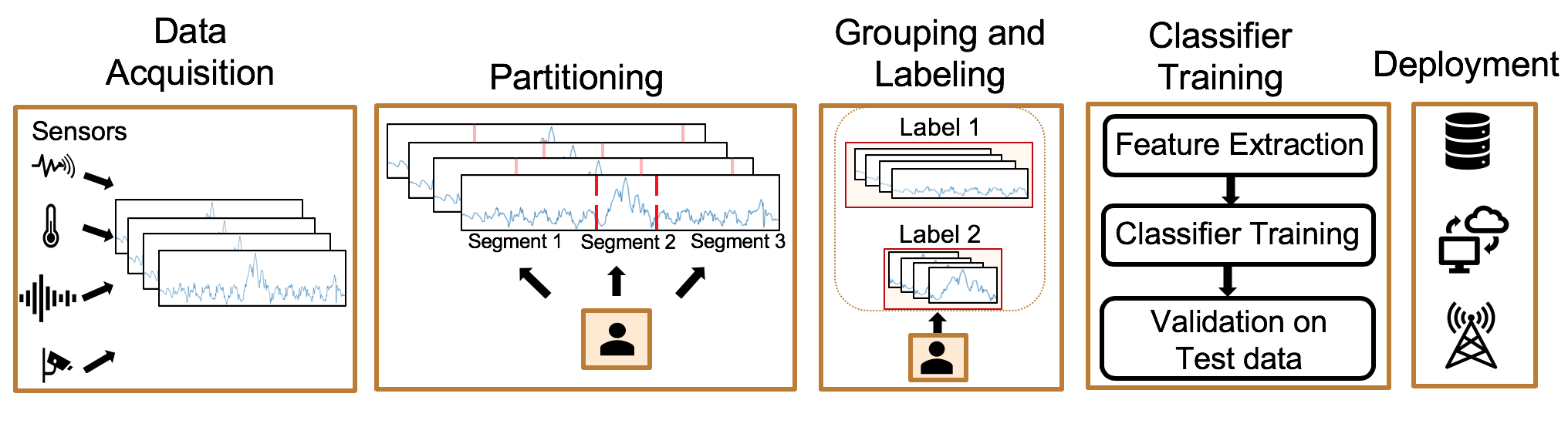}
  \caption{A general machine learning pipeline for sensor-based systems. Note, the partitioning step is crucial and non-trivial, as arbitrary partitioning can lead to arbitrary results~\cite{keogh2005clustering}.  %Partitioning must solve the sample delineation problem to achieve good performance downstream. 
  Applications such as occupancy detection and human activity recognition use manual annotations for partitioning and labeling (i.e. a change in activity or occupancy), increasing labeling effort, and is often inconsistent. Time-series change-point detection can help delineate segments in streaming data without human input, and %by grouping similar segments for annotation, can 
  decrease labeling effort.}
  \label{fig:senseMLP}
  \vspace{-4mm}
\end{figure*}

The number of Internet-of-Things (IoT) and edge devices has exploded in the last decade~\cite{IoT000,IoT00,AGG04}, providing new opportunities to transform everyday people's lives. Coupled with advances in learning technologies~\cite{ML00, ML01}, these can transform how people interact with their environment. Although research in this area has also increased dramatically~\cite{handsense2019, sensehar2019, Sinh2019, Wang2019}, some fundamental challenges remain. A typical machine learning workflow in sensor-based applications starts with unlabeled data. That data is visualized, featurized, and clustered in search of patterns. Typically, labels are obtained, and subsequent sample-label pairs are used to train a classifier. If the data is streaming, this becomes particularly challenging since it is unclear what constitutes the start and end of a training sample. Sensor-based and IoT applications generally follow a workflow where \emph{temporal sequence partitioning} is a typical pre-processing step \footnote{We observe that most machine-learning related papers in past SenSys, IPSN, and UBICOMP contain a segmentation section, whereby temporal partitioning is performed early in the pipeline.}. Data acquisition is fast and inexpensive, trivially involving turning on a sensing device and recording measurements. However, partitioning time series to reflect changes observed in the physical world is notoriously difficult, particularly for large-scale systems. It can either be done implicitly through label acquisition from an expert or explicitly using change-point or time series segmentation techniques. In this paper, we focus on \textit{time-series partitioning} step of sensor-based machine learning pipelines depicted in Figure~\ref{fig:senseMLP}.

Change-point detection (CPD) tries to identify when the probability distribution of a stochastic process changes~\cite{Aminikhanghahi2017}. While segmentation models a time series using piece-wise functional representations~\cite{keogh2004segmenting}, CPD is more useful for learning. Classifiers share similar goals to CPD algorithms. That is, they identify differences in the underlying distribution of the input. %Classifiers train with supervision (using labels). CPD algorithms are unsupervised (without labels).% 

The CPD literature is vast and spans several decades ~\cite{Lowry1992AME, Basseville93, Takuechi, Gustafsson, Kawahara07, IdeT07, Moskvina03} and has appeared across many disciplines and applications, including finance ~\cite{Banerjee_2020, 7288606}, genetic sequence analysis~\cite{Wang11}, anomaly and fault detection~\cite{Yamanishi00}, and human activity recognition~\cite{ni2018, Cook17}. There are two broad classes of CPD algorithms: parametric and nonparametric. Parametric approaches make strong assumptions about the data and do not generalize well across different application contexts~\cite{Adams2007BayesianOC, Chowdhury2012AHN, Nassar2010AnAB}. Nonparametric methods are data-driven and based on divergence metrics and kernel functions~\cite{Yunus2010, NIPS2015_5684}, each requiring a unique parameter configuration for a given application context. Deep learning models are most promising for general use but challenging to tune, as hyperparameter search is expensive. Proposed models are complex, making them difficult or impossible to run on edge devices themselves~\cite{chang2018kernel,LSTNet2017, zhang2020explainable}. Besides, these models are sensitive to the underlying data distribution and the model parameters and thus do not provide a generalized solution.

Developing a data-driven change point detection method that provides computational and sample efficiency and is robust performing across various data domains is non-trivial and poses several challenges that we address in this paper. \textbf{First}, we propose a new way to learn robust time-series features by explicitly learning the similarity (and dissimilarity) between samples through the optimization objective. We show that representations learned in this way are sample efficient and yield better results for change-point detection than computationally expensive, more complex methods. \textbf{Second}, we propose a neural network architecture that learns features from time-series samples by co-training two networks together so that the latent features learned are embedded with similarity characteristics from the samples. This results in a deep embedding learning in latent space that is favorable to change-point detection without distributional assumptions and data-specific hyper-parameter tuning. \textbf{Third}, we show that our method is robust to different parameter choices and yields state-of-the-art performance while being computationally inexpensive. This makes our method, \emph{Cadence}, a generalized change point detection mechanism for time series data – the most common data type in sensor-based, machine learning applications.

To demonstrate the proposed algorithm's performance, we evaluate its performance on four empirical datasets (popular in CPD literature) and compare using several state-of-the-art CPD methods as a baseline. We further evaluate it in a real-world scenario for an ambient sensing application and show that Cadence can detect changes in sensor observations and events in the environments without human input. We make the following contributions:

% To demonstrate the performance of the proposed algorithm, we evaluate its performance on four empirical datasets (popular in CPD literature) and compare using several state-of-the-art CPD methods as a baseline. We further evaluate it in a real-world scenario for an ambient sensing application and show that \emph{Cadence} can detect changes in sensor observations and events in the environments without human input.
% The contributions of this paper are the following:
\begin{itemize}
    \item We propose a new, unsupervised learning approach called Cadence that learns time series embeddings by learning the similarity (and dissimilarity) between samples by incorporating MMD in the objective function. 
	\item We show that it is sample efficient and much easier to train than state-of-the-art techniques.
	\item %To the best of our knowledge,%
	Cadence %is the first%
	offers a generalized change point detection algorithm based on learned embeddings that %is free from%
	does not require distributional assumptions and yields robust performance for different parameter settings.
	\item We show Cadence's use in a real-world IoT application and present an empirical evaluation of Cadence on four popular CPD datasets where Cadence significantly outperforms existing CPD techniques. 
\end{itemize}

Our algorithm is developed as a Python library and will be released as an open-source tool for use in end-to-end sensing applications.

The rest of the paper is organized as follows. We position the proposed approach to fill the gap in the existing change-point detection literature in Section~\ref{sec:11}. Sections~\ref{sec:2} and \ref{sec:3} discuss the motivation behind change-point detection %and its role%
in the machine learning pipeline and the design rationale, respectively. In Sections~\ref{sec:4}, ~\ref{sec:5}, and ~\ref{sec:6}, we discuss the problem formulation of change-point and the two critical components of our method; kernel two-sample test and auto-encoder based feature learning, respectively. We demonstrate its practical deployability %through our experiment%
in a real-world application for ambient-sensing in Section~\ref{sec:7}.  In Section~\ref{sec:8}, we describe the experimental evaluation procedure of the proposed approach. %and the algorithms and datasets used in the evaluation.%
In Section~\ref{sec:9}, we show that Cadence outperforms state-of-the-art techniques and we conclude by discussing %the properties and%
limitations of Cadence and future works in Section~\ref{sec:10}.

\section {Related Work}\label{sec:11}

In this section, we position our work in the context of previous works in the literature on change-point detection and kernel two-sample tests.

\subsection{Change Point Detection}

Many works in change-point detection are based on comparing probability distributions of time-series samples over past and current intervals \cite{Basseville93, Gustafsson}. In these methods, the common strategy is to detect a candidate change point when the two samples become significantly different by some statistical difference measure and the system issues an alarm for a change point \cite{Basseville93, 1101146}. Gustaffsson et al. \cite{Gustafsson} introduced marginalized likelihood ratio-based difference measure to eliminate the pre-defined threshold and reduced computational complexity for generality and robustness. Similar to them, our work refrains from setting a dataset-specific threshold.

Kawahara et al. utilized \cite{Kawahara07} subspace models, where a subspace of the data space is discovered and the distance between the sub-spaces is utilized as a difference measure to identify changes. While the methods in this line are computationally efficient, they require a pre-designed time-series parametric model to form the sub-space \cite{IdeT07, Moskvina03}. Both state-space models and probability distribution methods rely on assumptions about parameters such as mean, variance, and the spectrum of underlying distributions for tracking changes. In our proposed model, we make no such assumptions about distribution parameters. Instead, we track changes in the underlying distribution of the data from features learned via a deep neural network.

As an alternative to parametric assumptions, non-parametric techniques such as density estimation has been used to identify changes in the literature. However, the performance of such methods degrades significantly for high dimensional data \cite{hardle2004nonparametric}. To ease this problem \cite{sugiyama2012density} proposed direct density ratio estimation method, where the ratio of probability densities are estimated without actual calculation of probability density. Following this line, several techniques \textit{KLIEP} \cite{sugiyama2008direct}, \textit{uLSIF} \cite{kanamori2009least}, \textit{RuLSIF} \cite{liu2013change} have been introduced for change detection under non-parametric setting. Liu et al. \cite{liu2013change} used relative density ratio as a difference measure between samples to detect the change, but its performance is affected by the choice of hyperparameters such as window length and sample size. \cite{aminikhanghahi2017using, aminikhanghahi2018real} used a non-parametric change detection approach using Separation distance as the difference measure, but it requires pre-defined feature calculation to apply the algorithm for change-point detection.

\cite{lee2018time, de2020change} explored change point detection using auto-encoders, but their performance for different datasets is largely influenced by window size and parameter settings. Sinn et al. \cite{sinn2012detecting} used Maximum Mean Discrepancy (MMD) as the difference between the ordinal pattern in distribution (order structure in time series values) before and after a change-point under the assumption that the underlying time series is monotonically increasing or decreasing and for detecting multiple change-points requires iteratively applying the method to different blocks of the time series. The work in this paper is closely related to this line of work. While we use MMD as the score of change point possibility measure, we focused on computing change-point scores using kernel two-sample test on features learned from an autoencoder-based neural network.

\subsection{Kernel Two Sample Test}

Kernel two-sample test has been useful for statistical tests to identify the difference between two probability distributions. The two-sample tests are performed based on samples drawn from two probability distributions using a test statistic as the difference between two samples. Harchaoui et al. \cite{Harchaoui08} used a test statistic based upon maximum kernel Fisher discriminant ratio as a measure of homogeneity between segments in hypothesis testing. Along this line, Gretton et al. \cite{gretton2012kernel} introduced Maximum Mean Discrepancy (MMD) and non-parametric statistical tests based on MMD as a test statistic that is free from distribution assumptions. 

MMD-based kernel two-sample test has been effective in change point detection literature \cite{zou14, Harchaoui08, XieYchange}. However, the performance of these kernel methods requires tuning of the bandwidth parameter to obtain optimal performance from kernel measures. In \cite{gretton2012kernel, gretton2012optimal} Gretton et al. presented a bandwidth parameter selection strategy for kernel using median value, which is a simple heuristic for good performance without theoretical validation, as shown by Ramdas et al. \cite{ramdas2015decreasing}. In \cite{gretton2012optimal}, Gretton et al. showed that better kernels can be learned by optimizing test power, which can be less effective in CPD problems with insufficient samples. Chang et al. \cite{chang2018kernel} approached this problem by generating additional samples to mitigate the problem of insufficient samples and optimize test power towards better kernel learning. However, training generative models are computationally expensive and are thus of limited use in practice.

%The contribution of our work is a change point detection method that is free from distributional assumptions and is effective without being computationally expensive.%
We take inspiration from \cite{liu2013change} and compare subsequent samples based on a test statistic to identify a change in the distributions. We use Maximum Mean Discrepancy (MMD) as a test statistic as in \cite{chang2018kernel}. However, instead of generating additional samples towards kernel learning, we learn latent representations for change point detection from the samples by a deep neural network-based optimization. Our approach provides a generalized learning paradigm with simpler network architecture and faster training while achieving better detection performance.

\section{Motivation}\label{sec:2}
% A typical machine learning pipeline has multiple phases: 1) data acquisition, 2) data labeling, 3) classifier training and testing, and 4) deployment and execution. We argue that in sensor-based, streaming machine learning applications, the transition from step 1 to step 2 is particularly challenging.  %Raw, continuous readings do not present clear boundaries between measurement sequences that constitute a sample to capture high-level event information.  Delineation boundaries define discrete samples that are used to train a classifier or to perform further feature engineering before training.  
% %Several challenges emerge in delineating the start and end of a sample over the set of streaming measurements.  
% \textbf{First}, most sensor data are time series in nature, so the labeler can approximately delineate boundaries between events only if the data is put into the proper semantic context.  \textbf{Second}, raw sensor observations do not typically provide enough context, so labels are generated either by labeler \textit{in-situ} or recorded via a contextually richer sensing modality (i.e. video/audio) and labeled a posteriori through manual annotations.
% % This added \emph{context channel} is used to generate boundaries between samples and labels a posteriori.  
% \textbf{Third}, a common solution observed in the literature is to pick a fixed, sliding window over the raw streams and to acquire labels for them. A sensor-based machine learning pipeline is differentiated from a traditional one by these challenges and is illustrated in Figure~\ref{fig:senseMLP}.

A typical machine learning pipeline has multiple phases: 1) data acquisition, 2) data labeling, 3) classifier training and testing, and 4) deployment and execution. We argue that in sensor-based, streaming machine learning applications, the transition from step 1 to step 2 is particularly challenging:
\begin{enumerate}
    \item Most sensor data are time-series in nature, so the labeler can approximately delineate boundaries between events only if the data is put in the proper semantic context.
    \item Raw sensor observations do not typically provide enough context, so labels are generated either by labeler in-situ or recorded via a contextually richer sensing modality (i.e., video/audio) and labeled a posteriori through manual annotations.
    \item A standard solution is to pick a fixed, sliding window over the raw streams and acquire labels. 
    \item The exact delineation boundary for streaming data is not well defined (i.e., two different labelers may delineate the data in similar but unequal ways).
\end{enumerate}

These challenges differentiate a standard machine learning pipeline from a streaming sensor-based one, illustrated in Figure~\ref{fig:senseMLP}.

\begin{figure}[h]
  \centering
  \includegraphics[width=0.55\textwidth]{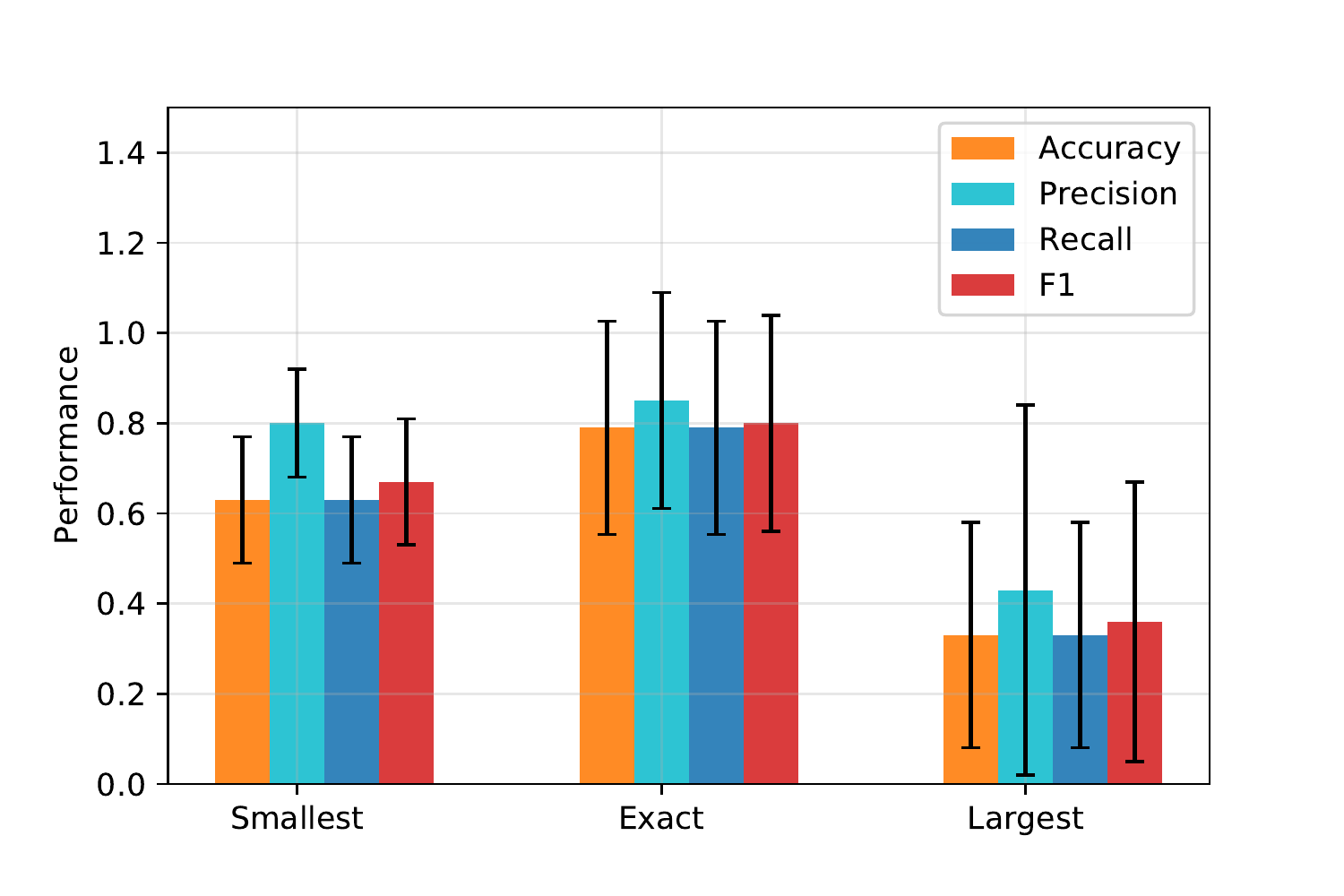}
  \caption{Classifier performance on the Beedance dataset under different delineation boundary sizes. Here, we create samples from the time series by sliding a fixed size window, where under Smallest and Largest condition window size is selected from the smallest and largest true segment present in the data. %Exact, smallest, and largest refer to the actual, smallest, and largest segment sizes in sample discretization, respectively.% 
  Under Exact condition, all samples are created from true segment boundaries in data based on labels. Selecting an arbitrarily small or large window introduces noise into individual samples and can influence the classifier's performance.}
  \label{fig:boundaries}
  %\vspace{-7mm}
\end{figure}

We make several observations of the pipeline presented in the figure. \textbf{First}, as the exact delineation boundary for streaming data is arbitrary, %not well defined (i.e. two different labelers may delineate the data in similar but not exactly equal ways).%
human inspection involves high effort and high cost. Humans are error-prone and can misidentify event transitions, label them inconsistently, or may miss them altogether. \textbf{Second}, the context modality must be synchronized with the sensing modality for proper labeling. A delay in the measurement platform (i.e., network, processing) can cause a delay between the physical observation of an event and when it is recorded.   %and timestamped%.
%Additionally, context-aware manual labeling in real-time by human subjects can interrupt the flow of ongoing activities and cause misalignment between events and labels.%
\textbf{Third}, poor delineation can lead to poor classifier performance and can cause information leakage between events. The chosen window size influences the classifier's performance as selecting an arbitrarily small or large window introduces noise into individual samples. A window too small may not contain sufficient relevant information about an event; a window too large may contain information about multiple events instead of a single event—this can result in biasing the classifier's view of the input distribution.

We make this observation in Figure~\ref{fig:boundaries}, whereby we measure a classifier's performance as we change the boundaries that define discrete samples. Note, there is a drop in performance for samples created using a sliding window of fixed size. We also observe that performance improves when the actual segment boundaries define samples. \cite{10.1145/3287064, aminikhanghahi2017using} made a similar observation that temporal segmentation as a pre-processing step in conjunction with a prediction or classification model leads to performance improvement in comparison to using a fixed-length sliding window, which is the common practice in many application areas including activity recognition \cite{10.1145/3448074, 10.1145/3397323, 10.1145/3448112, 10.1145/3448083}. We highlight some of these applications explored in recent literature in Table~\ref{tab:cpd_areas} that utilizes time-series partitioning procedure. More perniciously, Keogh et al.~\cite{keogh2005clustering} assert that  \emph{features extracted from arbitrary segments by sliding static windows result in arbitrary results}. This observation highlights the importance of finding good, representative sample boundaries in continuous measurements to capture physical events' dynamics.

In many real-world sensing applications~\cite{pan2019fine}, statistical properties of events change over time and expose the system to concept-drift~\cite{JANARDAN2017804, 1484-charconcept} and reduced performance for machine learning applications, as the training samples deviate from the observed ones over time. There is a need to incorporate change point detection techniques to enable continuous-learning applications. %As new sample boundaries are difficult to identify, CPD provides a statistically coherent view of samples from the new distribution and can detect a change in the distribution.%
However, while complex models~\cite{LSTNet2017, chang2018kernel, zhang2020explainable} may outperform simpler ones, they are difficult to use in real systems due to their computation cost.  Most IoT application developers desire an easy to train neural network, robust to different parameters and high performing across various application data. It relieves them from acquiring expertise in the underlying model to tune it for their specific purpose properly.

\begin{table*}[t]
%\color{red}
\begin{tabular}{c|c|c}
\hline
 \textbf{Category} & \textbf{Paper} & \textbf{Application Area}\\ \hline

\multirow{3}{*}{Activity Recognition} &  \cite{aminikhanghahi2017using} & Activity detection in smart home using sensors\\
                                      & \cite{10.1145/3287071} & Activity recognition using RFID \\ 
                                      & \cite{noor2017adaptive} & Activity recognition from accelerometer data \\ 
                                      &
                             \cite{imutube} & Activity recognition using virtual sensor extracted from video \\ \hline

\multirow{3}{*}{Event Detection} & \cite{guo2018device} & Workout event detection using WiFi \\ 
                                & \cite{10.1145/3131898} & Event detection of indoor objects using WiFi\\ 
                                & \cite{dailyevent} & Daily life event segmentation using wearable sensors \\ 
                                & \cite{2021} & Physiological event detection using WiFi and wearable sensor \\\hline

\multirow{2}{*}{Other} & \cite{10.1145/3328919} & User identification and authentication using RFID\\ 
                    & \cite{sarker2018individualized} & Smartphone usage behaviour analysis \\ 
                    & \cite{modelling1010001} & Industrial plant status monitoring using temperature sensor\\\hline

\end{tabular}
\caption{Some application areas in recent literature that use time series partitioning in the pipeline}
\label{tab:cpd_areas}
\vspace{-8mm}
\end{table*}

\section{Design Rationale}\label{sec:3}

\subsection{Properties of Practical CPD Algorithms}\label{sec:design_properties}

% \textcolor{red}{We begin our search in existing literature for change-point detection techniques that can be used for time series partitioning. We find that despite partitioning is an important step for time-series based machine learning applications, the existing solutions do not offer the properties to make them feasible for real-life applications. Here, we describe the properties we seek in a change point detection method and evaluate existing techniques with regards to these properties.}

Although partitioning is an essential step for time-series-based machine learning applications, the existing solutions do not offer the properties to make them feasible for real-world applications. Here, we describe the properties we need in a changepoint detection method to support real-world applications:

\begin{itemize}
    \item \textbf{Property 1: Ease of use.} CPD techniques should be usable without the need for application domain-specific knowledge or strong assumptions about the underlying data distribution. Most approaches in parametric CPD make strong assumptions about data distribution~\cite{adams2007bayesian, Nassar2010AnAB}, and even non-parametric ones are dependent on parameter choices such as window size and the number of training samples~\cite{liu2013change, sugiyama2012density, XieYchange}.
	\item \textbf{Property 2: Computational efficiency.} A CPD library needs to provide satisfactory results while minimizing computational cost and runtime. While some approaches are not parameter-dependent and can achieve good performance, they are computationally expensive and are thus infeasible for practical purposes~\cite{chang2018kernel, Liu2018ClassifierTS}.
	\item \textbf{Property 3: Performance.} CPD techniques should accurately identify changes in the data distribution. A well-performing algorithm may achieve its precise performance due to data-specific parameterization or computational overhead for complex modeling purposes~ \cite{chang2018kernel}, but have limited general use. On the other hand, relaxing the first two properties may degrade detection performance and become unreliable in practice.
\end{itemize}

%\rev{Although changepoint detection studies date back decades, existing techniques do not exhibit all three properties as we show in Table~\ref{tab:properties}.%
We now discuss the design choices we make in our proposed approach, \emph{Cadence}, to ensure these properties. %In this paper, we show the design of Cadence, a CPD algorithm and associated library with all the properties mentioned above. %  

\iffalse
\begin{table*}[t]
\centering
%\color{red}
\begin{tabular}{c|c|c|c}
\hline
 \textbf{Algorithms} & \textbf{Ease of use} & \textbf{Computational Efficiency} & \textbf{Performance}\\ \hline
RDR-CPD \cite{liu2013change} & \cmark & \xmark & \xmark \\ \hline
Mstats-CPD \cite{XieYchange} & \xmark & \cmark & \xmark\\ \hline
LSTNet \cite{Liu2018ClassifierTS} &\cmark & \xmark & \xmark\\ \hline
TIRE\cite{deryck2021change} &  \cmark &  \cmark & \xmark\\ \hline
KLCPD \cite{chang2018kernel} & \xmark & \xmark & \cmark\\ \hline
Cadence (Proposed Approach) & \cmark & \cmark & \cmark\\ \hline

\end{tabular}
\caption{Overview of properties exhibited by existing CPD algorithms. We design our proposed approach, \textit{Cadence}, to exhibit all three properties.}
\label{tab:properties}
\vspace{-4mm}
\end{table*}
\fi

\subsection{Design Discussion}

% \textcolor{red}{Our change point detection scheme is to compute a dissimilarity score between two consecutive segments in time series such that this score can act as a probability of change point. There are several design choices involved in this process to ensure that the mechanism exhibits the properties in section~\ref{sec:design_properties}.}
% \\

Our change point detection scheme computes a dissimilarity score between two consecutive segments in time series such that this score can act as a probability of change point. There are several design choices involved in this process to ensure that the mechanism exhibits the properties in Section~\ref{sec:design_properties}.

% \textbf{Maximum Mean Discrepancy.} We utilize Maximum Mean Discrepancy (MMD) to measure the dissimilarity between the segments. Maximum Mean Discrepancy (MMD) is a probabilistic distance measure that has been useful in non-parametric hypothesis testing \cite{gretton2012kernel}. It is attractive for our purpose for several reasons. First, MMD is a non-parametric distance measure, which relieves from parameter choice for estimation (Property 1). Second, MMD represents the distance between probability distributions as distance between mean embeddings of features and can be easily empirically computed with a given number of samples (Property 2). Based on these advantages, along with its symmetric and non-negative property, we choose MMD over other distance measures such as Kullback-Leibler divergence \cite{perez2008kullback} as our dissimilarity metric.

\rev{\textbf{Maximum Mean Discrepancy.} We utilize Maximum Mean Discrepancy (MMD) to measure the dissimilarity between the segments. MMD is a probabilistic distance measure that has been useful in non-parametric hypothesis testing~\cite{gretton2012kernel}. In our proposed technique, we perform hypothesis testing to determine change points in a two-sample test setting using Maximum Mean Discrepancy (MMD) as the test statistic. We perform the test between two consecutive segments (null hypothesis of they come from the same distribution vs. alternate hypothesis of they come from different distributions). While, there are other measures for the distance between two distributions such as Kullback-Leibler (KL) divergence~\cite{perez2008kullback}, %we use MMD as our test statistic due to its non-negative, symmetric and especially non-parametric properties.% 
MMD is attractive for our use case for several reasons. First, it is a non-parametric distance measure, which relieves from parameter choice for estimation (\textbf{Property 1}). Second, MMD represents the distance between probability distributions as the distance between mean embeddings of features and can be easily empirically computed with a given number of samples (\textbf{Property 2}). As we are interested in MMD between latent embeddings of two segments to be used as change-point score, the non-negative, symmetric, and non-parametric properties of MMD provide a solution that is easy to compute and high-performing. Based on these advantages, we choose MMD over other distance measures such as KL divergence as our dissimilarity metric.}

% \textbf{Autoencoder.} We choose a stacked auto-encoder to learn the feature representations from time-series samples in a lower dimension space and calculate MMD in that space instead of calculating in data space. MMD-based change point detection is grounded on measuring the distance by applying a function on the samples to learn feature embeddings. \cite{chang2018kernel} found that learning a better representation can help in the hypothesis testing-based change point detection in comparison to performing the test in the original data space. However, their approach based on generating additional samples is computationally expensive. 

\textbf{Autoencoder.} We choose a stacked auto-encoder to learn the feature representations from time-series samples in a lower dimension space and calculate MMD in that space instead of calculating in data space. MMD-based change point detection measures the distance by applying a function on the samples to learn feature embeddings. Chang et al.~\cite{chang2018kernel} found that learning a better representation can help hypothesis testing-based change point detection compared to performing the test in the original data space. However, their approach based on generating additional samples is computationally expensive (\textbf{Property 3}).

On the other hand, with Autoencoder, we use gradient descent via back-propagation to learn the representation that is suitable for detecting changes between the samples, instead of calculating MMD between two samples in a linearly embedded space as in principal component analysis (PCA) or non-negative matrix factorization (NMF). Recent research has shown that auto-encoders are capable of producing meaningful and well-distinguished representations on real-world datasets~\cite{vincent2010stacked, le2013building, hinton2006reducing} and exhibit denoising property in feature learning [72], which can be useful for noisy real-world time-series from IoT (\textbf{Property 2}). %LeCun et al.~\cite{lecun2015deep} show that deep learning algorithms are, in fact, representation learning frameworks with representations lying at multiple layers; at each layer of the multi-layer non-linear transformation of the input. Bengio et al.~\cite{bengio2013representation, bengio2009learning}  explain that the earlier layers of a deep neural network learn more generic features, where deeper layers learn more relevant and detailed features specific to the task.%
Furthermore, autoencoder training's unsupervised nature facilitates learning representations without explicitly utilizing labels.

We experimented with several other models described in the literature, including a variational autoencoder~\cite{kingma2013auto} and a variational recurrent neural network~\cite{chung2015recurrent}. Surprisingly, despite its simplicity, we found autoencoders outperform more complex models across various datasets.  It is also simpler to train.

\subsection{Caveats and Shortcomings}

% \textcolor{red}{Before describing our proposed method, we explain some additional design choices we make in our method.}
% \\
Before describing our proposed method, we explain some additional design choices we make in our method.

\begin{figure}[h]
  \centering
  \includegraphics[width=0.75\linewidth]{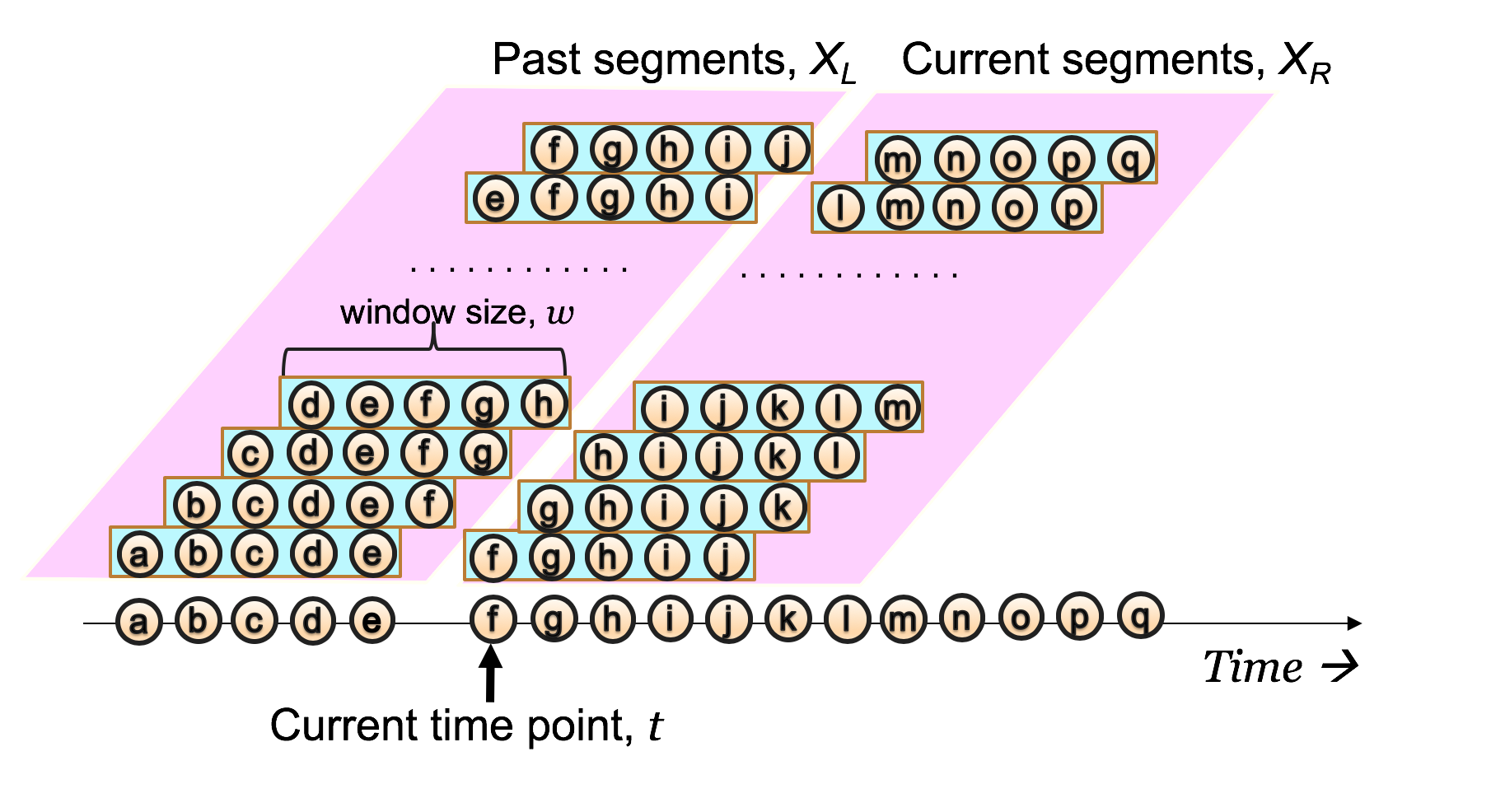}
  \caption{By sliding window $w$, pairs of past segment ${x_L} = \{{y_{t-w}},\dots,{y_{t-1}}\}$ and current segment ${x_R} = \{{y_{t}},\dots,{y_{t+w-1}}\}$ are compared to detect change point at every incoming time point. Note that, the objective of this window is to create non-overlapping segments for comparing their dissimilarity and use it as change-point probability at $t$ in two-sample test.}
  %\vspace{-6pt}
  \label{fig:segments_matrices}
\end{figure}

\textbf{Window.} In our proposed method, we use a sliding window to create segments for testing the hypothesis for change point. A sliding window may raise concern among the readers as the window size is a parameter choice, and this choice can influence algorithm performance. \teal{Note that the window size dictates the length of a segment.  We compare segments by calculating the dissimilarity between consecutive segment pairs (Note Figure~\ref{fig:segments_matrices}). However, our window-size choice does not dictate the start and end of an input sample used downstream in the rest of the processing pipeline. Here, we are interested in finding a homogeneous sample for an ML application, where the sample represents a single event or activity. To perform that comparison, we slide a window across the entire stream to compare consecutive segments created by the process above and use the dissimilarity to identify the start and end of a sample.} 

However, the fixed window size can still influence the algorithm's sensitivity to events.  %We found a heuristic that works well experimentally.  That is, to set the window size based on the changepoints' expected duration and frequency (i.e., its cadence). Data with oscillatory, frequent changes use smaller window sizes; larger window sizes capture higher-level context.%
We choose a window size of 25 timesteps for all of our experiments and experimentally explore how the choice influences performance in Section~\ref{impact_window}.

% \textbf{Threshold.}
% We use Maximum Mean Discrepancy between two consecutive segments as a dissimilarity score such that this score can act as a probability of change point. We report the performance of our proposed method using Area Under Curve Score (AUC) from true positive rate and false positive rate obtained using this change-point score. However, to partition a time series into segments based on the obtained change point score, we need to set a threshold on the score. While finding an adaptive threshold that works well across different datasets is outside of the scope of this paper, in our experiments, we empirically observed that using $0.4$ x maximum MMD score provides good detection performance.

%\vspace{0pt}

\textbf{Threshold.} We use the maximum mean discrepancy (MMD) score between consecutive segment pairs as the changepoint score. We report our proposed method's performance using Area Under Curve Score (AUC) from true-positive rate (TPR) and false-positive rate (FPR) obtained using this changepoint score. However, to partition a time series into segments based on the obtained changepoint score, we need to set a threshold on the score. While finding an adaptive threshold that works well across different datasets is outside of this paper's scope, in our experiments, we empirically observed that using a threshold of 40\% of the maximum change point score provides good detection performance.

\section{Problem Formulation}\label{sec:4}
In this section, we formulate the problem of change point detection for time series. Let us consider a time series of $d$-dimensional observations 
$Y =\{{y_1},{y_2},{y_3},\dots {y_t,\dots}\}$, where ${y_t}\in {\mathbb{R}^d}$. We create segments of window size $w$ and treat each segment as a single sample instead of a single $d$-dimensional observation. We follow the previous works in the literature in this choice \cite{liu2013change, sugiyama2008direct} as this captures the time-dependent information in the observations. Our objective is to detect a change point, where the observations undergo change.

%\begin{figure}[h]
  %\centering
  %\includegraphics[width=\linewidth]{Figures/past_present_segment.jpg}
  %\caption{Dissimilarity between past segment ${x_L}$ and current segment ${x_R}$ as a measure of probability of time point $t$ being a change point}
  %\label{fig:segments_pair}
%\end{figure}

Towards the goal of change point detection, we consider two consecutive segments of size $w$ at time $t$; past segment $x_L$ and current segment $x_R$. Our change point detection scheme is to compute a dissimilarity score between these two consecutive segments such that this score can act as a probability of change point at time $t$. %Figure~\ref{fig:segments_pair}%
 Specifically, we want to leverage the dissimilarity score to answer whether samples ${x_L}$ and ${x_R}$ are from different distributions and hence decide whether the current time $t$ is a change point. Here, $t$ is the first time-point in the current segment marking the boundary between the two segment pairs. We create such segment pairs by a sliding window approach for the entire time series. This results in two matrices of past segments ($X_L$) and current segments ($X_R$). The dissimilarity between each pair allows us to compute change point scores for ${{t},{t+1},{t+2},{t+3},\dots}$ as shown in Figure~\ref{fig:segments_matrices}. 

Figure~\ref{fig:segments_matrices} illustrates the change point detection strategy for one-dimensional time series. For multi-variate time series, we concatenate segments in all dimensions into one segment while maintaining the order of time. This allows our strategy to be valid regardless of the number of dimensions under consideration instead of detecting changes in each dimension individually. In the next section, we introduce the dissimilarity score used for measuring change point probability.

\blue{\textbf{Constraints.} Notice that following the previous literature \cite{liu2013change, sugiyama2008direct, chang2018kernel}, we create samples of window size $w$ to capture time-dependent information in our samples. In this setting, the method needs to have a specific number of observations available to create samples (which is dependent on the size of the window) for computing dissimilarity between sample pairs. As a result, the method cannot detect changes as soon as they occur as considered in the online changepoint detection setting. However, the smaller the window size is, the faster it can respond to a changepoint in inference mode (when a trained model is used to find change points on new data). Furthermore, we made piece-wise \emph{i.i.d.} assumption regarding the segments created in this setting following the previous works \cite{Harchaoui08, Kawahara07, XieYchange}. However, in many settings, samples may not follow such assumptions and can follow non-\emph{i.i.d} autoregressive process. We formulate the problem of change point detection for time series under these constraints and extending beyond this setting can be interesting which we leave for future work.} 

\section{MMD for two-sample test}\label{sec:5}

\subsection{Two-sample Test}

We utilize the concept of Maximum Mean Discrepancy (MMD) to measure the dissimilarity between the segments, ${x_L}$ and ${x_R}$. Maximum Mean Discrepancy is a probabilistic difference measure that is used for the two-sample test. In a two-sample test problem, a statistical test is performed under the null hypothesis that the probability distribution of the two samples is the same against the alternative hypothesis that the probability distribution of the samples is different. 

Let us consider sample \hbox{$X =\{{x_1},{x_2},\dots, {x_m}\} \sim P$} and 
sample $Y =\{{y_1},{y_2},\dots\
, {y_n}\} \sim Q$ are \textit{i.i.d} and drawn from distributions $P$ and $Q$ respectively. The definition of two-sample test can be formally established as:

${H_0}: P = Q$, samples X and Y are from same distributions.

${H_A}: P \neq Q$, samples X and Y are from different distributions.

Ideally, in a two-sample test, the test statistic should be such that when this measure is large, the samples are likely from different distributions, $ P \neq Q$. On the other hand, the statistic becomes 0 if and only if  $P = Q$. The closer the test statistic exhibits this behavior, the less susceptible it is to Type I and Type II errors.

\subsection{MMD as test statistic} 

Maximum Mean Discrepancy (MMD) has been used in the literature as the two-sample test statistic \cite{Harchaoui08}. MMD is defined as the difference between the mean embedding of the two samples. The hypothesis of two distributions $P$ and $Q$ being different is tested based on the samples drawn from each of them.  Following the notations in \cite{gretton2012kernel,gretton2012optimal,  Gretton2009}, MMD between two distributions $P$ and $Q$ can be thus expressed as distance between the mean embeddings of $P$ and $Q$ in the following way,
\begin{equation}
    {MMD}^2({P}, {Q}) =  {{M_k}}({P}, {Q}) = {{\parallel {\mu_P} - {\mu_Q} \parallel}^2}
\end{equation}

Given a kernel $k$ in the embedding space and the samples from $P$ and $Q$ are $X$ and $Y$, \textit{MMD} can be written in terms of  kernel functions as following,

\begin{equation}
    {{M_k}}({X}, {Y}) = \mathop{\mathbb{E}_P}[{k(x, x')}] - 2\mathop{\mathbb{E}_{PQ}}[{k(x, y)}] + \mathop{\mathbb{E}_Q}[{k(y,y')}]
\end{equation}

where $x'$ and $y'$ are an independent copy of $x$ and $y$ from the distributions $P$ and $Q$ respectively. It has been demonstrated in \cite{nishiyama2016} that for both Gaussian and Laplace kernels,  ${M_k}({P}, {Q}) = 0$ if and only if ${P} = {Q}$. Additionally, both kernels exhibit properties of a characteristic kernel where ${M_k}({P}, {Q})$ is non-negative.

The change point detection problem can be simplified into a binary classification problem where MMD at each time step acts as the change point probability score for the corresponding timestamp. ${M_k}({P}, {Q}) > 0$ between $x_L$ and $x_R$ thus denotes the probability of time step $t$ being a change point. Note that, the two-sample test is designed to inform on the hypothesis of two samples coming from the same distribution based on ${M_k}({P}, {Q}) = 0 $. We leverage its notation for the change-point detection setting by using the MMD statistic as a change point probability.

In practice, ${M_k}({P}, {Q})$ is estimated using finite samples from the two distributions $P$ and $Q$ through their kernel mean embedding. 
However, the test power is limited by the finite samples available for the estimation, as change points are rare in most cases. 

%However, this provides limited effectiveness in practice under finite sample size which can prove insufficient depending on the complexity of the data. We later describe our method to learn feature embedding that does not rely on the finite samples from the population for MMD estimation.

\subsection{Kernel Choice for MMD}

The motivation for the choice of kernel can depend on problem complexity and the information to be extracted from the data. MMD utilizes the difference between the embeddings of the samples to perform the two-sample test. Hence, the kernel or similarity metric used in the test is crucial to its performance \cite{gretton2012kernel}. Recall that the two-sample test should ensure that samples from unlike distributions should be distinguished with high probability. This minimizes Type II error, or false negative rate, which occurs if the test fails to show a difference when there is one. Similarly, the test statistic should be 0 (or very close to 0) when samples from the same distributions are compared.
 
The family of radial basis function (RBF) kernels exhibits such behavior. %We focus on two such characteristic kernels - Gaussian $\mathds{k}$($x$, $y$) =  $exp  (\frac{{\Vert x - y \Vert}^2} {\gamma^2})$  and Laplace $\mathds{k}$($x$, $y$) =  $exp  (\frac{{\Vert x - y \Vert}^2} {\gamma})$. 
An RBF kernel is defined as, 
\begin{equation}
    {k(x, y)} = exp  (- \gamma {\Vert x - y \Vert}^2 ) 
\end{equation}

where the value of $\gamma$ is an adjustable parameter that provides sensitivity. The common heuristic to choose the value of gamma for %both Gaussian and Laplace kernels%
RBF kernels in the literature is to use the median value \cite{scholkopf2002learning}. %This heuristic has been used frequently due to lack of theoretical understanding regarding kernel choice.%
However, the median heuristic does not guarantee maximum power and optimal performance of test statistics. Note that, the RBF kernel function is a non-linear euclidean distance whose value depends on the choice of $\gamma$. While the value of $\gamma$ contributes to the non-linear nature of this distance measure and the performance of the kernel, it requires hand-tuning for optimal performance depending on the data complexity and problem domain.

Considering $x = x'$ and $y = y'$ in (2), and combining (2) and (3), we get,
\begin{equation}
    {{M_k}}({X}, {Y}) = \mathop{\mathbb{E}_P}[{k(x, x)}] - 2\mathop{\mathbb{E}_{PQ}}[{k(x, y)}] + \mathop{\mathbb{E}_Q}[{k(y,y)}]
\end{equation}

As a result, the first and third term in (4) becomes 1, which results in the value of $ {{M_k}}({X},{Y})$ being dependent on the 2nd term only. There can be two cases here:

\begin{center}
    \textbf{Case 1}: $x=y: k(x, y) = 1 \rightarrow{} {{M_k}}({X}, {Y}) = 0$

    \textbf{Case 2}: $x \neq y: k(x, y) < 1 \rightarrow{} {{M_k}}({X}, {Y}) > 0$
\end{center}

%Note that this lays the foundation for test statistic suitable for the two-sample test in change-point detection setting. Whereas empirical estimation of $MMD$ relies on kernel embedding obtained via population samples, the test power is limited by the finite samples available for the estimation, as change points are rare in most cases. %particularly for scenarios where the data is highly imbalanced across the classes.%
Note that this lays the foundation for test statistics suitable for the two-sample test in the change-point detection setting. In the next section, we discuss an embedding learning method via a deep neural network that learns features useful towards change point detection.

\section{Deep Embedding Learning}\label{sec:6}

LeCun et al. \cite{lecun2015deep} show that deep learning algorithms are in fact representation learning frameworks with representations lying on multiple levels, which are obtained by multi-layer non-linear transformation of the input. While many works have been done in the computer vision and natural language processing literature that have focused on learning representations for a specific task, relatively fewer works have focused on unsupervised feature learning from raw time series for change point detection. 

%The key aspect of representation learning, also called feature learning is the intelligent extraction of informative features from the data so that the learned representation is useful for the target task. LeCun et al. \cite{lecun2015deep} show that deep learning algorithms are in fact representation learning frameworks with representations lying on multiple levels, which are obtained by multi-layer non-linear transformation of the input. %Each transformation converts the representation obtained at the current level to a more abstract, higher level.%
%Bengio et al. \cite{bengio2013representation, bengio2009learning} explain that the earlier layers of a deep neural network learn more generic features, where deeper layers learn more relevant and detailed features that are specific to the task in hand. 

\subsection{Feature Learning For Change Point Detection}

The foundation of MMD-based change point detection is grounded on measuring the distance between the samples by applying some kernel function on the samples that projects them in a new vector space. However, the feature space learned in this method is dependent on the number of samples available. %and the problem arises when sample size across positive and negative classes.%
Chang et al. \cite{chang2018kernel}  approached this problem by generating additional samples through a generative network to mitigate the problem of insufficient samples. Despite the computational complexity, it is clear from their findings that learning a better representation can help in the two-sample test-based change point detection in comparison to performing the test in the original data space. Which leaves us with the following question: \textit{Can we learn a feature representation from time-series data using deep neural networks, to detect change points in an unsupervised way, without generating new samples?}

%To this end, we define a parameterized non-linear mapping from the data space $X$ to a lower-dimensional feature space $z$.  %{This is not trivial in the unsupervised setting as we cannot train our deep network using the labeled data as done in supervised learning. We need to train our network in such a way  so that the optimization process is guided towards learning representation that is suitable for detecting change points without specifically using the labels in the optimization.} 

\vspace{0pt}
\begin{figure}[h]
  \centering
  \includegraphics[width=0.70\linewidth]{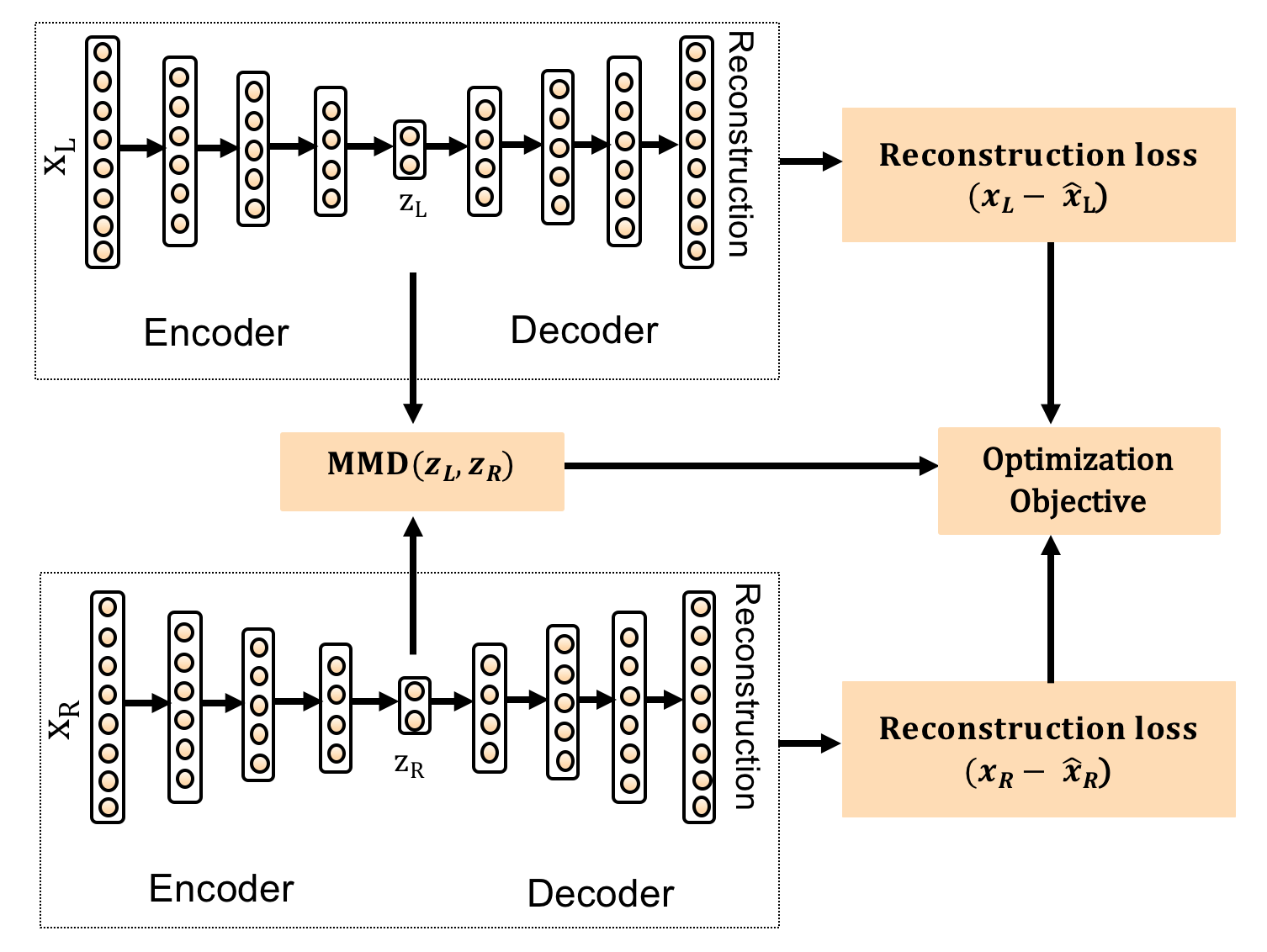}
 
  \vspace{-6pt}
  \caption{The architecture of the autoencoder network used in Cadence. We used three fully connected hidden layers with ReLU activation. For optimization, we used Adam optimizer and the learning rate was set to 0.0001. More implementation details are explained in Section~\ref{sec:7.3}}
  %\textcolor{red}{change figure 5 later}
  \vspace{-6pt}
  \label{fig:network}
\end{figure}

\subsection{Auto-encoder}
We propose to use a stacked auto-encoder to learn the feature representations from the data space to a lower dimension space. An auto-encoder is a pair of functions ($f$, $g$) designed to encode and decode data signals from an underlying domain. It takes an input ${x}$ = \{${x_1}, {x_2}, {x_3}, \dots, {x_d}\} \in {\mathbb{R}^d}$ and maps it to a hidden representation  ${z}$ = \{${z_1}, {z_2}, {z_3}, \dots, {z_{d'}}\} \in {\mathbb{R}^{d'}}$  through a deterministic mapping between the input and the code. 

%For a standard auto-encoder, the mapping  $z = f_{\theta}(x) = \phi ({Wx + b})$ is parameterized by $\theta = \{W, b\}$. Here $\phi$ is an activation function that provide non-linear transformation to the input, $W$ is a $d \times d'$ weight matrix and $b$ is the bias vector. 

%Let us consider $z$ as the feature set (latent representation) extracted by auto-encoder. Each $z_i$ is the output of a node in the bottleneck layer in the middle of auto-encoder.%

The latent representation $z$ is mapped back to a reconstructed input $\hat{x} = g_{\theta'}(z)$ = $\phi (W'x + b')$ in the input space with parameters $\theta' = \{W', b'\}$. The parameters of the autoencoder \{$\theta, \theta'$\} are optimized to minimize the $l_2$ reconstruction error $\mathcal{L}$($\boldsymbol{x}$, $\boldsymbol{\hat{x}}$):
\begin{equation}
    \theta, \theta' = {argmin_{\theta, \theta'}} \mathcal{L} (x, \hat{x}) 
\end{equation}
Here, $\mathcal{L}$  is a loss function which is a measure of the discrepancy between input $\boldsymbol{x}$ and its reconstruction $\boldsymbol{\hat{x}}$ over all available training samples.

\subsection{Learning Representation for Change Point}

In our architecture, we apply a non-linear transformation to the data X and map it to lower dimension feature space ${Z}, {f_{\theta}}$ $\colon$ ${X}$ $\rightarrow{Z}$, where $\theta$ are the learnable parameters. The optimization objective should allow learning such features that maximize the dissimilarity between the segments that are from different distributions in the latent space $Z$. Figure~\ref{fig:network} shows the neural network architecture used in Cadence.

\textbf{Objective Function Rationale.} Reconstruction loss is the commonly used loss function in neural network parameterization, defined as the squared loss between the input $x$ and its reconstructed output $\hat{x}$, ${{\Vert x-\hat{x} \Vert}^2}$. However, reconstruction loss minimizes towards better reconstruction output, whereas we are interested in better separation between segments that are from different distributions.

Towards this goal, we use an additional loss function to learn features useful to the change point detection task, which we name as MMD loss. It is defined as ${MMD}({z_L}, {z_R})$, where ${z_L}$ and ${z_R}$ are the latent representation of the past and current segments ${x_L}$ and ${x_R}$ learned in the optimization process. Note that, our objective is to learn a representation through the training that separates the dissimilar segments better in the learned feature space. While two-sample test demands for MMD to be large ($>0$) to detect a change-point, %maximizing MMD between segment pairs through loss function directly fails to accomplish that as MMD is then maximized not only for segments that are from different distributions but also for segments that \textit{are} from same distributions.%
maximizing MMD between the pairs directly will maximize the distance between samples for both when P = Q (same distribution) and P $\neq$ Q (different distribution). This will increase the false positive rate in the test statistic and will fail to achieve the change point detection goal.

Hence, we use MMD loss as an optimization objective that allows learning such features so that both reconstruction loss and MMD are minimized for segment pairs. %with the same distribution (close to 0).%
This strategy is motivated by the intuition that MMD contribution in the loss function (and optimization) comes from only pairs from different distributions. %We also experimented by explicitly maximizing MMD loss through the objective function and the proposed strategy performed better.%
One reason for this strategy's effectiveness can be minimizing MMD loss helps the latent space towards better segmentation in conjunction with reconstruction objective, whereas explicitly maximizing MMD loss moves further away from reconstruction objective space. The magnitude of MMD between the pairs in the lower dimension is much less than the reconstruction loss (between the sample and its reconstructed approximation). Hence, the total loss function value per epoch is dominated by reconstruction loss. We therefore use a weighting factor $\beta$ to MMD to balance its effect on the loss function value (please see section \ref{sec:7.3} and \ref{sec:8.2} for details on the choice of $\beta$ value).

%For segment pairs from different distributions, the dominant component of the loss function, reconstruction loss, is prioritized and minimized. MMD becomes greater than 0 in such cases, which allows detecting change point among the dissimilar samples. 

The objective function for our algorithm (Algorithm 1) thus becomes,
\textit{Reconstruction loss} + \textit{MMD loss}
\begin{equation}
    {{\Vert x_L-\hat{x_L} \Vert}^2} +{{\Vert x_R-\hat{x_R} \Vert}^2} + {\beta}{M_k}({z_L}, {z_R})
\end{equation}

\begin{algorithm}
    \SetKwInOut{Input}{Input}
    \SetKwInOut{Output}{Output}

    \Input{$w$ window size, $\alpha$ learning rate, ${n_c}$ number of iterations}
    
    \Output{change point scores}
    
    \underline{Learning}
    
    \While{${M_k}({Z^L}, {Z^R}) < \epsilon$}
    {
    \For{$t = 1, \dots {n_c}$}
    {
    Sample a minibatch ${X_t} = {({X^L_t}, {X^R_t})}$\;
        
    gradient $(\theta)$ $\leftarrow{}$ $MSE$ ($X^L$, $\hat{X^L}$) + $MSE$ ($X^R$, $\hat{X^R}$) + $M_k(Z^L, Z^R) $\;
        
    $\theta$ $\leftarrow{}$ $\alpha \cdot$ Adam ($\theta$, gradient $(\theta)$)\;
    }
    }
    
    \underline{Inference} 
    
    feature space mapping, ${f_{\theta}}$ $\colon$ $X^L, X^R$ $\rightarrow$ ${Z^L, Z^R}$\;
    
    ${M_k}({Z^L}, {Z^R}) \sim$ change point scores\;
    
    \Return change point scores;

\caption{\textbf{our proposed algorithm}}
\end{algorithm}

 \rev{For change point detection, we train the autoencoder in the learning phase (Algorithm 1) to learn a latent feature representation from each sample pair specifically with the objective function in (6). In the inference phase, we use the trained network to project incoming sample pairs in the latent space and perform a two-sample test based on the MMD computed in the latent space.

 \textbf{Change Point Detection.} To identify change points,  we first apply a smoothing filter on this score and then use the resultant score as change point probability for each time point. From this set of probabilities, we find local maxima points using a threshold of 40\% of the maximum change point score. We empirically observed that this threshold has worked well for the datasets used in the experiments. %We compare score at each time point to the threshold to obtain local maxima points as change point decisions.%
This relieves from relying on a user-defined threshold decided from domain-specific knowledge.
}
 
 \begin{figure}[h]
  \centering
  \includegraphics[width=0.45\textwidth]{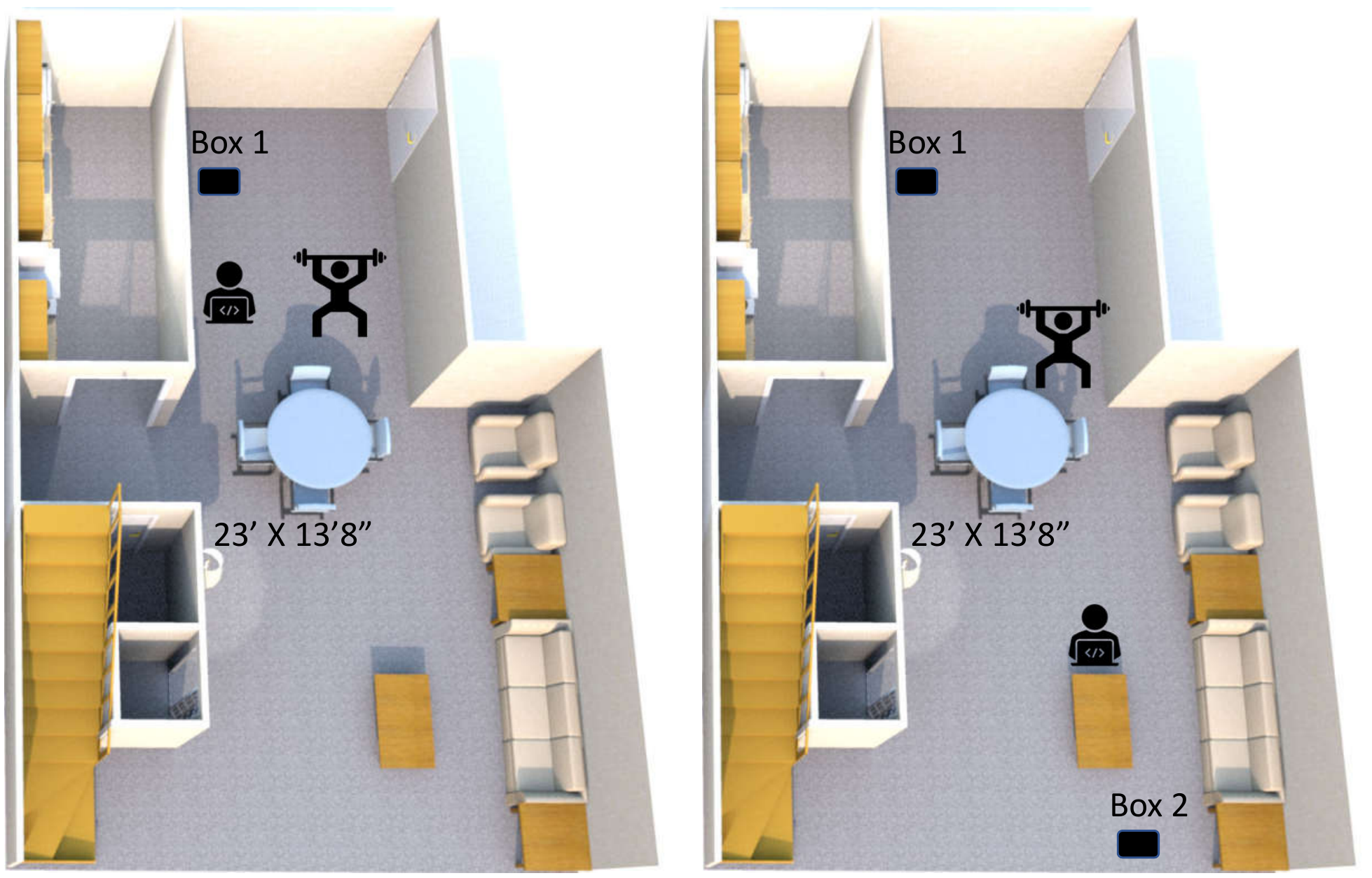}
  \caption{Data collection setup using multiple sensor boxes of Maestro in a room}
  \label{fig:maestro_layout}
\end{figure}

\section{Application: Activity Sensing}\label{sec:7}

To demonstrate the use of our proposed change-point detection method in a real-world sensing application, in this section we apply Cadence for ambient sensing. \rev{ In our application, we used Cadence to sense changes in a physical environment for human activity sensing from time series consisting of ambient sensor observations.} We use a sensor suite comprised of non-intrusive sensors to detect events in daily life in a room setting for indoor activity sensing, which we call \textit{Maestro}. Maestro includes various off-the-shelf low-cost sensors such as accelerometer, magnetometer, illumination sensor, humidity, temperature, and audio sensor to capture occupant interaction with ambient objects using 18 channels of information from 9 physical sensors. \rev{A Maestro unit is powered by a Raspberry Pi3B+©, and the data collection is managed by an Arduino Uno©. The sampling rate for the sensors is 30 samples/sec. Data collected from a Maestro unit over every 10 seconds are stacked and uploaded to a remote server via WIFI to be stored in a TimeScaleDB database. %This also allows real-time labeling, where a label can be assigned by specifying the start and end of the event to enable fine-grained labeling for short-duration event(s) as well.%
}

\textbf{Experiment.} We conduct a real-world experiment using a typical household of 3 people as the test-bed. \rev{We put the sensor suite at a location of the room around which most interactions of the room occupants occurred during the experiment. We allow Maestro, the sensor suite, to record all activities for several hours and periodically upload them to the database in batches of 10 seconds. The data does not contain any identifying information to be linked back to the subjects from whom it was collected. During the experiment, occupants of the room naturally interacted with different ambient objects and household appliances.} The experimenter (also a member of the household) annotated the occupancy of the room during the experiment period which we use as ground-truth labels. %from a camera feed, which is also part of the sensor suite but used only for annotation purposes.%
The layout of the testbed is shown in Figure~\ref{fig:maestro_layout}.

\begin{figure}[t]
  \centering
  \includegraphics[width=0.85\linewidth]{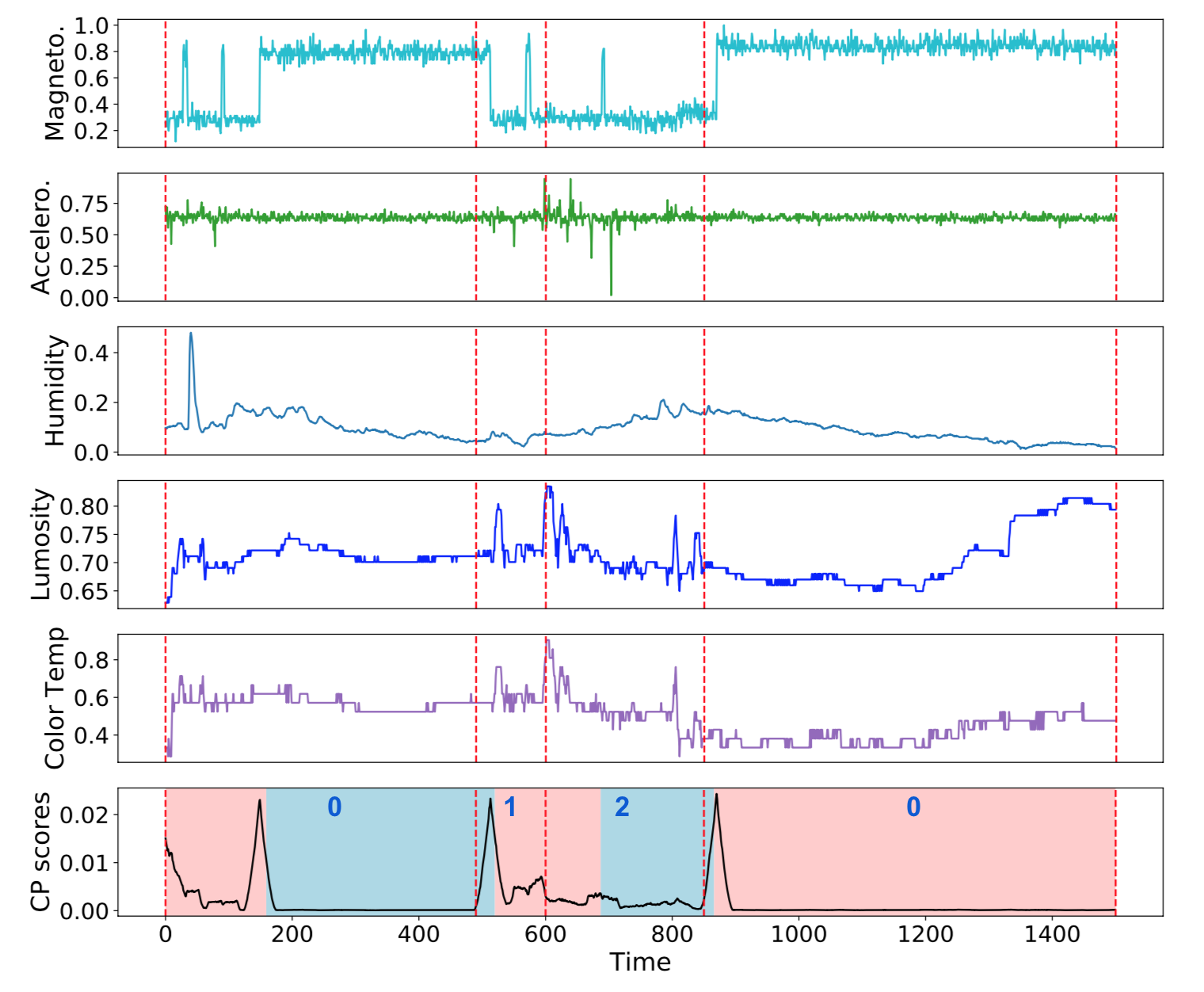}
  \caption{Performance of Cadence in an ambient sensing application. The first five plots show sensor readings obtained from different sensors. Red dashed vertical lines indicate annotations from the human labeler. The bottom plot shows change-point scores in black. \rev{Segments based on identified change point decisions are shown in alternating shaded regions of red and blue.} True occupant levels in each segment are indicated in blue. We observe that there is a rough alignment between human-labeled occupancy change and change point possibility score from Cadence. A change-point score above the baseline of 0 occurs when changes are observed in raw sensor readings. Also note that there are events that are apparent from the sensor readings (around t=180), which is ignored by the occupancy annotator (absent among ground truth change points), but is apparent from change-point score.}
  \label{fig:maestro_cpd}
  \vspace{-1mm}
\end{figure}

\textbf{Result.} We consider the occupancy detection application where our goal is to detect the locations in the time series when the occupancy level of the room has changed based on activities captured by the sensor readings. In Figure~\ref{fig:maestro_cpd}, top five plots show raw readings from five selected sensors. The bottom plot shows change point scores per time-point to denote the possibility for each time-point of being a change-point using our method. \rev{The segments obtained from Cadence are shown in alternating shaded red and blue regions.} \textit{We note that the change point scores above 0 correlate with sensor events and hence do not always align with human-provided occupancy change-point.} Ground-truth annotations are obtained from human, which is dependent on the human perception scale about context. 

Qualitatively, we observe that there is a rough alignment between human annotation of change-point and the change point score, and do not always match the peak score. \textit{One explanation for this can be events are happening in the ambient environment of that room that is ignored by the application-specific annotation (room occupancy, in this case), yet observed by sensors that %produce a high change point score (notice the segments labeled with  $0$ occupant).%
can be useful for fine-grained sensing and labeling.} This type of small interval events or actions may not be indicative of
context-level changes such as room occupancy but can be important for in-situ sensing of interesting contextual events, partitioning, labeling, and classification in the wild (recall Partitioning step in Figure~\ref{fig:senseMLP}). Our change point detection algorithm identifies these events as change points. 

\rev{\textbf{Evaluation.} We obtain Area under the ROC (Receiver Operating Characteristic) Curve, or AUC score, of 0.827 for our experiment to detect change points in the sensor streams from Maestro %into contextual segments%
using Cadence. There are existing works that perform occupancy detection using data from ambient sensors \cite{ANG, Zimmermann} as well as electricity consumption data from smart meter \cite{Feng} in a supervised setting. To compare with existing occupancy detection systems, we use supervised occupancy detection systems using ambient sensor data based on machine learning methods such as Multilayer perceptron (MLP), Random Forest (RF), and Support Vector Machine (SVM) for performance evaluation. Our results show that Cadence, being unsupervised, provides performance (AUC: 0.827) comparable with supervised learning methods used in the literature (SVM: 0.826, MLP: 0.496, RF: 0.097). 
Cadence provides an easy to train, fast, unsupervised approach to identify change-points in multi-dimensional unlabeled sensor streams that can reduce human effort in partitioning time-series.
}

\section{Experimental Evaluation}\label{sec:8}

In this section, we describe details of the experimental evaluation of the proposed Cadence framework. We highlight the existing algorithms with which we compare for benchmark evaluation, the datasets used for evaluation, followed by implementation notes on our proposed framework.

\subsection{Algorithms}

% \textcolor{red}{We evaluate the performance of our algorithm, \textit{Cadence} with other CPD methods available in the literature. Towards this goal, we select 4 existing methods to use as the baseline which use approaches related to Cadence. We include \textit{RDR-CPD} \cite{liu2013change}, a density-ratio estimation-based method to detect change-points using Gaussian Kernel, and \textit{Mstats-CPD} \cite{XieYchange}, which uses kernel MMD for two-sample test under the condition that large amount of background data is available. These methods measure distribution distance using a kernel-based measure in the original data space, hence performance can be dependent on the amount of data available. \textit{LSTNet} \cite{Liu2018ClassifierTS} uses a Recurrent Neural Network (RNN) parameterized kernels for the kernel two-sample test. 
% %and performs better compared to methods testing in the data space.%
% \textit{KL-CPD} \cite{chang2018kernel}, which is the current baseline method, uses a RNN parameterized kernel learning approach by creating additional samples to mitigate the problem of insufficient samples when sufficient data is not available.}

\begin{table}[]
\centering
    \begin{tabular}{l||l||l||l||l}
    \hline
    Dataset & \textbf{Application Domain} & \textbf{\# of instances} & \textbf{\# of CPs} & \textbf{\% of CPs} \\ \hline
    \textbf{Beedance} & Biology & 1057         & 19                 & 1.8\%                    \\ \hline
    \textbf{Yahoo}   & Network Traffic & 1440         & 32                 & 2.2\%                    \\ \hline
    \textbf{HASC}    &  Human Activity Sensing & 39397        & 65                 & 0.16\%                   \\ \hline
    \textbf{Fishkiller} & Environmental Science & 45175        & 899                & 1.99\%                   \\ \hline
    \end{tabular}
   
\caption{Dataset Statistics. CP indicates change-points based on annotation}
\label{tab:dataset}
\vspace{-10mm}
\end{table}

We evaluate the performance of our algorithm, \textit{Cadence}, with other CPD methods available in the literature for the properties mentioned in section 4.1. We select five existing methods to use to compare against Cadence. We include \textit{RDR-CPD}~\cite{liu2013change}, a density-ratio estimation-based method to detect changepoints using a Gaussian Kernel, and \textit{Mstats-CPD}~\cite{XieYchange} using an MMD kernel for the two-sample test under the condition that a large amount of data is available. %These methods measure distribution distance using a kernel-based measure in the original data space; hence, performance can depend on the amount of data available.%
\textit{LSTNet}~\cite{LSTNet2017} uses an RNN parameterized kernel for the kernel two-sample test. \rev{\textit{TIRE}~\cite{deryck2021change} uses an autoencoder to extract time and frequency domain features and detect changepoints using euclidean distance-based dissimilarity.} \textit{KL-CPD}~\cite{chang2018kernel}, the current top-performing method in the literature, uses an RNN kernel learning approach by creating additional samples when sufficient data is not available.

\subsection{Dataset}

\begin{figure}[h]
  \centering
  \includegraphics[width=0.80\linewidth]{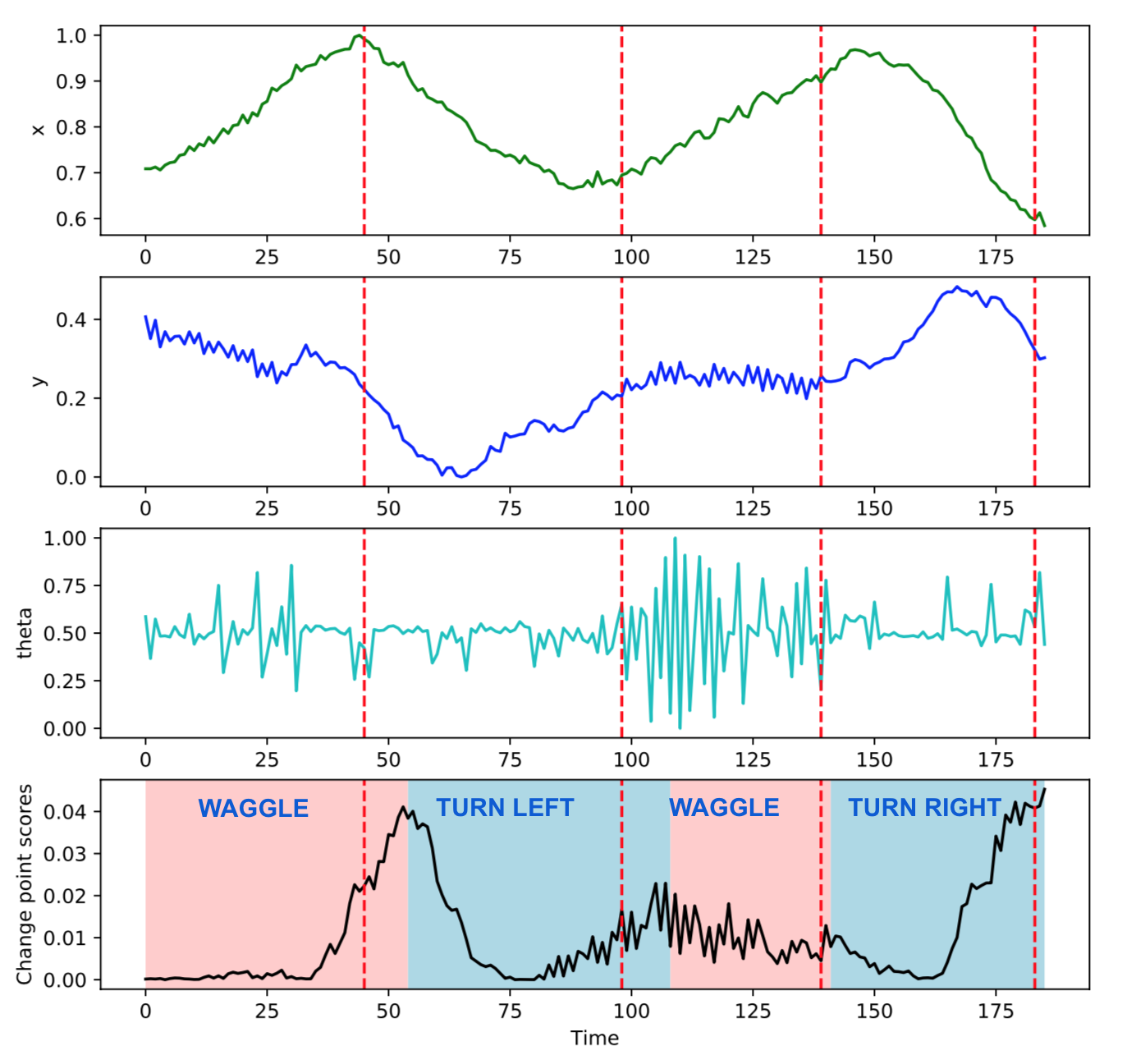}
  \caption{Predicted change point score on Beedance data using our method. In the first three plots are waggle dance signals of x, y co-ordinates, and phase angles of honey bee movement. The last plot (in black) is MMD scores obtained from our framework. Red dashed vertical lines denote ground truth change point labels (not used in prediction). \rev{Segments based on identified change point decisions are shown in alternating shaded regions of red and blue.}}
  \label{fig:beedance}
  \vspace{-4mm}
\end{figure}

\rev{
We evaluate our proposed approach on 4 publicly available datasets used in change point detection literature for benchmarking. The datasets come from real-life application domains of biology, environmental science, network traffic loads, and human activity recognition. We use these datasets in our evaluation as they are the closest approximations to IoT applications that use time series partitioning, are publicly available and, are frequently used in CPD literature for benchmarking.
%The 4 datasets we have used cover different application areas where time series change point detection can be useful. They all consist of time series related to the IoT domain that senses dynamics of a physical system and changes observed in the system as a sensor.%
We perform the evaluation on the following datasets:

\begin{itemize}
    \item {\bf Bee-dance.} %Honey bee waggle dance is one of the most popular communication systems in the animal world. %
    The bee-dance dataset \cite{beedance} consists of the pixel locations in x and y dimensions and angle differences of dancer bee
movements. The dance movements represent a three-stage communication method %through which%
 between forager bees %and unemployed forager bees %
 about the orientation and distance to the food sources and water. Biologists have been interested in identifying the change point of transitioning from one stage to another of the dance and map the pattern of each stage with the information it conveys. \rev{Change point labels denote different segments (phases) of the dance in this case.} Figure~\ref{fig:beedance} shows a subset of the signals in the data along with predicted change point scores from the proposed method.

    \item {\bf Yahoo.} The yahoo dataset \cite{yahoo} consists of real time-series of network traffic data with anomalies, which has been manually annotated as change-points. The data consists of time-series representing metrics of various Yahoo services (e.g. CPU
utilization, memory, network traffic, etc.). \rev{Change point labels here denote an abrupt rise in the incoming traffic.}

    \item { \bf HASC.} This dataset is a subset of the Human Activity Sensing Consortium (HASC) challenge 2011 dataset \cite{hasc}, which provides human activity information collected by portable three-axis accelerometers. The goal of change-point detection is to segment the time series data into 6 activities: stay, walk, jog, skip, stair up, and stair down. \rev{Change point labels denote the segment boundaries of the activities for this dataset.}
    
    \item { \bf Fishkiller.} Fishkiller dataset \cite{osborne2011machine} records sensor readings of water level from a river dam in British Columbia, Canada. When the dam does not operate normally, the water level rapidly oscillates in a short period, %causing trouble for the fish. This%
    which leads to rapid water level drops followed by salmon fish becoming stranded and suffocating. \rev{The beginning and end of every water oscillation (the reason for fish-killing) are treated as change points and change point labels, in this case, denote abnormal functioning of the dam.}

\end{itemize}
}

A summary of the data statistics is described in Table~\ref{tab:dataset}. We preprocess all 4 datasets by normalizing each dimension of the data in the range of [0, 1]. Note that the datasets have a varied sampling rate and are available with manual annotation of change points which we use as ground truth in evaluation. The unit of the window size is hence represented by the number of steps on the time dimension. Following \cite{liu2013change, saatcci2010gaussian, chang2018kernel}, the datasets are divided into train-validation-test split of 60\%, 20\%, and 20\% in chronological order. %Note that the class balance of change-points between the splits largely varies due to the highly non-stationary nature of these real-life observations. All 4 datasets used in the evaluation experiments are publicly available.

\subsection{Evaluation Metric}
For the quantitative evaluation of our algorithm, %the standard evaluation metric for CPD techniques in the literature. Towards this goal, %
we use the Receiver Operating Characteristic (ROC) curve of change point detection results as the evaluation metric. ROC curve is a plot of true positive rate (the fraction of positive examples that are correctly labeled by the classifier) against the false positive rate (the fraction of negative examples that are misclassified as positive) under different classification thresholds of the test statistic. The area under the ROC curve (AUC) is an aggregate measure of the classifier's performance across all possible classification thresholds. While there are alternate evaluation metrics for binary classifiers such as Precision-Recall (PR) curve, ROC curve and AUC score are not biased towards classification models that perform well on the minority class at the expense of the majority class \cite{he2009learning, davis2006relationship}. This is a property that is attractive when dealing with imbalanced data.
Additionally, AUC is the metric commonly used in CPD literature \cite{ chang2018kernel, liu2013change, XieYchange} and is used as the performance metric to obtain a comparative evaluation of detection performance in this work. \rev{To calculate the AUC from the ROC curve, we obtain true positive rate and false positive rate under different change point probability thresholds, %using scikit-learn implementation of ROC curve function,%
which is used to plot the ROC curve. AUC is calculated by aggregating the entire two-dimensional area under the ROC curve.}

%ROC curve may be less informative in evaluating classifier performance on imbalanced data as it focuses equally on positive and negative class (to)

\subsection{Implementation Notes}\label{sec:7.3} Determining the hyperparameters of the deep neural network for unsupervised learning is non-trivial. % Hence, we avoid tuning the hyperparameters based on cross-validation to retain the class imbalance between the data splits.%
Inspired by van der Maaten et al, \cite{maaten2008visualizing}, we set the autoencoder dimensions to ${{d}-40-30-20-{z}}$ for all the datasets, where $d$ denotes the dimension of the time-series sample (data space) and $z$ denotes the dimension of latent representation (feature space). All the layers are fully connected dense layers with ReLU activation functions. %Note that unlike in their paper where image data was used, our data samples are created by sliding window approach meaning that the dimension of the data space is dependent on the window size. Here $\mathds{d}$ is the dimension of data that varies across all datasets. Tuning the hyperparameters based on cross-validation was avoided to retain the class imbalance between the data splits.%
We have used the same architecture and hyperparameters for all the datasets unless specifically mentioned otherwise.

We initialize the weights prior to training with Kaiming initialization for training deep neural networks with ReLU activation for better convergence \cite{he2015delving}. The deep autoencoder is trained for 2000 iterations without using dropout. The learning rate is set to 0.0001. We used Adam optimizer \cite{kingma2014adam} for all our experiments. All of the above parameters are kept fixed to achieve a reasonably low objective function value and are held constant across all datasets. Default value of $\beta = 1.0$ , $w = 25$ and $z = 3$ are used across all datasets unless mentioned otherwise. %While dataset-specific tuning may provide more accurate results (as we observe in our experimentation with $\beta$ value in section~\ref{sec:8.2}), we refrain from doing such unrealistic fine-tuning to focus on finding a method that is generalizable. 

%We run each of our experiments for 10 restarts and pick the average.% 
Note that training is fully unsupervised for our method with no information to change-point labels. We use performance on the validation set as a stopping criterion for the training procedure and select the best-parameterized model. All final results are reported based on the test set.

\section{Results and Analysis}\label{sec:9}

\subsection{Benchmark Evaluation}
%We evaluate the performance of our algorithm with other CPD methods available in the literature.%
In Table~\ref{tab:comparion_cpd}, we report the results of performance evaluation based on Area-Under-Curve (AUC) score from the ROC curve. AUC performances of existing baseline CPD techniques (except \textit{TIRE}) are reported from \cite{chang2018kernel}.

%\begin{figure*}[htp]
  %\centering
  %\subfigure[random caption 1]{\includegraphics[scale=0.30]{Figures/ablation.png}}\quad
  %\subfigure[random caption 2]{\includegraphics[scale=0.30]{Figures/ablation_window.png}}
%\end{figure*}

\iffalse
\begin{table*}[t] 
\centering

%\color{red}
\begin{tabular}{*{7}{c} }
%{c||c||c||c||c||c||c}
\hline
Dataset  & \textbf{RDR-CPD \cite{liu2013change}} & \textbf{Mstats-CPD \cite{XieYchange}} & \textbf{LSTNet \cite{Liu2018ClassifierTS}} & \textbf{TIRE\cite{deryck2021change}} &  \textbf{KLCPD \cite{chang2018kernel}}  & \textbf{Cadence} \\ \hline
\textbf{Beedance}   & 0.5197  & 0.5616     & 0.6168 & 0.7001 & 0.6767 & \textbf{0.7541} \\ \hline
\textbf{Yahoo}      & 0.6029  & 0.6961     & 0.8863 & 0.7083 & 0.9146 & \textbf{0.9774} \\ \hline
\textbf{HASC}       & 0.4217  & 0.5199     & 0.5077 & 0.6504 & 0.6490 & \textbf{0.6525} \\ \hline
\textbf{Fishkiller} & 0.4942  & 0.6392     & 0.9127 & 0.5753 & \textbf{0.9596} & 0.9477 \\ \hline
\end{tabular}

\caption{Comparative Evaluation on four datasets based on AUC score. Our method achieves best AUC in three out of the four datasets. On Fishkiller dataset it is second best by a small margin (1.2\%)}
\label{tab:comparion_cpd}
\vspace{-4mm}
\end{table*}
\fi

\begin{table}[]
\begin{tabular}{c||cccc}
\hline
Dataset    & \textbf{Beedance} & \textbf{Yahoo}  & \textbf{HASC}   & \textbf{Fishkiller} \\ \hline
\textbf{RDR-CPD}    & 0.5197   & 0.6029 & 0.4217 & 0.4942     \\ \hline
\textbf{Mstats-CPD} & 0.5616   & 0.6961 & 0.5199 & 0.6392     \\ \hline
\textbf{LSTNet}     & 0.6168   & 0.8863 & 0.5077 & 0.9127     \\ \hline
\textbf{TIRE}       & 0.7001   & 0.7083 & 0.6504 & 0.5753     \\ \hline
\textbf{KLCPD}      & 0.6767   & 0.9146 & 0.6490 & \textbf{0.9596}     \\ \hline
\textbf{Cadence}    & \textbf{0.7541}   & \textbf{0.9774} & \textbf{0.6525} & 0.9477   \\  \hline
\end{tabular}

\caption{Comparative Evaluation on four datasets based on AUC score. Our method achieves best AUC in three out of the four datasets. On Fishkiller dataset it is second best by a small margin (1.2\%)}
\label{tab:comparion_cpd}
\vspace{-4mm}
\end{table}

\begin{figure}[ht]
\centering
\begin{minipage}[b]{0.49\textwidth}
\centering
\includegraphics[width=0.95\textwidth]{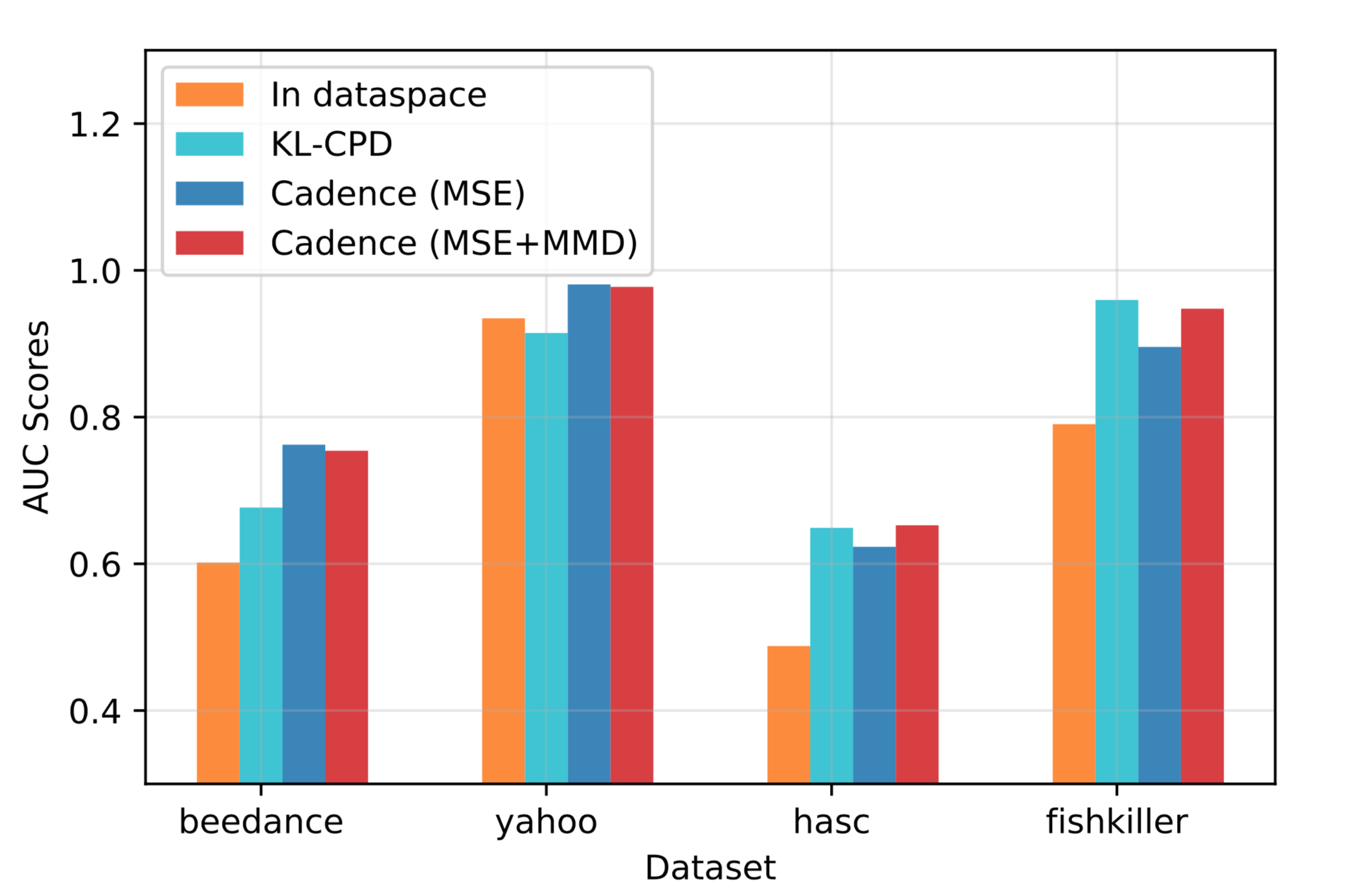}
\caption{Ablation test of our method}
\label{fig:ablation}
\end{minipage}
\begin{minipage}[b]{0.49\textwidth}
\centering
\includegraphics[width=0.94\textwidth]{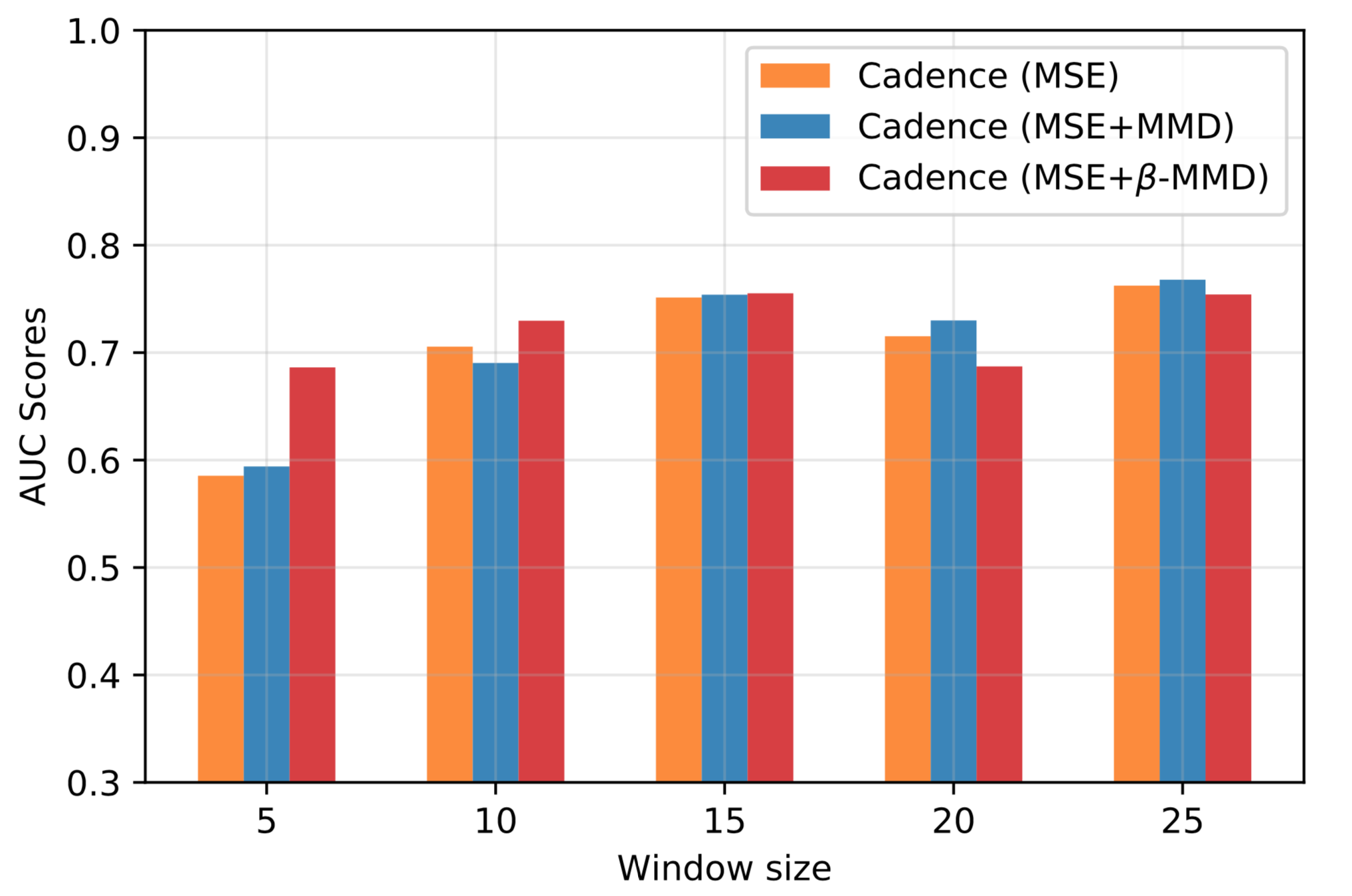}
\caption{Ablation test for different window size}
\label{fig:window_size}
\end{minipage}
%\caption{}
\vspace{-4mm}
\end{figure}

We compare with baseline methods \textit{RDR-CPD} \cite{liu2013change}, which uses density-ratio estimation to detect change-points using Gaussian Kernel, and \textit{Mstats-CPD} \cite{XieYchange}, which uses kernel MMD for two-sample test when a large amount of background data is available. These methods measure distribution distance using a kernel-based measure in the original data space, hence performance can be dependent on the amount of data available. \textit{LSTNet} \cite{Liu2018ClassifierTS} uses RNN parameterized kernels for the kernel two-sample test and performs better compared to methods testing in the data space. However, for highly non-stationary and complex data (Beedance and HASC), the results are much inferior. As the changes in Yahoo and Fishkiller dataset more resembles anomaly and fault detection problem, this also shows the effectiveness of RNN-based methods in anomaly data and limitation to generalize to broader CPD problems. 
\rev{\textit{TIRE}~\cite{deryck2021change} uses an autoencoder-based approach similar to \textit{Cadence} to extract time and frequency domain features and finding change points using euclidean distance-based dissimilarity from the features. While it performs well on Beedance and HASC, on datasets such as Yahoo and Fishkiller that resemble anomaly data it performs worse due to its dependence on parameters based on domain knowledge.}

\textit{KL-CPD} \cite{chang2018kernel}, the current baseline method, uses an RNN parameterized kernel learning approach by creating additional samples to mitigate the problem of insufficient samples when sufficient data are not available. Notice that our proposed method, Cadence, shows performance comparable for both datasets (HASC and Fishkiller) where the sample size is large, but significantly outperforms KL-CPD for smaller datasets (Beedance and Yahoo). The AUC scores achieved by our method on these two datasets are \textbf{10.2\%} and \textbf{6.4\%} higher than that of KL-CPD respectively. This indicates the capability of deep autoencoders to learn meaningful, well-distinguished features through unsupervised training that is useful in performing kernel two-sample tests in the feature space.

\subsection{Ablation Test on Learning Embedding}\label{sec:8.2}

To explore how Cadence performs under different feature spaces, we perform an ablation study where our method performs kernel two-sample test with different feature mappings. \rev{To validate training a deep network for feature learning, in \textbf{In dataspace} the kernel-based two-sample tests are performed in the original dataspace between two segments without using a deep autoencoder.} We use \textbf{KL-CPD} %, the RNN-parameterized kernel learning method using synthetic samples developed%
by Chang et al.~\cite{chang2018kernel} as a baseline method. For \textbf{Cadence (MSE)}, kernel two-sample tests are performed on the feature space learned by optimizing reconstruction loss only. \textbf{Cadence (MSE+MMD)}, uses maximum mean discrepancy in its objective function alongside reconstruction loss.

The results are shown in Figure~\ref{fig:ablation}. \rev{We observed improvement in AUC scores in all 4 datasets with up to 30\% improvement over solution achieved without using a deep autoencoder (In dataspace).} We further notice the generation of additional synthetic samples does offer an advantage over performing the two-sample test in data space [cyan bars rises over orange]. %However, the objective of computationally expensive generation approach is to mitigate the problem of class imbalance that is present in these datasets.%
Cadence (MSE) [blue bars] learns the feature space without generating additional samples to balance the classes with sufficient samples. In Cadence (MSE+MMD) [red bars], our network learns more fine-grained parameterization geared towards identifying changes between segments by incorporating MMD between segments in the optimization objective. Notice that our method performs significantly better for smaller datasets (Beedance and Yahoo) than KL-CPD. We observe the advantage of adding a discrepancy measure (MMD) in the objective function for larger datasets (HASC and Fishkiller), where the composite loss function provides leverage over reconstruction loss only (red bar rises over blue bars). This validates our choice of the loss function and its advantage in our CPD goal.

\textbf{Ablation Test for Window Size.} We further demonstrate how Cadence performs with different optimization objectives across different window sizes, as shown in Figure~\ref{fig:window_size}. We trained Cadence with segments created using a sliding window, $w$ of size \{5, 10, 15, 20, 25\} and three versions of optimization objective: reconstruction loss only (MSE), composite loss of reconstruction plus MMD (MSE + MMD), and weighted composite loss (MSE +$\beta$-MMD). 
The value of $\beta$ was selected from {0.01, 0.1, 10, 100} and Cadence (MSE+MMD) is the baseline where $\beta$=1.0. Our results show that for smaller window sizes, tuning the $\beta$ parameter in Cadence-$\beta$-MMD provides leverage over the other two cases, and $\beta$=10 performs the best in such case. As the window size increases, the effect of tuning becomes negligible, and all three versions of Cadence perform similarly ($w=25$).  We also see that Cadence performs better with more data (upward trend as $w$ gets larger).

We note that the two-sample test on features learned via Cadence (MSE) provides better performance over testing in the original dataspace (Figure~\ref{fig:ablation}).  We do not observe a similar gain for Cadence (MSE + MMD) %. Using Cadence (MSE + MMD), we obtained %
 (up to 7\% performance improvement above Cadence (MSE)). We make the same observation in Figure~\ref{fig:window_size}.  Using Cadence (MSE) achieves good performance for larger window sizes. However, for small window sizes, Cadence (MSE + $\beta$-MMD) provides more robust performance.
%\begin{figure*}
%\centering
%\begin{minipage}[b]{.45\textwidth}
%\includegraphics[scale=0.21]{Figures/impact_window.png}
%\caption{Effect of different window size}
%\label{fig:effect_window}
%\end{minipage}
%\qquad
%\begin{minipage}[b]{.45\textwidth}
%\includegraphics[scale=0.21]{Figures/impact_feature.png}
%\caption{Effect of different encoding dimension}
%\label{fig:effect_dim}
%\end{minipage}
%\caption{}
%\end{figure*}

\begin{figure*}
\centering
\begin{minipage}[b]{.45\textwidth}
\includegraphics[scale=0.20]{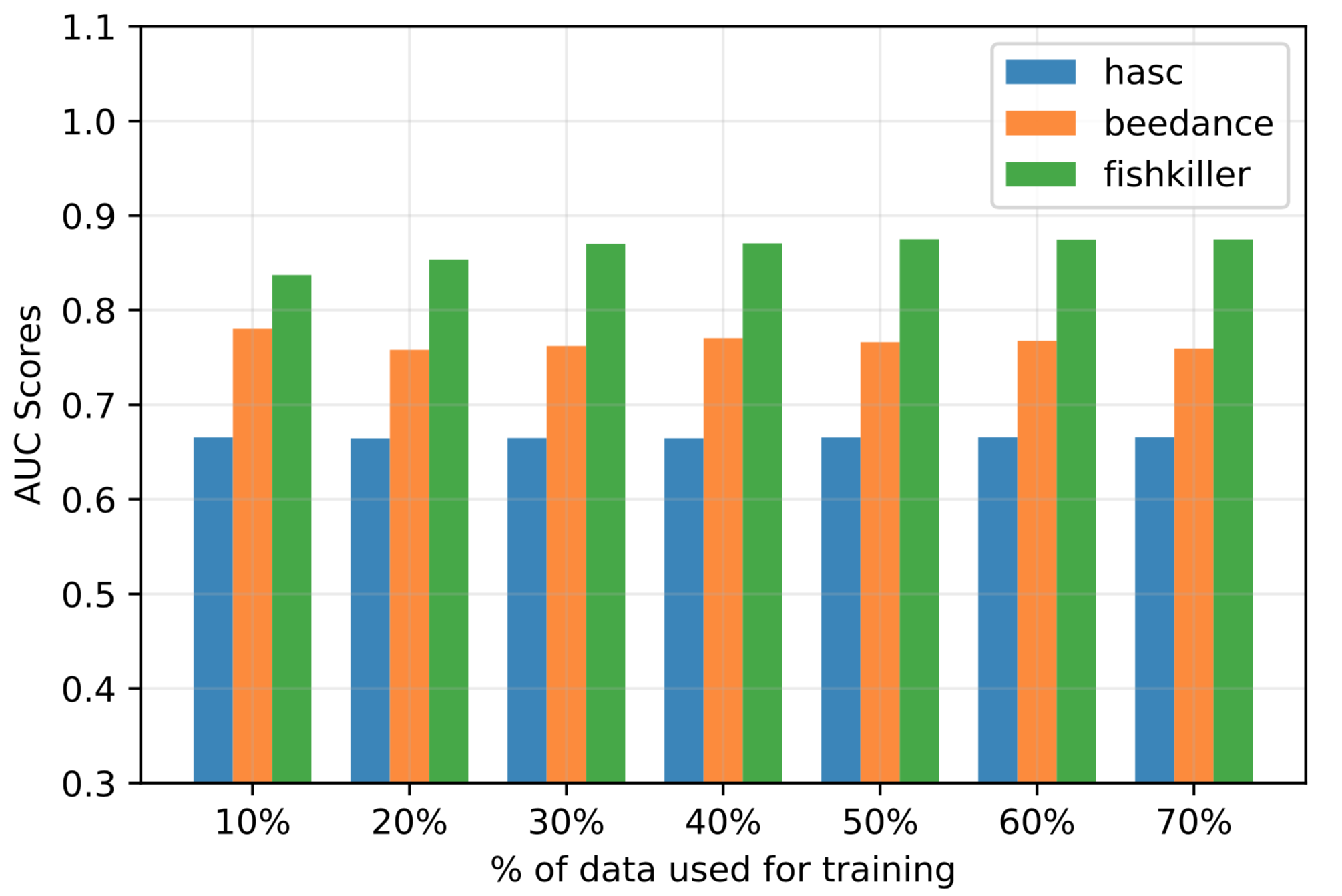}
%\caption{Caption}
\end{minipage}\qquad
\begin{minipage}[b]{.45\textwidth}
\includegraphics[scale=0.20]{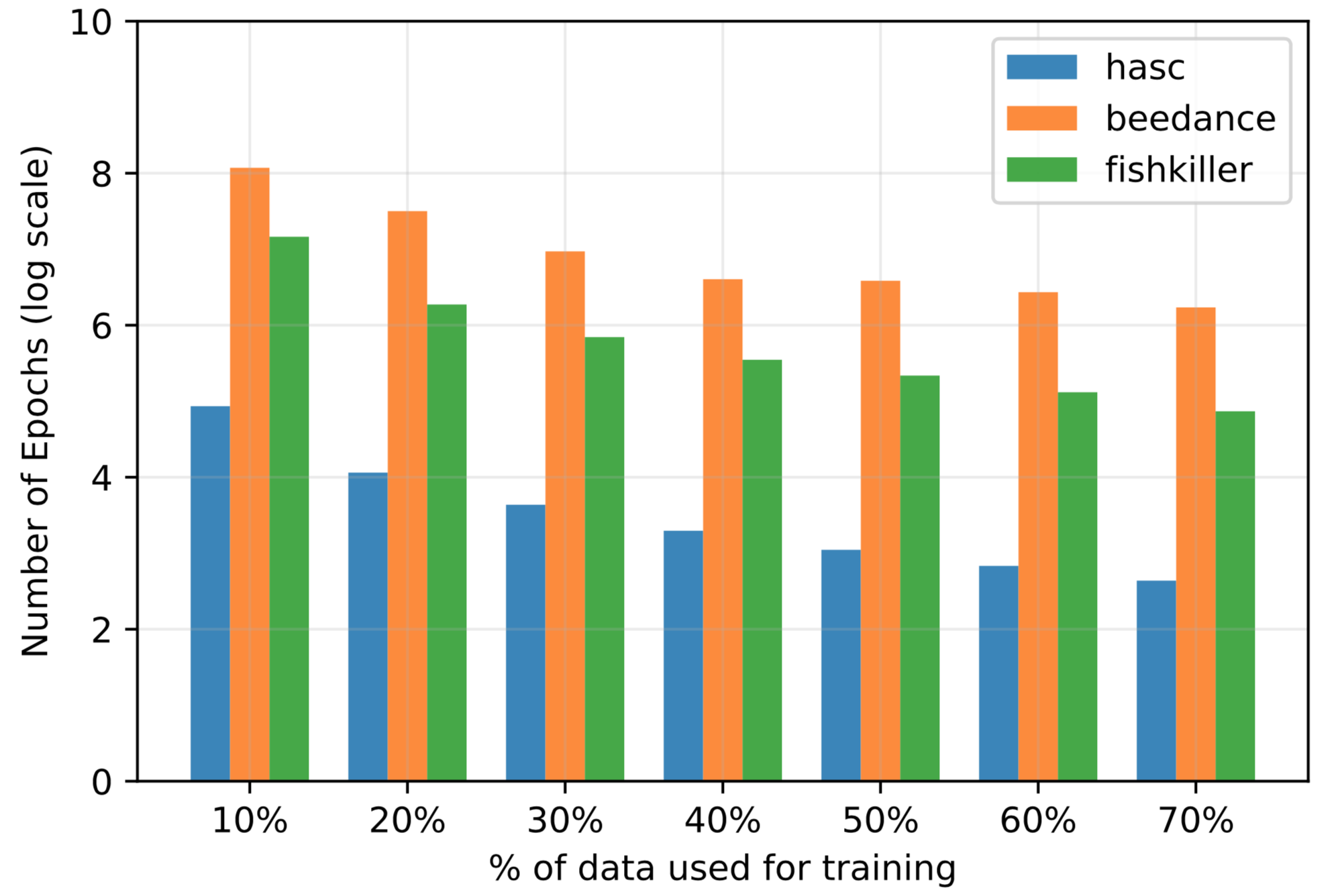}
%\caption{Caption}
\end{minipage}
\caption{Our method shows stable performance under limited training data conditions as low as 10\% of the available data. As the amount of training data decreases, our method still reaches the solution (left), however at the cost of more training epochs (right), which demonstrates sample efficiency of our proposed approach.
}
\label{fig:samples_auc_epochs}
\vspace{-4.5mm}
\end{figure*}

\subsection{Impact of number of training samples}

Recall that the amount of data available has been studied in the CPD literature as it impacts the success of kernel two-sample test based change-point detection. To demonstrate the effect of the amount of available training data on Cadence, we sample different subsets of the datasets and trained our network using fractions of training data between 10\%-70\% and report the achieved AUC on the test set (20\%). We exclude the yahoo dataset in this experiment as its change-point labels are distributed across different fractions unevenly and thus performance on data subsets does not reflect true detection performance.

\begin{figure}[h]
    \centering
    \includegraphics[width=0.45\linewidth]{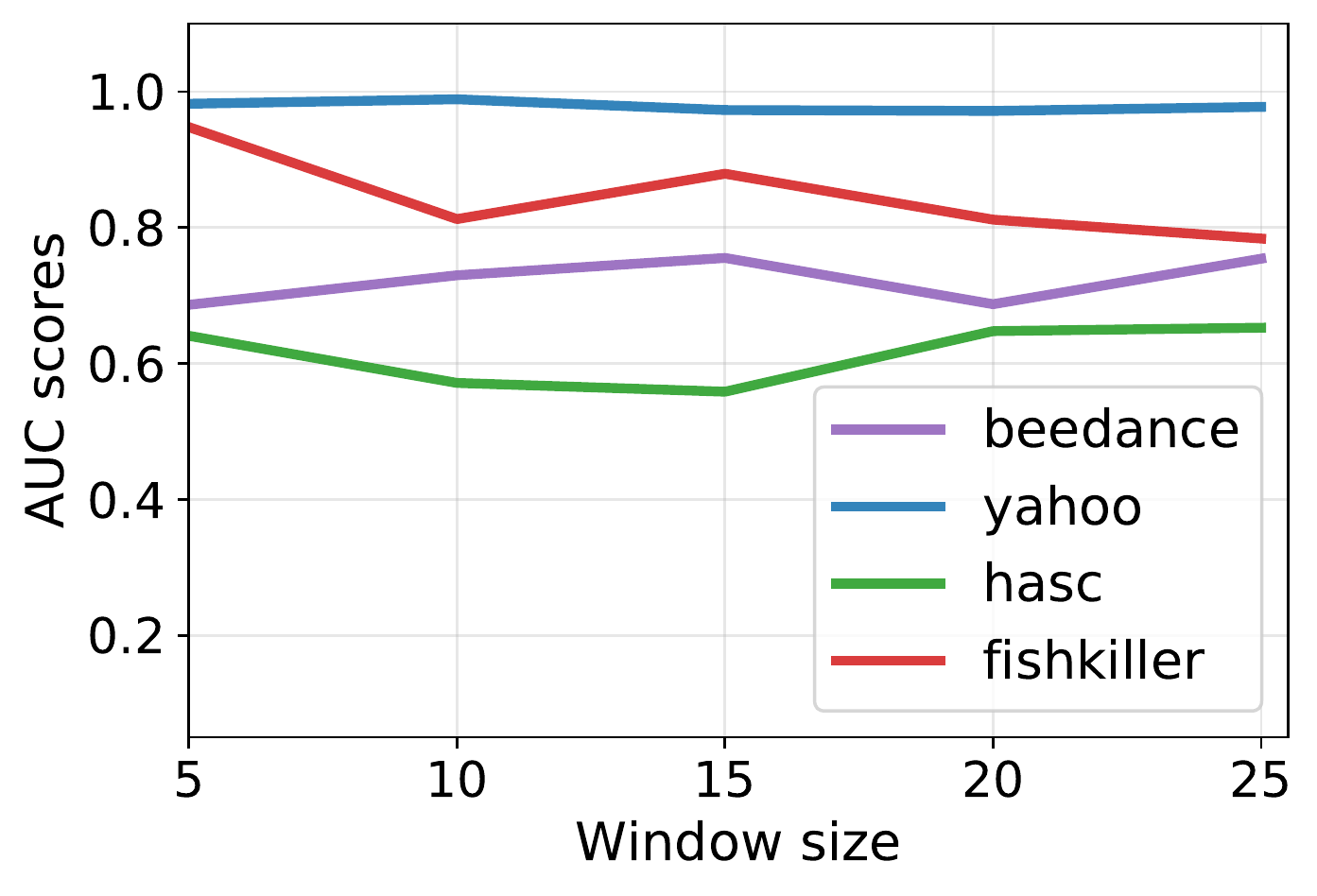}
    %\label{fig:effect_window}
    \hfil
    \includegraphics[width=0.45\linewidth]{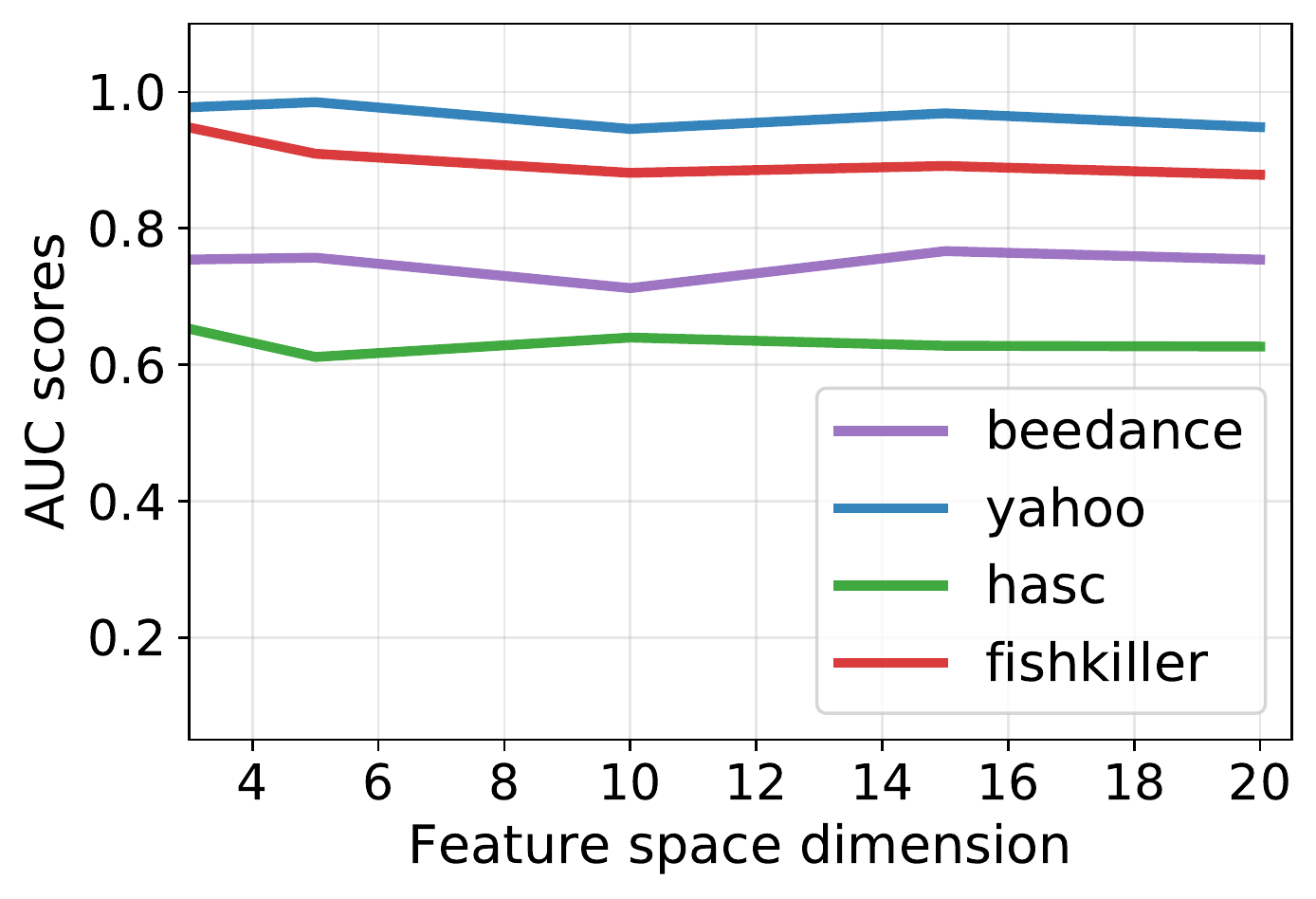}
    \caption{Effect of different window size (left) and different encoding dimension (right)}
    \label{fig:effect_dim_window}
\vspace{-2mm}
\end{figure}

\teal{We present the results obtained in Figure ~\ref{fig:samples_auc_epochs}. Cadence shows robust performance for limited training data condition as low as 10\% for all three of the datasets in Figure~\ref{fig:samples_auc_epochs} (left). This may raise concern among readers since our proposed method is unsupervised and performance deterioration is expected with less available data. However, Figure~\ref{fig:samples_auc_epochs} (right) demonstrates that robust performance is achieved at the cost of more training time to learn the useful features for a two-sample test. Specifically, as training data decreases, Cadence still reaches the solution, however at the cost of more training epochs. In other words, Cadence leverages deep neural network parameterization to obtain useful feature mapping when training data is not sufficient. This relaxes the need for creating additional samples as suggested in \cite{chang2018kernel}, but the network requires more iterations to obtain that mapping when lesser data is available.}

%put two figures in combined columns

% put two figures in single column side by side

\subsection{Impact of window size}\label{impact_window}

Finding the right size for the sliding window is a common hyperparameter choice that can influence CPD performance. \teal{We experimented with the window size, $w$, of the segments %as the tolerance of delay%
for all 4 datasets. We varied $w$ to \{5, 10, 15, 20, 25\} to observe how performance varies for different window sizes. Except for Fishkiller, all three other datasets improve or provide robust performance under different window sizes in Figure~\ref{fig:effect_dim_window} (left).} One explanation of Fishkiller performance degradation for large window size can be the oscillatory nature of the observations. The oscillatory, seasonality pattern observed in the data, which monitors river dam functioning via sensors, can influence the MMD-based test statistic between the segments depending on the duration and frequency of the pattern.

\subsection{Impact of feature space dimension}

Another implementation choice for autoencoder-based optimization is the dimension of the encoded space (feature space). We study how different feature space dimensions would affect the change point detection performance of Cadence. We varied the size of the bottleneck layer of the auto-encoder, which encodes the segments in the feature space. Deep embedding learned in the lower dimension space is dependent on the size of this layer, hence it is going to influence the performance of MMD-based two-sample test statistic as well. We varied the size of this encoder layer, $z$,  to \{3, 5, 10, 15, 20\}. We observe AUC remaining stable across these parameter settings for all four datasets in Figure~\ref{fig:effect_dim_window} (right).

%\begin{figure}[h]
  %\centering
  %\includegraphics[width=0.42\textwidth]{Figures/impact_kernel.png}
  %\setlength{\belowcaptionskip}{0.0pt}
  %\caption{Effect of different kernel choice}
  %\label{}
 %\vspace{-1mm}
%\end{figure}

%\begin{figure}[h]
  %\centering
  %\includegraphics[width=0.40\textwidth]{Figures/auc_loss.png}
  %\setlength{\belowcaptionskip}{0.0pt}
  %\caption{Performance improvement over different epochs}
  %\label{fig:performance_epochs}
  %\vspace{-1mm}
%\end{figure}

\subsection{Impact of different kernel choices}

So far we have assumed the choice of the kernel is fixed as Gaussian for all our experiments. To study the effect of kernel choices on the performance of our method, we experimented with different kernel choices from the RBF kernel family. To this end, we repeat our experiments on all 4 datasets using Gaussian (our choice of the kernel) along with Laplace and Cauchy kernels. We choose these kernels %for our experiments%
due to their positive definite properties \cite{fasshauer2011positive} which is fundamental to our MMD based kernel two-sample test (recall eqn. 4).

Figure~\ref{fig:effect_kernel} shows the results of our experiments. In general, we see a very small effect of kernel choice in the performance across all 4 datasets. Both Gaussian and Laplace kernel exhibit performance superior to Cauchy kernel. This suggests the suitability of RBF kernels (such as Gaussian kernel) as the dissimilarity metric for the proposed method.

% to save space
\iffalse
\begin{figure}[t]
  \centering
  \begin{minipage}[b]{.49\textwidth}
    %\centering
    \includegraphics[width=0.49\linewidth]{Figures/impact_window.pdf}
    %\label{fig:effect_window}
    \hfil
    \includegraphics[width=0.49\linewidth]{Figures/impact_feature.pdf}
    \caption{Effect of different window size (left) and different encoding dimension (right)}
    \label{fig:effect_dim_window}
  \end{minipage}%
  %\hfill
  \begin{minipage}[b]{.49\textwidth}
    %\centering
    \includegraphics[width=.90\linewidth]{Figures/impact_kernel.png}
    \caption{Effect of different kernel choice}
    \label{fig:effect_kernel}
  \end{minipage}%
    %\caption{}
\vspace{-4mm}
\end{figure}
\fi

\begin{figure}[t]
  \centering
  \begin{minipage}[b]{.49\textwidth}
    %\centering
    \includegraphics[width=.90\linewidth]{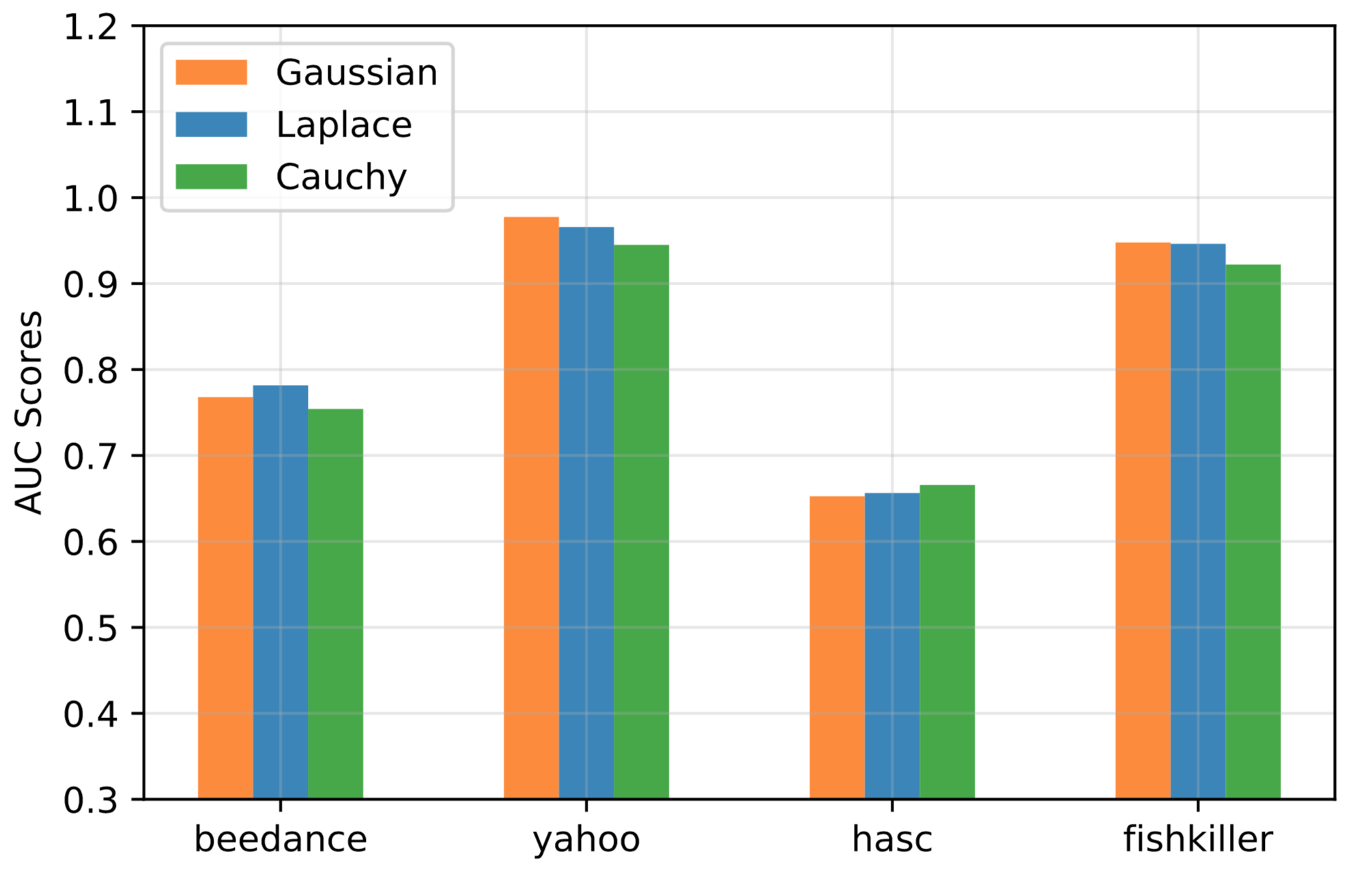}
    \caption{Effect of different kernel choice}
    \label{fig:effect_kernel}
  \end{minipage}%
  %\hfill
  \begin{minipage}[b]{.49\textwidth}
    %\centering
    \includegraphics[width=.88\linewidth]{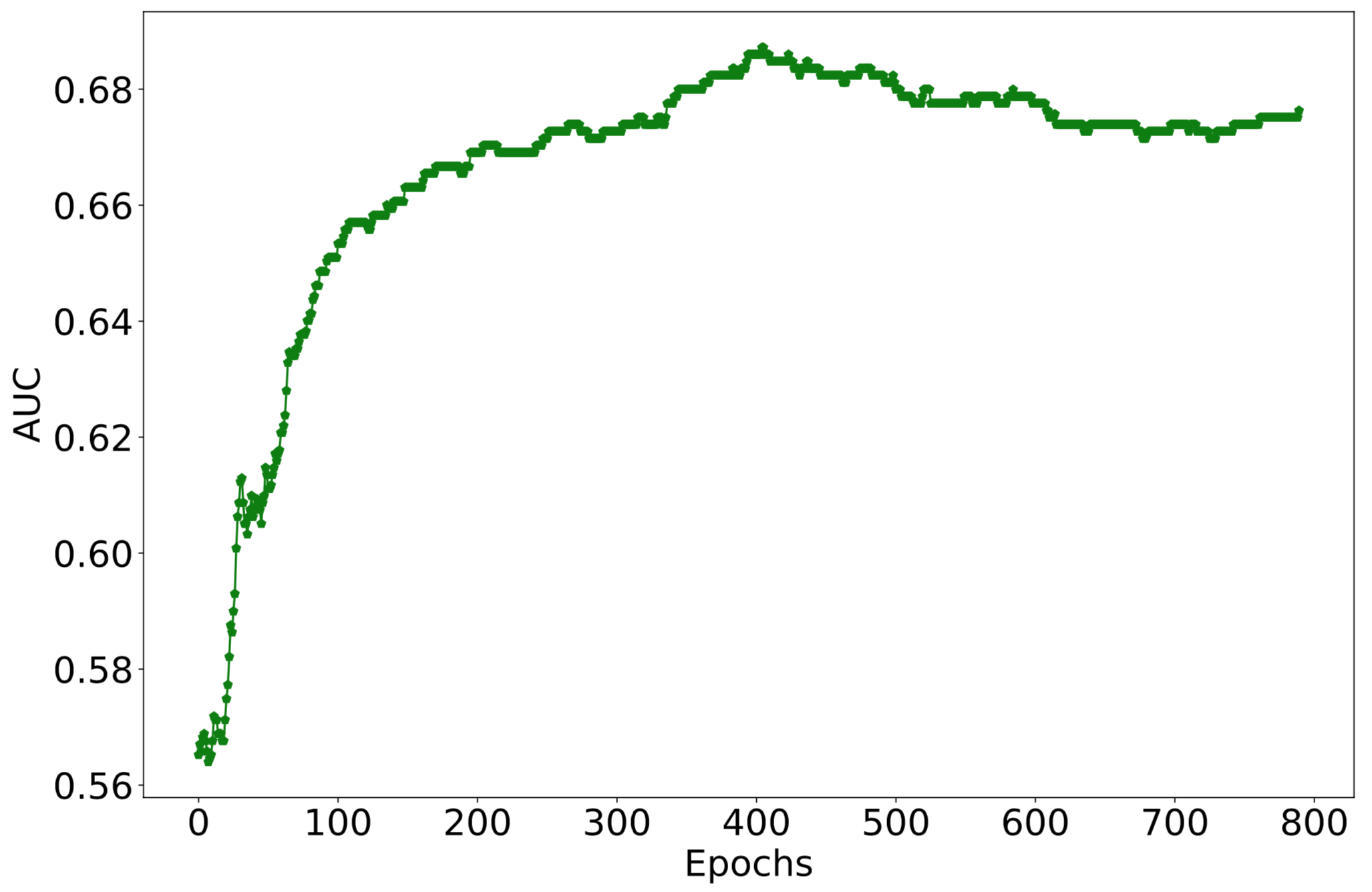}
    \caption{Performance over different epochs}
    \label{fig:performance_epochs}
  \end{minipage}%
    %\caption{}
\vspace{-2mm}
\end{figure}

\subsection{Computation cost}

Recall from our motivation for auto-encoder based approach for MMD-based kernel two-sample test was to obtain good performance without generating additional samples. %As evident from Table~\ref{tab:dataset}, all four of our datasets are class imbalanced with ~ 2\% of the data points being change points.% 
\cite{chang2018kernel} utilized a Generative Adversarial Network-based model to create additional samples for CPD problem. %for mitigating the problem of insufficient samples and optimize test power by leveraging this surplus of data.% 
However, such generative models are computationally expensive and require long training time to achieve desirable performance, which can be of limited practice for deployment.

\teal{
In Table~\ref{tab:cost}, we present a computation complexity analysis in terms of CPU utilization, memory consumption and processing time for our method. We ran our model on a Tesla K80 machine with 12GB RAM. Processing time includes the time required for training %and testing%
using the best model found via early stopping criteria. %We include the time required for processing Beedance and Yahoo datasets with KL-CPD (current baseline) for the purpose of comparison. 
We include the computational cost of KLCPD and TIRE for the purpose of comparison. For all four cases, Cadence shows the best performance in terms of CPU utilization and memory consumption (except Beedance), and processing time (except Beedance and yahoo). While TIRE also shows low CPU, memory, and time requirement because of its autoencoder-based architecture as Cadence, Cadence shows lower computation cost as it processes raw time series and has an efficient batch processing mechanism based on PyTorch. In terms of processing time, Cadence shows comparable performance to KL-CPD (in Table~\ref{tab:cost}) within under 1.5 minutes whereas for hasc and fishkiller, KLCPD cannot complete processing due to computational exhaustion (denoted by Out of Memory (OOM)). \rev{Unlike KL-CPD, a generative network, Cadence relies on the autoencoder's data compression capability in the hidden layer as a latent feature representation learning method. As a result, the model complexity and computational cost are relatively lower in comparison to other deep networks. We utilize several techniques based on the observations made in the existing literature which we believe has aided in lower computation cost while providing good performance such as the use of ReLU activation function \cite{10.5555/2968826.2968922} and unsupervised training for better generalization in smaller networks \cite{erhan2010does}.} %Note that this is promising for edge computing cases.%
Datasets representative of anomaly and fault detection problems (Yahoo and Fishkiller) require less time to process due to their subtle seasonality in variation. Beedance and HASC, which capture complex distributions of biological and human activity, require longer processing time to learn the representations and perform MMD-based two-sample test. Cadence exhibits all three properties outlined in section~\label{sec:3} as we show in Table~\ref{tab:properties}.}

%\begin{wraptable}{r}{8.5cm}
\iffalse
\begin{table}[t]
\begin{centering}
\begin{tabular}{c||c||c|}
\cline{2-3}
                 \multicolumn{1}{c|}{}        & \multicolumn{2}{|c|}{Total Processing Time (AUC)} \\  \hline
                   \multicolumn {1}{|c||}{Dataset}       & \textbf{Cadence}         & \textbf{KLCPD}        \\ \hline
\multicolumn{1}{|c||}{\textbf{Beedance}}  & \textbf{53.2s}  (0.75412) &   76.36m (0.6799)              \\ \hline
\multicolumn{1}{|c||}{\textbf{Yahoo}}  &  \textbf{36.5s} (0.97747) & 361.24m (0.96488)                     \\ \hline
\multicolumn{1}{|c||}{\textbf{HASC}}   & \textbf{93s} (0.66562)  &  -                      \\ \hline
\multicolumn{1}{|c||}{\textbf{Fishkiller}} & \textbf{ 8.34s} (0.9477) & -                     \\ \hline
\end{tabular}
\caption{Average Processing Time of Cadence on all 4 datasets. We include processing time required using KL-CPD for Beedance and Yahoo datasets, containing less than 1500 instances, and omit the remaining two due to computational exhaustion. Corresponding AUC obtained are included in parentheses.}
\label{tab:cost}
\vspace{-8mm}
\end{centering}
\end{table}
%\end{wraptable}
%{\lipsum[2] 
%\par
%Table~\ref{wrap-tab:1} is a wrapped table.}
\fi

\begin{table}[]
\begin{centering}
\begin{tabular}{l|l|llll}
\hline
\multicolumn{2}{l}{}           & \textbf{Beedance} & \textbf{yahoo} & \textbf{hasc} & \textbf{fishkiller} \\ \hline
\multirow{3}{*}{}              & KLCPD   &  59.1        &  31.2     & %44.1% 
OOM &   %38.2% 
OOM\\ \cline{2-6}
\textbf{CPU Utilization (\%)}       & TIRE    &  18.2        &  13.2     & 43.7     &   45.2        \\ \cline{2-6}
                               & Cadence &   \textbf{2.7}       &    \textbf{6.2}     & \textbf{24.8}     &  \textbf{20.3}       \\ \hline
                               \hline
\multirow{3}{*}{}              & KLCPD   &   8.3       &  6.4     &  OOM    &   OOM         \\ \cline{2-6}
\textbf{Memory Consumption (\%)}    & TIRE    &   \textbf{0.5}      &   1.8    &   2.3   &    7.2       \\ \cline{2-6}
                               & Cadence &   1.4      &   \textbf{0.7}    &   \textbf{1.5}   &    \textbf{0.8}      \\ \hline
                               \hline
\multirow{3}{*}{}              & KLCPD   &  4570.82        & 21674.40      & OOM     &   OOM         \\ \cline{2-6}
\textbf{Processing Time (seconds)}       & TIRE    &   \textbf{44.72}       &   \textbf{24.60}    &   954.62   &   470.47         \\ \cline{2-6}
                               & Cadence &   53.2       &    36.5   &    \textbf{93}  &      \textbf{8.34}      \\ \hline 
                        
\end{tabular}
\caption{\teal{Comparison of average CPU utilization, memory consumption, and processing time on all 4 datasets. We include KLCPD and TIRE in this comparison as they are the best performing methods. KLCPD's CPU utilization, memory consumption, and processing time for hasc and fishkiller datasets are omitted due to computational exhaustion (OOM: Out of Memory).}}
\label{tab:cost}
\vspace{-4mm}
\end{centering}
\end{table}

\section{Discussion and Future Work}\label{sec:10}

%In this section, we analyze different aspects of our change point detection method proposed in this paper.
%\subsection{Representation Learning for CPD} Our deep autoencoder (DAE) based CPD method, Cadence, leverages auto-encoder to learn informative representations to be used for two-sample test in the latent space. The network used in Cadence allows us to learn such latent representations, where the distance between the features is utilized as a difference measure to identify change points. Our deep network learns parameters to maximize the distance measure between the samples, which in this case is maximum mean discrepancy (MMD). 

%when data is highly non-stationary, comparing the immediate past segment and the current segment can satisfactorily measure the difference between them.% 

\textbf{Representation learning.} Training of the auto-encoder is fully unsupervised in Cadence and %with no access to training labels.%
validation performance is used as a stopping criterion. %, but do not tune parameters based on that.% 
The gradual performance improvement in Figure~\ref{fig:performance_epochs} on unseen test data as training progresses implies that the training aided towards better change point detection performance on unseen data. However, following the CPD literature and prior works on MMD, we made piece-wise \textit{i.i.d} assumption regarding the time series segments. Real-world datasets may not always follow this assumption. %Despite this assumption, one explanation for the generalizability and performance of Cadence over the baseline methods can be its simplified representation learning mechanism specifically towards change-point detection objective without making a data-specific assumption about parameter choice.%
One direction for future work is to incorporate recurrent neural networks (RNN) to capture temporal dynamics in the representation learning to overcome the assumption of \textit{i.i.d} samples. \blue{When data is highly non-stationary, using Variational Autoencoder \cite{kingma2013auto} has been explored for anomalies \cite{9053558, 9208761} and can provide leverage in better representation learning for change-point detection in future work.}
\\
\textbf{Time series partitioning.} Cadence uses MMD between each consecutive segment pairs as a change-point score to indicate the possibility of a given point being a change-point.
%While this strategy allows us to compute the change-point score for every point in a time series,% 
However, a threshold needs to be set on MMD values to determine the location of change points in the sensor stream. In this work, we focus on finding a method that is generalizable and well-performing across various domains and Cadence shows that property using the same threshold (40\% of the maximum change point score) on all the experiment datasets. However, change-point detection performance will be sensitive to the choice of threshold and further exploration for adaptive threshold choice is needed in future work. %However, extreme cases can occur in the real-world such as data being very high dimensional (video), missing data (packet drop), etc. Finding an appropriate threshold and window can be desirable in such cases and depends on the type of data and the number of dimensions. 
%Another direction for future work in this line can be partitioning the entire time series into segments based on different ranges of MMD scores (or change point scores), where segments with similar MMD scores can be grouped as similar events or classes. \rev{For practical use case, change point models can be trained on historical data and can use transfer learning \cite{zhuang2020comprehensive} approach to adapt trained models to newly appearing data (e.g. new class of activities in indoor human activity detection), to ensure being well-performing and computationally inexpensive for massive-scale use case. This can be another direction for future work.}  
\\
\rev{\textbf{Generalizability and Scale.} %The computation cost of Cadence for training is lower due to its simpler architecture and design choices.%
Our model is developed and used in two separate phases: development (training) and inference. The training time for Cadence is dependent on the amount of data available for training. Training from scratch may require a large dataset and increased training time.  However, %and is low due to its simpler architecture and design choices.%
this can be reduced with transfer learning~\cite{zhuang2020comprehensive, hpc} where a previously trained model can be trained with much fewer iterations using new data and pre-trained weights. %This will reduce the training time as observed in other transfer learning-based approaches \cite{zhuang2020comprehensive, hpc}%.
In the inference phase, the trained model can be applied to
streaming data in batches %(the size of the input to the model)%
and can thus scale to large data. The inference-generation throughput may be a bottleneck for large-scale streaming applications.  Cadence's inference time is $\sim{30}$ milliseconds (KLCPD: $\sim{4.45}$ seconds, TIRE: $\sim{225}$ milliseconds) which should cover a large class of IoT applications. However, we leave related scaling challenges for massive streaming applications for future work.}

\begin{table*}[t]
\centering
%\color{red}
\begin{tabular}{c|c|c|c}
\hline
 \textbf{Algorithms} & \textbf{Ease of use} & \textbf{Computational Efficiency} & \textbf{Performance}\\ \hline
RDR-CPD \cite{liu2013change} & \cmark & \xmark & \xmark \\ \hline
Mstats-CPD \cite{XieYchange} & \xmark & \cmark & \xmark\\ \hline
LSTNet \cite{Liu2018ClassifierTS} &\cmark & \xmark & \xmark\\ \hline
TIRE\cite{deryck2021change} &  \cmark &  \cmark & \xmark\\ \hline
KLCPD \cite{chang2018kernel} & \xmark & \xmark & \cmark\\ \hline
Cadence (Proposed Approach) & \cmark & \cmark & \cmark\\ \hline

\end{tabular}
\caption{\purple{Overview of properties exhibited by existing CPD algorithms. We design our proposed approach, \textit{Cadence}, to exhibit all three properties.}}
\label{tab:properties}
\vspace{-4mm}
\end{table*}

\section{Conclusion}\label{sec:12}

This paper presents Cadence - a framework that projects a set of time series samples in a jointly optimized feature space towards change-point detection. It works by iteratively optimizing a custom representation learning objective based on the discrepancy between time-series samples. Using this novel feature mapping, our approach provides a way to detect time-series events without using change-point labels provided by a human. We evaluate our proposed algorithm, Cadence, with public change point detection datasets and present its performance for different parameter settings in the paper.
We compared Cadence's performance with the state-of-the-art techniques and showed its superior performance as well as robustness to hyperparameter tuning and sample efficiency. %Cadence provides such performance utilizing available data or a subset of it.%
Its good performance is obtained within surprisingly less time, with a simpler, easy to train, and generalizable model. With these properties, Cadence has the potential to scale well for %applications with %large, complex data streams towards%
practical deployment in applications of general-purpose sensing and continuous learning without human input. %algorithm for time-series partitioning.

%Empirical studies presented in this paper demonstrate the strength of our proposed method.%

% put two figures in single column side by side
%\begin{figure}[h]
    %\centering
    %\caption{Two generator partitions for two properties} 
    
    %\includegraphics[width=0.45\linewidth]{Figures/figure 4.jpg}
    %\hfil
    %\includegraphics[width=0.45\linewidth]{Figures/figure 5.jpg}  
%\end{figure}

\bibliographystyle{ACM-Reference-Format}
\bibliography{sample-sigconf}


\begin{thebibliography}{101}

%%% the Web address.

\ifx \showCODEN    \undefined \def \showCODEN     #1{\unskip}     \fi
\ifx \showDOI      \undefined \def \showDOI       #1{#1}\fi
\ifx \showISBNx    \undefined \def \showISBNx     #1{\unskip}     \fi
\ifx \showISBNxiii \undefined \def \showISBNxiii  #1{\unskip}     \fi
\ifx \showISSN     \undefined \def \showISSN      #1{\unskip}     \fi
\ifx \showLCCN     \undefined \def \showLCCN      #1{\unskip}     \fi
\ifx \shownote     \undefined \def \shownote      #1{#1}          \fi
\ifx \showarticletitle \undefined \def \showarticletitle #1{#1}   \fi
\ifx \showURL      \undefined \def \showURL       {\relax}        \fi
% invisible to TeX
\providecommand\bibfield[2]{#2}
\providecommand\bibinfo[2]{#2}
\providecommand\natexlab[1]{#1}
\providecommand\showeprint[2][]{arXiv:#2}

\bibitem[\protect\citeauthoryear{??}{bee}{2008}]%
        {beedance}
 \bibinfo{year}{2008}\natexlab{}.
\newblock \bibinfo{title}{Honey Bee Waggle Data}.
\newblock
\newblock
\urldef\tempurl%
\url{https://www.cc.gatech.edu/~borg/ijcv_psslds/}
\showURL{%
\tempurl}


\bibitem[\protect\citeauthoryear{??}{has}{2011}]%
        {hasc}
 \bibinfo{year}{2011}\natexlab{}.
\newblock \bibinfo{title}{HASC 2011 Data}.
\newblock
\newblock
\urldef\tempurl%
\url{http://hasc.jp/hc2011/}
\showURL{%
\tempurl}


\bibitem[\protect\citeauthoryear{??}{yah}{2017}]%
        {yahoo}
 \bibinfo{year}{2017}\natexlab{}.
\newblock \bibinfo{title}{Yahoo Web Data}.
\newblock
\newblock
\urldef\tempurl%
\url{https://webscope.sandbox.yahoo.com}
\showURL{%
\tempurl}


\bibitem[\protect\citeauthoryear{Abedin, Ehsanpour, Shi, Rezatofighi, and
  Ranasinghe}{Abedin et~al\mbox{.}}{2021}]%
        {10.1145/3448083}
\bibfield{author}{\bibinfo{person}{Alireza Abedin}, \bibinfo{person}{Mahsa
  Ehsanpour}, \bibinfo{person}{Qinfeng Shi}, \bibinfo{person}{Hamid
  Rezatofighi}, {and} \bibinfo{person}{Damith~C. Ranasinghe}.}
  \bibinfo{year}{2021}\natexlab{}.
\newblock \showarticletitle{Attend and Discriminate: Beyond the
  State-of-the-Art for Human Activity Recognition Using Wearable Sensors}.
\newblock \bibinfo{journal}{\emph{Proc. ACM Interact. Mob. Wearable Ubiquitous
  Technol.}} \bibinfo{volume}{5}, \bibinfo{number}{1}, Article
  \bibinfo{articleno}{1} (\bibinfo{date}{March} \bibinfo{year}{2021}),
  \bibinfo{numpages}{22}~pages.
\newblock
\urldef\tempurl%
\url{https://doi.org/10.1145/3448083}
\showDOI{\tempurl}


\bibitem[\protect\citeauthoryear{Adams and Mackay}{Adams and Mackay}{2007}]%
        {Adams2007BayesianOC}
\bibfield{author}{\bibinfo{person}{Ryan~P. Adams} {and}
  \bibinfo{person}{David~J.C. Mackay}.} \bibinfo{year}{2007}\natexlab{}.
\newblock \showarticletitle{Bayesian Online Changepoint Detection}.
\newblock


\bibitem[\protect\citeauthoryear{Adams and MacKay}{Adams and MacKay}{2007}]%
        {adams2007bayesian}
\bibfield{author}{\bibinfo{person}{Ryan~Prescott Adams} {and}
  \bibinfo{person}{David~JC MacKay}.} \bibinfo{year}{2007}\natexlab{}.
\newblock \showarticletitle{Bayesian online changepoint detection}.
\newblock \bibinfo{journal}{\emph{arXiv preprint arXiv:0710.3742}}
  (\bibinfo{year}{2007}).
\newblock


\bibitem[\protect\citeauthoryear{Aminikhanghahi and Cook}{Aminikhanghahi and
  Cook}{2017}]%
        {Aminikhanghahi2017}
\bibfield{author}{\bibinfo{person}{S. Aminikhanghahi} {and}
  \bibinfo{person}{D.~J. Cook}.} \bibinfo{year}{2017}\natexlab{}.
\newblock \showarticletitle{A Survey of Methods for Time Series Change Point
  Detection}.
\newblock \bibinfo{journal}{\emph{Knowledge and information systems}}
  \bibinfo{volume}{51}, \bibinfo{number}{2} (\bibinfo{date}{May}
  \bibinfo{year}{2017}), \bibinfo{pages}{339--367}.
\newblock


\bibitem[\protect\citeauthoryear{{Aminikhanghahi} and {Cook}}{{Aminikhanghahi}
  and {Cook}}{2017}]%
        {Cook17}
\bibfield{author}{\bibinfo{person}{S. {Aminikhanghahi}} {and}
  \bibinfo{person}{D.~J. {Cook}}.} \bibinfo{year}{2017}\natexlab{}.
\newblock \showarticletitle{Using change point detection to automate daily
  activity segmentation}. In \bibinfo{booktitle}{\emph{2017 IEEE International
  Conference on Pervasive Computing and Communications Workshops}}.
  \bibinfo{pages}{262--267}.
\newblock


\bibitem[\protect\citeauthoryear{Aminikhanghahi and Cook}{Aminikhanghahi and
  Cook}{2017}]%
        {aminikhanghahi2017using}
\bibfield{author}{\bibinfo{person}{Samaneh Aminikhanghahi} {and}
  \bibinfo{person}{Diane~J Cook}.} \bibinfo{year}{2017}\natexlab{}.
\newblock \showarticletitle{Using change point detection to automate daily
  activity segmentation}. In \bibinfo{booktitle}{\emph{2017 IEEE International
  Conference on Pervasive Computing and Communications Workshops (PerCom
  Workshops)}}. IEEE, \bibinfo{pages}{262--267}.
\newblock


\bibitem[\protect\citeauthoryear{Aminikhanghahi, Wang, and Cook}{Aminikhanghahi
  et~al\mbox{.}}{2018}]%
        {aminikhanghahi2018real}
\bibfield{author}{\bibinfo{person}{Samaneh Aminikhanghahi},
  \bibinfo{person}{Tinghui Wang}, {and} \bibinfo{person}{Diane~J Cook}.}
  \bibinfo{year}{2018}\natexlab{}.
\newblock \showarticletitle{Real-time change point detection with application
  to smart home time series data}.
\newblock \bibinfo{journal}{\emph{IEEE Transactions on Knowledge and Data
  Engineering}} \bibinfo{volume}{31}, \bibinfo{number}{5}
  (\bibinfo{year}{2018}), \bibinfo{pages}{1010--1023}.
\newblock


\bibitem[\protect\citeauthoryear{Ang, Dilys~Salim, and Hamilton}{Ang
  et~al\mbox{.}}{2016}]%
        {ANG}
\bibfield{author}{\bibinfo{person}{Irvan Bastian~Arief Ang},
  \bibinfo{person}{Flora Dilys~Salim}, {and} \bibinfo{person}{Margaret
  Hamilton}.} \bibinfo{year}{2016}\natexlab{}.
\newblock \showarticletitle{Human occupancy recognition with multivariate
  ambient sensors}. In \bibinfo{booktitle}{\emph{2016 IEEE International
  Conference on Pervasive Computing and Communication Workshops (PerCom
  Workshops)}}. \bibinfo{pages}{1--6}.
\newblock
\urldef\tempurl%
\url{https://doi.org/10.1109/PERCOMW.2016.7457116}
\showDOI{\tempurl}


\bibitem[\protect\citeauthoryear{Atzori, Iera, and Morabito}{Atzori
  et~al\mbox{.}}{2010}]%
        {IoT00}
\bibfield{author}{\bibinfo{person}{L. Atzori}, \bibinfo{person}{A. Iera}, {and}
  \bibinfo{person}{G. Morabito}.} \bibinfo{year}{2010}\natexlab{}.
\newblock \showarticletitle{The {Internet of Things}: A survey}.
\newblock \bibinfo{journal}{\emph{Computer Networks}} \bibinfo{volume}{54},
  \bibinfo{number}{15} (\bibinfo{date}{Oct.} \bibinfo{year}{2010}),
  \bibinfo{pages}{2787--2805}.
\newblock


\bibitem[\protect\citeauthoryear{Bai, Yao, Wang, Kanhere, Guo, and Yu}{Bai
  et~al\mbox{.}}{2020}]%
        {10.1145/3397323}
\bibfield{author}{\bibinfo{person}{Lei Bai}, \bibinfo{person}{Lina Yao},
  \bibinfo{person}{Xianzhi Wang}, \bibinfo{person}{Salil~S. Kanhere},
  \bibinfo{person}{Bin Guo}, {and} \bibinfo{person}{Zhiwen Yu}.}
  \bibinfo{year}{2020}\natexlab{}.
\newblock \showarticletitle{Adversarial Multi-View Networks for Activity
  Recognition}.
\newblock \bibinfo{journal}{\emph{Proc. ACM Interact. Mob. Wearable Ubiquitous
  Technol.}} \bibinfo{volume}{4}, \bibinfo{number}{2}, Article
  \bibinfo{articleno}{42} (\bibinfo{date}{June} \bibinfo{year}{2020}),
  \bibinfo{numpages}{22}~pages.
\newblock
\urldef\tempurl%
\url{https://doi.org/10.1145/3397323}
\showDOI{\tempurl}


\bibitem[\protect\citeauthoryear{Banerjee and Guhathakurta}{Banerjee and
  Guhathakurta}{2020}]%
        {Banerjee_2020}
\bibfield{author}{\bibinfo{person}{Sayantan Banerjee} {and}
  \bibinfo{person}{Kousik Guhathakurta}.} \bibinfo{year}{2020}\natexlab{}.
\newblock \showarticletitle{Change‐point analysis in financial networks}.
\newblock \bibinfo{journal}{\emph{Stat}} \bibinfo{volume}{9},
  \bibinfo{number}{1} (\bibinfo{date}{Jan} \bibinfo{year}{2020}).
\newblock
\showISSN{2049-1573}


\bibitem[\protect\citeauthoryear{Basseville and Nikiforov}{Basseville and
  Nikiforov}{1993}]%
        {Basseville93}
\bibfield{author}{\bibinfo{person}{Mich\`{e}le Basseville} {and}
  \bibinfo{person}{Igor~V. Nikiforov}.} \bibinfo{year}{1993}\natexlab{}.
\newblock \bibinfo{booktitle}{\emph{Detection of Abrupt Changes: Theory and
  Application}}.
\newblock \bibinfo{publisher}{Prentice-Hall, Inc.}, \bibinfo{address}{Upper
  Saddle River, NJ, USA}.
\newblock
\showISBNx{0-13-126780-9}


\bibitem[\protect\citeauthoryear{Bonacina, Miele, and Corsini}{Bonacina
  et~al\mbox{.}}{2020}]%
        {modelling1010001}
\bibfield{author}{\bibinfo{person}{Fabrizio Bonacina},
  \bibinfo{person}{Eric~Stefan Miele}, {and} \bibinfo{person}{Alessandro
  Corsini}.} \bibinfo{year}{2020}\natexlab{}.
\newblock \showarticletitle{Time Series Clustering: A Complex Network-Based
  Approach for Feature Selection in Multi-Sensor Data}.
\newblock \bibinfo{journal}{\emph{Modelling}} \bibinfo{volume}{1},
  \bibinfo{number}{1} (\bibinfo{year}{2020}), \bibinfo{pages}{1--21}.
\newblock
\showISSN{2673-3951}
\urldef\tempurl%
\url{https://doi.org/10.3390/modelling1010001}
\showDOI{\tempurl}


\bibitem[\protect\citeauthoryear{Chang, Li, Yang, and Póczos}{Chang
  et~al\mbox{.}}{2019}]%
        {chang2018kernel}
\bibfield{author}{\bibinfo{person}{Wei-Cheng Chang},
  \bibinfo{person}{Chun-Liang Li}, \bibinfo{person}{Yiming Yang}, {and}
  \bibinfo{person}{Barnabás Póczos}.} \bibinfo{year}{2019}\natexlab{}.
\newblock \showarticletitle{Kernel Change-point Detection with Auxiliary Deep
  Generative Models}. In \bibinfo{booktitle}{\emph{International Conference on
  Learning Representations}}.
\newblock


\bibitem[\protect\citeauthoryear{Chowdhury, Selouani, and
  O'Shaughnessy}{Chowdhury et~al\mbox{.}}{2012}]%
        {Chowdhury2012AHN}
\bibfield{author}{\bibinfo{person}{Md~Foezur~Rahman Chowdhury},
  \bibinfo{person}{Sid-Ahmed Selouani}, {and} \bibinfo{person}{Douglas~D.
  O'Shaughnessy}.} \bibinfo{year}{2012}\natexlab{}.
\newblock \showarticletitle{A highly non-stationary noise tracking and
  compensation algorithm, with applications to speech enhancement and on-line
  ASR}.
\newblock \bibinfo{journal}{\emph{2012 IEEE International Conference on
  Acoustics, Speech and Signal Processing (ICASSP)}} (\bibinfo{year}{2012}),
  \bibinfo{pages}{4337--4340}.
\newblock


\bibitem[\protect\citeauthoryear{Chung, Kastner, Dinh, Goel, Courville, and
  Bengio}{Chung et~al\mbox{.}}{2015}]%
        {chung2015recurrent}
\bibfield{author}{\bibinfo{person}{Junyoung Chung}, \bibinfo{person}{Kyle
  Kastner}, \bibinfo{person}{Laurent Dinh}, \bibinfo{person}{Kratarth Goel},
  \bibinfo{person}{Aaron Courville}, {and} \bibinfo{person}{Yoshua Bengio}.}
  \bibinfo{year}{2015}\natexlab{}.
\newblock \showarticletitle{A recurrent latent variable model for sequential
  data}.
\newblock \bibinfo{journal}{\emph{arXiv preprint arXiv:1506.02216}}
  (\bibinfo{year}{2015}).
\newblock


\bibitem[\protect\citeauthoryear{Davis and Goadrich}{Davis and
  Goadrich}{2006}]%
        {davis2006relationship}
\bibfield{author}{\bibinfo{person}{Jesse Davis} {and} \bibinfo{person}{Mark
  Goadrich}.} \bibinfo{year}{2006}\natexlab{}.
\newblock \showarticletitle{The relationship between Precision-Recall and ROC
  curves}. In \bibinfo{booktitle}{\emph{Proceedings of the 23rd international
  conference on Machine learning}}. \bibinfo{pages}{233--240}.
\newblock


\bibitem[\protect\citeauthoryear{De~Ryck, De~Vos, and Bertrand}{De~Ryck
  et~al\mbox{.}}{2020}]%
        {de2020change}
\bibfield{author}{\bibinfo{person}{Tim De~Ryck}, \bibinfo{person}{Maarten
  De~Vos}, {and} \bibinfo{person}{Alexander Bertrand}.}
  \bibinfo{year}{2020}\natexlab{}.
\newblock \showarticletitle{Change Point Detection in Time Series Data using
  Autoencoders with a Time-Invariant Representation}.
\newblock \bibinfo{journal}{\emph{arXiv preprint arXiv:2008.09524}}
  (\bibinfo{year}{2020}).
\newblock


\bibitem[\protect\citeauthoryear{Erhan, Courville, Bengio, and Vincent}{Erhan
  et~al\mbox{.}}{2010}]%
        {erhan2010does}
\bibfield{author}{\bibinfo{person}{Dumitru Erhan}, \bibinfo{person}{Aaron
  Courville}, \bibinfo{person}{Yoshua Bengio}, {and} \bibinfo{person}{Pascal
  Vincent}.} \bibinfo{year}{2010}\natexlab{}.
\newblock \showarticletitle{Why does unsupervised pre-training help deep
  learning?}. In \bibinfo{booktitle}{\emph{Proceedings of the thirteenth
  international conference on artificial intelligence and statistics}}. JMLR
  Workshop and Conference Proceedings, \bibinfo{pages}{201--208}.
\newblock


\bibitem[\protect\citeauthoryear{Fasshauer}{Fasshauer}{2011}]%
        {fasshauer2011positive}
\bibfield{author}{\bibinfo{person}{G.~E. Fasshauer}.}
  \bibinfo{year}{2011}\natexlab{}.
\newblock \showarticletitle{Positive definite kernels: past, present and
  future}.
\newblock  (\bibinfo{year}{2011}).
\newblock


\bibitem[\protect\citeauthoryear{Feng, Mehmani, and Zhang}{Feng
  et~al\mbox{.}}{2020}]%
        {Feng}
\bibfield{author}{\bibinfo{person}{Cong Feng}, \bibinfo{person}{Ali Mehmani},
  {and} \bibinfo{person}{Jie Zhang}.} \bibinfo{year}{2020}\natexlab{}.
\newblock \showarticletitle{Deep Learning-Based Real-Time Building Occupancy
  Detection Using AMI Data}.
\newblock \bibinfo{journal}{\emph{IEEE Transactions on Smart Grid}}
  \bibinfo{volume}{11}, \bibinfo{number}{5} (\bibinfo{year}{2020}),
  \bibinfo{pages}{4490--4501}.
\newblock
\urldef\tempurl%
\url{https://doi.org/10.1109/TSG.2020.2982351}
\showDOI{\tempurl}


\bibitem[\protect\citeauthoryear{Greenwald and Oertel}{Greenwald and
  Oertel}{2017}]%
        {ML00}
\bibfield{author}{\bibinfo{person}{Hal~S. Greenwald} {and}
  \bibinfo{person}{Carsten~K. Oertel}.} \bibinfo{year}{2017}\natexlab{}.
\newblock \showarticletitle{Future Directions in Machine Learning}.
\newblock \bibinfo{journal}{\emph{Frontiers in Robotics and AI}}
  \bibinfo{volume}{3} (\bibinfo{year}{2017}), \bibinfo{pages}{79}.
\newblock
\showISSN{2296-9144}


\bibitem[\protect\citeauthoryear{Gretton, Borgwardt, Rasch, Sch{\"o}lkopf, and
  Smola}{Gretton et~al\mbox{.}}{2012a}]%
        {gretton2012kernel}
\bibfield{author}{\bibinfo{person}{Arthur Gretton}, \bibinfo{person}{Karsten~M
  Borgwardt}, \bibinfo{person}{Malte~J Rasch}, \bibinfo{person}{Bernhard
  Sch{\"o}lkopf}, {and} \bibinfo{person}{Alexander Smola}.}
  \bibinfo{year}{2012}\natexlab{a}.
\newblock \showarticletitle{A kernel two-sample test}.
\newblock \bibinfo{journal}{\emph{Journal of Machine Learning Research}}
  \bibinfo{volume}{13}, \bibinfo{number}{Mar} (\bibinfo{year}{2012}),
  \bibinfo{pages}{723--773}.
\newblock


\bibitem[\protect\citeauthoryear{Gretton, Fukumizu, Harchaoui, and
  Sriperumbudur}{Gretton et~al\mbox{.}}{2009}]%
        {Gretton2009}
\bibfield{author}{\bibinfo{person}{Arthur Gretton}, \bibinfo{person}{Kenji
  Fukumizu}, \bibinfo{person}{Za\"{\i}d Harchaoui}, {and}
  \bibinfo{person}{Bharath~K. Sriperumbudur}.} \bibinfo{year}{2009}\natexlab{}.
\newblock \showarticletitle{A Fast, Consistent Kernel Two-Sample Test}.
\newblock In \bibinfo{booktitle}{\emph{Advances in Neural Information
  Processing Systems 22}}. \bibinfo{pages}{673--681}.
\newblock


\bibitem[\protect\citeauthoryear{Gretton, Sejdinovic, Strathmann, Balakrishnan,
  Pontil, Fukumizu, and Sriperumbudur}{Gretton et~al\mbox{.}}{2012b}]%
        {gretton2012optimal}
\bibfield{author}{\bibinfo{person}{A. Gretton}, \bibinfo{person}{D.
  Sejdinovic}, \bibinfo{person}{H. Strathmann}, \bibinfo{person}{S.
  Balakrishnan}, \bibinfo{person}{M. Pontil}, \bibinfo{person}{K. Fukumizu},
  {and} \bibinfo{person}{B.~K Sriperumbudur}.}
  \bibinfo{year}{2012}\natexlab{b}.
\newblock \showarticletitle{Optimal kernel choice for large-scale two-sample
  tests}. In \bibinfo{booktitle}{\emph{Advances in neural information
  processing systems}}. \bibinfo{pages}{1205--1213}.
\newblock


\bibitem[\protect\citeauthoryear{Gubbia, Buyya, Marusic, and
  Palaniswami}{Gubbia et~al\mbox{.}}{2013}]%
        {AGG04}
\bibfield{author}{\bibinfo{person}{J. Gubbia}, \bibinfo{person}{R. Buyya},
  \bibinfo{person}{S. Marusic}, {and} \bibinfo{person}{M. Palaniswami}.}
  \bibinfo{year}{2013}\natexlab{}.
\newblock \showarticletitle{{Internet of Things (IoT)}: A vision, architectural
  elements, and future directions}.
\newblock \bibinfo{journal}{\emph{Future Generation Computer Systems}}
  \bibinfo{volume}{29}, \bibinfo{number}{7} (\bibinfo{date}{Sept.}
  \bibinfo{year}{2013}), \bibinfo{pages}{1645--1660}.
\newblock


\bibitem[\protect\citeauthoryear{Guo, Liu, Shi, Liu, Chen, and Chuah}{Guo
  et~al\mbox{.}}{2018}]%
        {guo2018device}
\bibfield{author}{\bibinfo{person}{Xiaonan Guo}, \bibinfo{person}{Jian Liu},
  \bibinfo{person}{Cong Shi}, \bibinfo{person}{Hongbo Liu},
  \bibinfo{person}{Yingying Chen}, {and} \bibinfo{person}{Mooi~Choo Chuah}.}
  \bibinfo{year}{2018}\natexlab{}.
\newblock \showarticletitle{Device-free personalized fitness assistant using
  WiFi}.
\newblock \bibinfo{journal}{\emph{Proceedings of the ACM on Interactive,
  Mobile, Wearable and Ubiquitous Technologies}} \bibinfo{volume}{2},
  \bibinfo{number}{4} (\bibinfo{year}{2018}), \bibinfo{pages}{1--23}.
\newblock


\bibitem[\protect\citeauthoryear{Guo, Ji, Wang, Yu, Min, and Li}{Guo
  et~al\mbox{.}}{2020}]%
        {9208761}
\bibfield{author}{\bibinfo{person}{Yifan Guo}, \bibinfo{person}{Tianxi Ji},
  \bibinfo{person}{Qianlong Wang}, \bibinfo{person}{Lixing Yu},
  \bibinfo{person}{Geyong Min}, {and} \bibinfo{person}{Pan Li}.}
  \bibinfo{year}{2020}\natexlab{}.
\newblock \showarticletitle{Unsupervised Anomaly Detection in IoT Systems for
  Smart Cities}.
\newblock \bibinfo{journal}{\emph{IEEE Transactions on Network Science and
  Engineering}} \bibinfo{volume}{7}, \bibinfo{number}{4}
  (\bibinfo{year}{2020}), \bibinfo{pages}{2231--2242}.
\newblock
\urldef\tempurl%
\url{https://doi.org/10.1109/TNSE.2020.3027543}
\showDOI{\tempurl}


\bibitem[\protect\citeauthoryear{{Gustafsson}}{{Gustafsson}}{1996}]%
        {Gustafsson}
\bibfield{author}{\bibinfo{person}{F. {Gustafsson}}.}
  \bibinfo{year}{1996}\natexlab{}.
\newblock \showarticletitle{The marginalized likelihood ratio test for
  detecting abrupt changes}.
\newblock \bibinfo{journal}{\emph{IEEE Trans. Automat. Control}}
  \bibinfo{volume}{41}, \bibinfo{number}{1} (\bibinfo{year}{1996}),
  \bibinfo{pages}{66--78}.
\newblock


\bibitem[\protect\citeauthoryear{Harchaoui, Bach, and Moulines}{Harchaoui
  et~al\mbox{.}}{2008}]%
        {Harchaoui08}
\bibfield{author}{\bibinfo{person}{Za\"{\i}d Harchaoui},
  \bibinfo{person}{Francis Bach}, {and} \bibinfo{person}{\'{E}ric Moulines}.}
  \bibinfo{year}{2008}\natexlab{}.
\newblock \showarticletitle{Kernel Change-Point Analysis}. In
  \bibinfo{booktitle}{\emph{21st International Conference on Neural Information
  Processing Systems}} \emph{(\bibinfo{series}{NIPS’08})}.
  \bibinfo{pages}{609–616}.
\newblock


\bibitem[\protect\citeauthoryear{H{\"a}rdle, Werwatz, M{\"u}ller, and
  Sperlich}{H{\"a}rdle et~al\mbox{.}}{2004}]%
        {hardle2004nonparametric}
\bibfield{author}{\bibinfo{person}{W. H{\"a}rdle}, \bibinfo{person}{A.
  Werwatz}, \bibinfo{person}{M. M{\"u}ller}, {and} \bibinfo{person}{S.
  Sperlich}.} \bibinfo{year}{2004}\natexlab{}.
\newblock \showarticletitle{Nonparametric density estimation}.
\newblock In \bibinfo{booktitle}{\emph{Nonparametric and semiparametric
  models}}. \bibinfo{pages}{39--83}.
\newblock


\bibitem[\protect\citeauthoryear{He and Garcia}{He and Garcia}{2009}]%
        {he2009learning}
\bibfield{author}{\bibinfo{person}{Haibo He} {and} \bibinfo{person}{Edwardo~A
  Garcia}.} \bibinfo{year}{2009}\natexlab{}.
\newblock \showarticletitle{Learning from imbalanced data}.
\newblock \bibinfo{journal}{\emph{IEEE Transactions on knowledge and data
  engineering}} \bibinfo{volume}{21}, \bibinfo{number}{9}
  (\bibinfo{year}{2009}), \bibinfo{pages}{1263--1284}.
\newblock


\bibitem[\protect\citeauthoryear{He, Zhang, Ren, and Sun}{He
  et~al\mbox{.}}{2015}]%
        {he2015delving}
\bibfield{author}{\bibinfo{person}{Kaiming He}, \bibinfo{person}{Xiangyu
  Zhang}, \bibinfo{person}{Shaoqing Ren}, {and} \bibinfo{person}{Jian Sun}.}
  \bibinfo{year}{2015}\natexlab{}.
\newblock \showarticletitle{Delving deep into rectifiers: Surpassing
  human-level performance on imagenet classification}. In
  \bibinfo{booktitle}{\emph{Proceedings of the IEEE international conference on
  computer vision}}. \bibinfo{pages}{1026--1034}.
\newblock


\bibitem[\protect\citeauthoryear{Hinton and Salakhutdinov}{Hinton and
  Salakhutdinov}{2006}]%
        {hinton2006reducing}
\bibfield{author}{\bibinfo{person}{Geoffrey~E Hinton} {and}
  \bibinfo{person}{Ruslan~R Salakhutdinov}.} \bibinfo{year}{2006}\natexlab{}.
\newblock \showarticletitle{Reducing the dimensionality of data with neural
  networks}.
\newblock \bibinfo{journal}{\emph{science}} \bibinfo{volume}{313},
  \bibinfo{number}{5786} (\bibinfo{year}{2006}), \bibinfo{pages}{504--507}.
\newblock


\bibitem[\protect\citeauthoryear{Huang, Wang, Zhao, and Zhang}{Huang
  et~al\mbox{.}}{2019}]%
        {10.1145/3328919}
\bibfield{author}{\bibinfo{person}{Anna Huang}, \bibinfo{person}{Dong Wang},
  \bibinfo{person}{Run Zhao}, {and} \bibinfo{person}{Qian Zhang}.}
  \bibinfo{year}{2019}\natexlab{}.
\newblock \showarticletitle{Au-Id: Automatic User Identification and
  Authentication through the Motions Captured from Sequential Human Activities
  Using RFID}.
\newblock \bibinfo{journal}{\emph{Proc. ACM Interact. Mob. Wearable Ubiquitous
  Technol.}} \bibinfo{volume}{3}, \bibinfo{number}{2}, Article
  \bibinfo{articleno}{48} (\bibinfo{date}{June} \bibinfo{year}{2019}),
  \bibinfo{numpages}{26}~pages.
\newblock
\urldef\tempurl%
\url{https://doi.org/10.1145/3328919}
\showDOI{\tempurl}


\bibitem[\protect\citeauthoryear{Huynh, Balan, Ko, and Lee}{Huynh
  et~al\mbox{.}}{2019}]%
        {Sinh2019}
\bibfield{author}{\bibinfo{person}{Sinh Huynh}, \bibinfo{person}{Rajesh~Krishna
  Balan}, \bibinfo{person}{JeongGil Ko}, {and} \bibinfo{person}{Youngki Lee}.}
  \bibinfo{year}{2019}\natexlab{}.
\newblock \showarticletitle{VitaMon: Measuring Heart Rate Variability Using
  Smartphone Front Camera}. In \bibinfo{booktitle}{\emph{17th Conference on
  Embedded Networked Sensor Systems}} \emph{(\bibinfo{series}{SenSys ’19})}.
  \bibinfo{pages}{1–14}.
\newblock


\bibitem[\protect\citeauthoryear{Id{\'{e}} and Tsuda}{Id{\'{e}} and
  Tsuda}{2007}]%
        {IdeT07}
\bibfield{author}{\bibinfo{person}{Tsuyoshi Id{\'{e}}} {and}
  \bibinfo{person}{Koji Tsuda}.} \bibinfo{year}{2007}\natexlab{}.
\newblock \showarticletitle{Change-Point Detection using Krylov Subspace
  Learning}. In \bibinfo{booktitle}{\emph{Proceedings of the Seventh {SIAM}
  International Conference on Data Mining, April 26-28, 2007, Minneapolis,
  Minnesota, {USA}}}. \bibinfo{pages}{515--520}.
\newblock


\bibitem[\protect\citeauthoryear{Janardan and Mehta}{Janardan and
  Mehta}{2017}]%
        {JANARDAN2017804}
\bibfield{author}{\bibinfo{person}{Janardan} {and} \bibinfo{person}{Shikha
  Mehta}.} \bibinfo{year}{2017}\natexlab{}.
\newblock \showarticletitle{Concept drift in Streaming Data Classification:
  Algorithms, Platforms and Issues}.
\newblock   \bibinfo{volume}{122} (\bibinfo{year}{2017}), \bibinfo{pages}{804
  -- 811}.
\newblock
\newblock
\shownote{5th International Conference on Information Technology and
  Quantitative Management, ITQM 2017.}


\bibitem[\protect\citeauthoryear{Jeyakumar, Lai, Suda, and
  Srivastava}{Jeyakumar et~al\mbox{.}}{2019}]%
        {sensehar2019}
\bibfield{author}{\bibinfo{person}{Jeya~Vikranth Jeyakumar},
  \bibinfo{person}{Liangzhen Lai}, \bibinfo{person}{Naveen Suda}, {and}
  \bibinfo{person}{Mani Srivastava}.} \bibinfo{year}{2019}\natexlab{}.
\newblock \showarticletitle{SenseHAR: A Robust Virtual Activity Sensor for
  Smartphones and Wearables}. In \bibinfo{booktitle}{\emph{17th Conference on
  Embedded Networked Sensor Systems}} \emph{(\bibinfo{series}{SenSys ’19})}.
  \bibinfo{pages}{15–28}.
\newblock


\bibitem[\protect\citeauthoryear{Kanamori, Hido, and Sugiyama}{Kanamori
  et~al\mbox{.}}{2009}]%
        {kanamori2009least}
\bibfield{author}{\bibinfo{person}{T. Kanamori}, \bibinfo{person}{S. Hido},
  {and} \bibinfo{person}{M. Sugiyama}.} \bibinfo{year}{2009}\natexlab{}.
\newblock \showarticletitle{A least-squares approach to direct importance
  estimation}.
\newblock \bibinfo{journal}{\emph{Journal of Machine Learning Research}}
  \bibinfo{volume}{10}, \bibinfo{number}{Jul} (\bibinfo{year}{2009}),
  \bibinfo{pages}{1391--1445}.
\newblock


\bibitem[\protect\citeauthoryear{Kawahara, Yairi, and Machida}{Kawahara
  et~al\mbox{.}}{2007}]%
        {Kawahara07}
\bibfield{author}{\bibinfo{person}{Y. Kawahara}, \bibinfo{person}{T. Yairi},
  {and} \bibinfo{person}{K. Machida}.} \bibinfo{year}{2007}\natexlab{}.
\newblock \showarticletitle{Change-Point Detection using Krylov Subspace
  Learning}. In \bibinfo{booktitle}{\emph{Proceedings of the Seventh {IEEE}
  International Conference on Data Mining, 2007}}. \bibinfo{pages}{559--564}.
\newblock


\bibitem[\protect\citeauthoryear{Keogh, Chu, Hart, and Pazzani}{Keogh
  et~al\mbox{.}}{2004}]%
        {keogh2004segmenting}
\bibfield{author}{\bibinfo{person}{Eamonn Keogh}, \bibinfo{person}{Selina Chu},
  \bibinfo{person}{David Hart}, {and} \bibinfo{person}{Michael Pazzani}.}
  \bibinfo{year}{2004}\natexlab{}.
\newblock \showarticletitle{Segmenting time series: A survey and novel
  approach}.
\newblock In \bibinfo{booktitle}{\emph{Data mining in time series databases}}.
  \bibinfo{publisher}{World Scientific}, \bibinfo{pages}{1--21}.
\newblock


\bibitem[\protect\citeauthoryear{Keogh and Lin}{Keogh and Lin}{2005}]%
        {keogh2005clustering}
\bibfield{author}{\bibinfo{person}{Eamonn Keogh} {and} \bibinfo{person}{Jessica
  Lin}.} \bibinfo{year}{2005}\natexlab{}.
\newblock \showarticletitle{Clustering of time-series subsequences is
  meaningless: implications for previous and future research}.
\newblock \bibinfo{journal}{\emph{Knowledge and information systems}}
  \bibinfo{volume}{8}, \bibinfo{number}{2} (\bibinfo{year}{2005}),
  \bibinfo{pages}{154--177}.
\newblock


\bibitem[\protect\citeauthoryear{Kingma and Ba}{Kingma and Ba}{2014}]%
        {kingma2014adam}
\bibfield{author}{\bibinfo{person}{Diederik~P Kingma} {and}
  \bibinfo{person}{Jimmy Ba}.} \bibinfo{year}{2014}\natexlab{}.
\newblock \showarticletitle{Adam: A method for stochastic optimization}.
\newblock \bibinfo{journal}{\emph{arXiv preprint arXiv:1412.6980}}
  (\bibinfo{year}{2014}).
\newblock


\bibitem[\protect\citeauthoryear{Kingma and Welling}{Kingma and
  Welling}{2013}]%
        {kingma2013auto}
\bibfield{author}{\bibinfo{person}{Diederik~P Kingma} {and}
  \bibinfo{person}{Max Welling}.} \bibinfo{year}{2013}\natexlab{}.
\newblock \showarticletitle{Auto-encoding variational bayes}.
\newblock \bibinfo{journal}{\emph{arXiv preprint arXiv:1312.6114}}
  (\bibinfo{year}{2013}).
\newblock


\bibitem[\protect\citeauthoryear{Kwon, Tong, Haresamudram, Gao, Abowd, Lane,
  and Pl\"{o}tz}{Kwon et~al\mbox{.}}{2020}]%
        {imutube}
\bibfield{author}{\bibinfo{person}{Hyeokhyen Kwon}, \bibinfo{person}{Catherine
  Tong}, \bibinfo{person}{Harish Haresamudram}, \bibinfo{person}{Yan Gao},
  \bibinfo{person}{Gregory~D. Abowd}, \bibinfo{person}{Nicholas~D. Lane}, {and}
  \bibinfo{person}{Thomas Pl\"{o}tz}.} \bibinfo{year}{2020}\natexlab{}.
\newblock \showarticletitle{IMUTube: Automatic Extraction of Virtual on-Body
  Accelerometry from Video for Human Activity Recognition}.
\newblock \bibinfo{journal}{\emph{Proc. ACM Interact. Mob. Wearable Ubiquitous
  Technol.}} \bibinfo{volume}{4}, \bibinfo{number}{3}, Article
  \bibinfo{articleno}{87} (\bibinfo{date}{Sept.} \bibinfo{year}{2020}),
  \bibinfo{numpages}{29}~pages.
\newblock
\urldef\tempurl%
\url{https://doi.org/10.1145/3411841}
\showDOI{\tempurl}


\bibitem[\protect\citeauthoryear{Lai, Chang, Yang, and Liu}{Lai
  et~al\mbox{.}}{2017}]%
        {LSTNet2017}
\bibfield{author}{\bibinfo{person}{Guokun Lai}, \bibinfo{person}{Wei{-}Cheng
  Chang}, \bibinfo{person}{Yiming Yang}, {and} \bibinfo{person}{Hanxiao Liu}.}
  \bibinfo{year}{2017}\natexlab{}.
\newblock \showarticletitle{Modeling Long- and Short-Term Temporal Patterns
  with Deep Neural Networks}.
\newblock \bibinfo{journal}{\emph{CoRR}}  \bibinfo{volume}{abs/1703.07015}
  (\bibinfo{year}{2017}).
\newblock
\urldef\tempurl%
\url{http://arxiv.org/abs/1703.07015}
\showURL{%
\tempurl}


\bibitem[\protect\citeauthoryear{Le}{Le}{2013}]%
        {le2013building}
\bibfield{author}{\bibinfo{person}{Quoc~V Le}.}
  \bibinfo{year}{2013}\natexlab{}.
\newblock \showarticletitle{Building high-level features using large scale
  unsupervised learning}. In \bibinfo{booktitle}{\emph{2013 IEEE international
  conference on acoustics, speech and signal processing}}. IEEE,
  \bibinfo{pages}{8595--8598}.
\newblock


\bibitem[\protect\citeauthoryear{LeCun, Bengio, and Hinton}{LeCun
  et~al\mbox{.}}{2015}]%
        {lecun2015deep}
\bibfield{author}{\bibinfo{person}{Yann LeCun}, \bibinfo{person}{Yoshua
  Bengio}, {and} \bibinfo{person}{Geoffrey Hinton}.}
  \bibinfo{year}{2015}\natexlab{}.
\newblock \showarticletitle{Deep learning}.
\newblock \bibinfo{journal}{\emph{nature}} \bibinfo{volume}{521},
  \bibinfo{number}{7553} (\bibinfo{year}{2015}), \bibinfo{pages}{436--444}.
\newblock


\bibitem[\protect\citeauthoryear{Lee, Ortiz, Ko, and Lee}{Lee
  et~al\mbox{.}}{2018}]%
        {lee2018time}
\bibfield{author}{\bibinfo{person}{Wei-Han Lee}, \bibinfo{person}{Jorge Ortiz},
  \bibinfo{person}{Bongjun Ko}, {and} \bibinfo{person}{Ruby Lee}.}
  \bibinfo{year}{2018}\natexlab{}.
\newblock \showarticletitle{Time series segmentation through automatic feature
  learning}.
\newblock \bibinfo{journal}{\emph{arXiv preprint arXiv:1801.05394}}
  (\bibinfo{year}{2018}).
\newblock


\bibitem[\protect\citeauthoryear{Li, Xie, Dai, and Song}{Li
  et~al\mbox{.}}{2015a}]%
        {NIPS2015_5684}
\bibfield{author}{\bibinfo{person}{S. Li}, \bibinfo{person}{Y. Xie},
  \bibinfo{person}{H. Dai}, {and} \bibinfo{person}{L. Song}.}
  \bibinfo{year}{2015}\natexlab{a}.
\newblock \showarticletitle{M-Statistic for Kernel Change-Point Detection}.
\newblock In \bibinfo{booktitle}{\emph{Advances in Neural Information
  Processing Systems 28}}. \bibinfo{pages}{3366--3374}.
\newblock


\bibitem[\protect\citeauthoryear{Li, Xie, Dai, and Song}{Li
  et~al\mbox{.}}{2015b}]%
        {XieYchange}
\bibfield{author}{\bibinfo{person}{Shuang Li}, \bibinfo{person}{Yao Xie},
  \bibinfo{person}{Hanjun Dai}, {and} \bibinfo{person}{Le Song}.}
  \bibinfo{year}{2015}\natexlab{b}.
\newblock \showarticletitle{M-Statistic for Kernel Change-Point Detection}. In
  \bibinfo{booktitle}{\emph{28th International Conference on Neural Information
  Processing Systems}} \emph{(\bibinfo{series}{NIPS’15},
  Vol.~\bibinfo{volume}{2})}. \bibinfo{pages}{3366–3374}.
\newblock


\bibitem[\protect\citeauthoryear{Li}{Li}{[n.d.]}]%
        {dailyevent}
\bibfield{author}{\bibinfo{person}{Wei Z. Jia W. Sun~M. Li, Z.}}
  \bibinfo{year}{[n.d.]}\natexlab{}.
\newblock \showarticletitle{Daily life event segmentation for lifestyle
  evaluation based on multi-sensor data recorded by a wearable device}.
\newblock  (\bibinfo{year}{[n.\,d.]}).
\newblock


\bibitem[\protect\citeauthoryear{{Lin}, {Clark}, {Birke}, {Schönborn},
  {Trigoni}, and {Roberts}}{{Lin} et~al\mbox{.}}{2020}]%
        {9053558}
\bibfield{author}{\bibinfo{person}{S. {Lin}}, \bibinfo{person}{R. {Clark}},
  \bibinfo{person}{R. {Birke}}, \bibinfo{person}{S. {Schönborn}},
  \bibinfo{person}{N. {Trigoni}}, {and} \bibinfo{person}{S. {Roberts}}.}
  \bibinfo{year}{2020}\natexlab{}.
\newblock \showarticletitle{Anomaly Detection for Time Series Using VAE-LSTM
  Hybrid Model}. In \bibinfo{booktitle}{\emph{ICASSP 2020 - 2020 IEEE
  International Conference on Acoustics, Speech and Signal Processing
  (ICASSP)}}. \bibinfo{pages}{4322--4326}.
\newblock
\urldef\tempurl%
\url{https://doi.org/10.1109/ICASSP40776.2020.9053558}
\showDOI{\tempurl}


\bibitem[\protect\citeauthoryear{Liu, Yamada, Collier, and Sugiyama}{Liu
  et~al\mbox{.}}{2013}]%
        {liu2013change}
\bibfield{author}{\bibinfo{person}{Song Liu}, \bibinfo{person}{Makoto Yamada},
  \bibinfo{person}{Nigel Collier}, {and} \bibinfo{person}{Masashi Sugiyama}.}
  \bibinfo{year}{2013}\natexlab{}.
\newblock \showarticletitle{Change-point detection in time-series data by
  relative density-ratio estimation}.
\newblock \bibinfo{journal}{\emph{Neural Networks}}  \bibinfo{volume}{43}
  (\bibinfo{year}{2013}), \bibinfo{pages}{72--83}.
\newblock


\bibitem[\protect\citeauthoryear{Liu, Alibhai, Wang, Liu, He, and Wu}{Liu
  et~al\mbox{.}}{2019}]%
        {hpc}
\bibfield{author}{\bibinfo{person}{Tong Liu}, \bibinfo{person}{Shakeel
  Alibhai}, \bibinfo{person}{Jinzhen Wang}, \bibinfo{person}{Qing Liu},
  \bibinfo{person}{Xubin He}, {and} \bibinfo{person}{Chentao Wu}.}
  \bibinfo{year}{2019}\natexlab{}.
\newblock \showarticletitle{Exploring Transfer Learning to Reduce Training
  Overhead of HPC Data in Machine Learning}. In \bibinfo{booktitle}{\emph{2019
  IEEE International Conference on Networking, Architecture and Storage
  (NAS)}}. \bibinfo{pages}{1--7}.
\newblock
\urldef\tempurl%
\url{https://doi.org/10.1109/NAS.2019.8834723}
\showDOI{\tempurl}


\bibitem[\protect\citeauthoryear{Liu, Qin, Shao, Chen, and Bian}{Liu
  et~al\mbox{.}}{[n.d.]}]%
        {ML01}
\bibfield{author}{\bibinfo{person}{T.~Y. Liu}, \bibinfo{person}{T. Qin},
  \bibinfo{person}{B. Shao}, \bibinfo{person}{W. Chen}, {and}
  \bibinfo{person}{J. Bian}.} \bibinfo{year}{[n.d.]}\natexlab{}.
\newblock \bibinfo{title}{{Machine Learning: Research hotspots in the next ten
  years}}.
\newblock
  \bibinfo{howpublished}{\url{https://www.microsoft.com/en-us/research/lab/microsoft-research-asia/articles/machine-learning-research-hotspots/
  }}.
\newblock


\bibitem[\protect\citeauthoryear{Liu, Li, and P{\'o}czos}{Liu
  et~al\mbox{.}}{2018}]%
        {Liu2018ClassifierTS}
\bibfield{author}{\bibinfo{person}{Yusha Liu}, \bibinfo{person}{Chun-Liang Li},
  {and} \bibinfo{person}{Barnab{\'a}s P{\'o}czos}.}
  \bibinfo{year}{2018}\natexlab{}.
\newblock \showarticletitle{Classifier Two Sample Test for Video Anomaly
  Detections}. In \bibinfo{booktitle}{\emph{BMVC}}.
\newblock


\bibitem[\protect\citeauthoryear{Livni, Shalev-Shwartz, and Shamir}{Livni
  et~al\mbox{.}}{2014}]%
        {10.5555/2968826.2968922}
\bibfield{author}{\bibinfo{person}{Roi Livni}, \bibinfo{person}{Shai
  Shalev-Shwartz}, {and} \bibinfo{person}{Ohad Shamir}.}
  \bibinfo{year}{2014}\natexlab{}.
\newblock \showarticletitle{On the Computational Efficiency of Training Neural
  Networks}. In \bibinfo{booktitle}{\emph{Proceedings of the 27th International
  Conference on Neural Information Processing Systems - Volume 1}} (Montreal,
  Canada) \emph{(\bibinfo{series}{NIPS'14})}. \bibinfo{publisher}{MIT Press},
  \bibinfo{address}{Cambridge, MA, USA}, \bibinfo{pages}{855–863}.
\newblock


\bibitem[\protect\citeauthoryear{Lowry, Woodall, Champ, and Rigdon}{Lowry
  et~al\mbox{.}}{1992}]%
        {Lowry1992AME}
\bibfield{author}{\bibinfo{person}{Cynthia~A. Lowry},
  \bibinfo{person}{William~H. Woodall}, \bibinfo{person}{Charles~W. Champ},
  {and} \bibinfo{person}{Steven~E. Rigdon}.} \bibinfo{year}{1992}\natexlab{}.
\newblock \showarticletitle{A multivariate exponentially weighted moving
  average control chart}.
\newblock


\bibitem[\protect\citeauthoryear{Ma, Zhang, Li, and Lu}{Ma
  et~al\mbox{.}}{2021}]%
        {10.1145/3448074}
\bibfield{author}{\bibinfo{person}{Haojie Ma}, \bibinfo{person}{Zhijie Zhang},
  \bibinfo{person}{Wenzhong Li}, {and} \bibinfo{person}{Sanglu Lu}.}
  \bibinfo{year}{2021}\natexlab{}.
\newblock \showarticletitle{Unsupervised Human Activity Representation Learning
  with Multi-Task Deep Clustering}.
\newblock \bibinfo{journal}{\emph{Proc. ACM Interact. Mob. Wearable Ubiquitous
  Technol.}} \bibinfo{volume}{5}, \bibinfo{number}{1}, Article
  \bibinfo{articleno}{48} (\bibinfo{date}{March} \bibinfo{year}{2021}),
  \bibinfo{numpages}{25}~pages.
\newblock
\urldef\tempurl%
\url{https://doi.org/10.1145/3448074}
\showDOI{\tempurl}


\bibitem[\protect\citeauthoryear{Maaten and Hinton}{Maaten and Hinton}{2008}]%
        {maaten2008visualizing}
\bibfield{author}{\bibinfo{person}{Laurens van~der Maaten} {and}
  \bibinfo{person}{Geoffrey Hinton}.} \bibinfo{year}{2008}\natexlab{}.
\newblock \showarticletitle{Visualizing data using t-SNE}.
\newblock \bibinfo{journal}{\emph{Journal of machine learning research}}
  \bibinfo{volume}{9}, \bibinfo{number}{Nov} (\bibinfo{year}{2008}),
  \bibinfo{pages}{2579--2605}.
\newblock


\bibitem[\protect\citeauthoryear{Moskvina and Zhigljavsky}{Moskvina and
  Zhigljavsky}{2003}]%
        {Moskvina03}
\bibfield{author}{\bibinfo{person}{V. Moskvina} {and} \bibinfo{person}{A.
  Zhigljavsky}.} \bibinfo{year}{2003}\natexlab{}.
\newblock \showarticletitle{An Algorithm Based on Singular Spectrum Analysis
  for Change-Point Detection}.
\newblock \bibinfo{journal}{\emph{Communications in Statistics-simulation and
  Computation}}  \bibinfo{volume}{32} (\bibinfo{date}{01}
  \bibinfo{year}{2003}), \bibinfo{pages}{319--352}.
\newblock


\bibitem[\protect\citeauthoryear{Nassar, Wilson, Heasly, and Gold}{Nassar
  et~al\mbox{.}}{2010}]%
        {Nassar2010AnAB}
\bibfield{author}{\bibinfo{person}{Matthew~R. Nassar},
  \bibinfo{person}{Robert~C. Wilson}, \bibinfo{person}{Benjamin~S. Heasly},
  {and} \bibinfo{person}{Joshua~I. Gold}.} \bibinfo{year}{2010}\natexlab{}.
\newblock \showarticletitle{An approximately Bayesian delta-rule model explains
  the dynamics of belief updating in a changing environment.}
\newblock \bibinfo{journal}{\emph{The Journal of neuroscience : the official
  journal of the Society for Neuroscience}} \bibinfo{volume}{30},
  \bibinfo{number}{37} (\bibinfo{year}{2010}), \bibinfo{pages}{12366--78}.
\newblock


\bibitem[\protect\citeauthoryear{Nguyen, Rupavatharam, Liu, Howard, and
  Gruteser}{Nguyen et~al\mbox{.}}{2019}]%
        {handsense2019}
\bibfield{author}{\bibinfo{person}{Viet Nguyen}, \bibinfo{person}{Siddharth
  Rupavatharam}, \bibinfo{person}{Luyang Liu}, \bibinfo{person}{Richard
  Howard}, {and} \bibinfo{person}{Marco Gruteser}.}
  \bibinfo{year}{2019}\natexlab{}.
\newblock \showarticletitle{HandSense: Capacitive Coupling-Based Dynamic, Micro
  Finger Gesture Recognition}. In \bibinfo{booktitle}{\emph{Proceedings of the
  17th Conference on Embedded Networked Sensor Systems}}
  \emph{(\bibinfo{series}{SenSys ’19})}. \bibinfo{pages}{285–297}.
\newblock


\bibitem[\protect\citeauthoryear{Ni, Zhang, and Li}{Ni et~al\mbox{.}}{2018}]%
        {ni2018}
\bibfield{author}{\bibinfo{person}{Qin Ni}, \bibinfo{person}{Lei Zhang}, {and}
  \bibinfo{person}{Luqun Li}.} \bibinfo{year}{2018}\natexlab{}.
\newblock \showarticletitle{A Heterogeneous Ensemble Approach for Activity
  Recognition with Integration of Change Point-Based Data Segmentation}.
\newblock \bibinfo{journal}{\emph{Applied Sciences}} \bibinfo{volume}{8},
  \bibinfo{number}{9} (\bibinfo{year}{2018}), \bibinfo{pages}{1695}.
\newblock


\bibitem[\protect\citeauthoryear{Nishiyama and Fukumizu}{Nishiyama and
  Fukumizu}{2016}]%
        {nishiyama2016}
\bibfield{author}{\bibinfo{person}{Yu Nishiyama} {and} \bibinfo{person}{Kenji
  Fukumizu}.} \bibinfo{year}{2016}\natexlab{}.
\newblock \showarticletitle{Characteristic kernels and infinitely divisible
  distributions}.
\newblock \bibinfo{journal}{\emph{The Journal of Machine Learning Research}}
  \bibinfo{volume}{17}, \bibinfo{number}{1} (\bibinfo{year}{2016}),
  \bibinfo{pages}{6240--6267}.
\newblock


\bibitem[\protect\citeauthoryear{Noor, Salcic, Kevin, and Wang}{Noor
  et~al\mbox{.}}{2017}]%
        {noor2017adaptive}
\bibfield{author}{\bibinfo{person}{Mohd Halim~Mohd Noor},
  \bibinfo{person}{Zoran Salcic}, \bibinfo{person}{I Kevin}, {and}
  \bibinfo{person}{Kai Wang}.} \bibinfo{year}{2017}\natexlab{}.
\newblock \showarticletitle{Adaptive sliding window segmentation for physical
  activity recognition using a single tri-axial accelerometer}.
\newblock \bibinfo{journal}{\emph{Pervasive and Mobile Computing}}
  \bibinfo{volume}{38} (\bibinfo{year}{2017}), \bibinfo{pages}{41--59}.
\newblock


\bibitem[\protect\citeauthoryear{Ohara, Maekawa, and Matsushita}{Ohara
  et~al\mbox{.}}{2017}]%
        {10.1145/3131898}
\bibfield{author}{\bibinfo{person}{Kazuya Ohara}, \bibinfo{person}{Takuya
  Maekawa}, {and} \bibinfo{person}{Yasuyuki Matsushita}.}
  \bibinfo{year}{2017}\natexlab{}.
\newblock \showarticletitle{Detecting State Changes of Indoor Everyday Objects
  Using Wi-Fi Channel State Information}.
\newblock \bibinfo{journal}{\emph{Proc. ACM Interact. Mob. Wearable Ubiquitous
  Technol.}} \bibinfo{volume}{1}, \bibinfo{number}{3}, Article
  \bibinfo{articleno}{88} (\bibinfo{date}{Sept.} \bibinfo{year}{2017}),
  \bibinfo{numpages}{28}~pages.
\newblock
\urldef\tempurl%
\url{https://doi.org/10.1145/3131898}
\showDOI{\tempurl}


\bibitem[\protect\citeauthoryear{Osborne, Garnett, Swersky, and
  De~Freitas}{Osborne et~al\mbox{.}}{2011}]%
        {osborne2011machine}
\bibfield{author}{\bibinfo{person}{Michael Osborne}, \bibinfo{person}{Roman
  Garnett}, \bibinfo{person}{Kevin Swersky}, {and} \bibinfo{person}{Nando
  De~Freitas}.} \bibinfo{year}{2011}\natexlab{}.
\newblock \showarticletitle{A machine learning approach to pattern detection
  and prediction for environmental monitoring and water sustainability}. In
  \bibinfo{booktitle}{\emph{ICML Workshop on Machine Learning for Global
  Challenges}}.
\newblock


\bibitem[\protect\citeauthoryear{Pan, Berges, Rodakowski, Zhang, and Noh}{Pan
  et~al\mbox{.}}{2019}]%
        {pan2019fine}
\bibfield{author}{\bibinfo{person}{Shijia Pan}, \bibinfo{person}{Mario Berges},
  \bibinfo{person}{Juleen Rodakowski}, \bibinfo{person}{Pei Zhang}, {and}
  \bibinfo{person}{Hae~Young Noh}.} \bibinfo{year}{2019}\natexlab{}.
\newblock \showarticletitle{Fine-grained recognition of activities of daily
  living through structural vibration and electrical sensing}. In
  \bibinfo{booktitle}{\emph{6th ACM International Conference on Systems for
  Energy-Efficient Buildings, Cities, and Transportation}}.
  \bibinfo{pages}{149--158}.
\newblock


\bibitem[\protect\citeauthoryear{P{\'e}rez-Cruz}{P{\'e}rez-Cruz}{2008}]%
        {perez2008kullback}
\bibfield{author}{\bibinfo{person}{Fernando P{\'e}rez-Cruz}.}
  \bibinfo{year}{2008}\natexlab{}.
\newblock \showarticletitle{Kullback-Leibler divergence estimation of
  continuous distributions}. In \bibinfo{booktitle}{\emph{2008 IEEE
  international symposium on information theory}}. IEEE,
  \bibinfo{pages}{1666--1670}.
\newblock


\bibitem[\protect\citeauthoryear{Ramdas, Reddi, P{\'o}czos, Singh, and
  Wasserman}{Ramdas et~al\mbox{.}}{2015}]%
        {ramdas2015decreasing}
\bibfield{author}{\bibinfo{person}{A. Ramdas}, \bibinfo{person}{S.~J. Reddi},
  \bibinfo{person}{B. P{\'o}czos}, \bibinfo{person}{A. Singh}, {and}
  \bibinfo{person}{L. Wasserman}.} \bibinfo{year}{2015}\natexlab{}.
\newblock \showarticletitle{On the decreasing power of kernel and distance
  based nonparametric hypothesis tests in high dimensions}. In
  \bibinfo{booktitle}{\emph{Twenty-Ninth AAAI Conference on Artificial
  Intelligence}}.
\newblock


\bibitem[\protect\citeauthoryear{Ryck, Vos, and Bertrand}{Ryck
  et~al\mbox{.}}{2021}]%
        {deryck2021change}
\bibfield{author}{\bibinfo{person}{Tim~De Ryck}, \bibinfo{person}{Maarten~De
  Vos}, {and} \bibinfo{person}{Alexander Bertrand}.}
  \bibinfo{year}{2021}\natexlab{}.
\newblock \bibinfo{title}{Change Point Detection in Time Series Data using
  Autoencoders with a Time-Invariant Representation}.
\newblock
\newblock
\showeprint[arxiv]{2008.09524}~[cs.LG]


\bibitem[\protect\citeauthoryear{Saat\c{c}i, Turner, and Rasmussen}{Saat\c{c}i
  et~al\mbox{.}}{2010}]%
        {Yunus2010}
\bibfield{author}{\bibinfo{person}{Yunus Saat\c{c}i}, \bibinfo{person}{Ryan
  Turner}, {and} \bibinfo{person}{Carl~Edward Rasmussen}.}
  \bibinfo{year}{2010}\natexlab{}.
\newblock \showarticletitle{Gaussian Process Change Point Models}. In
  \bibinfo{booktitle}{\emph{Proceedings of the 27th International Conference on
  International Conference on Machine Learning}}
  \emph{(\bibinfo{series}{ICML’10})}. \bibinfo{pages}{927–934}.
\newblock


\bibitem[\protect\citeauthoryear{Saat{\c{c}}i}{Saat{\c{c}}i}{2010}]%
        {saatcci2010gaussian}
\bibfield{author}{\bibinfo{person}{Yunus Saat{\c{c}}i}.}
  \bibinfo{year}{2010}\natexlab{}.
\newblock \showarticletitle{Gaussian Process Change Point Models.}
\newblock


\bibitem[\protect\citeauthoryear{Sadri, Salim, Ren, Shao, Krumm, and
  Mascolo}{Sadri et~al\mbox{.}}{2018}]%
        {10.1145/3287064}
\bibfield{author}{\bibinfo{person}{Amin Sadri}, \bibinfo{person}{Flora~D.
  Salim}, \bibinfo{person}{Yongli Ren}, \bibinfo{person}{Wei Shao},
  \bibinfo{person}{John~C. Krumm}, {and} \bibinfo{person}{Cecilia Mascolo}.}
  \bibinfo{year}{2018}\natexlab{}.
\newblock \showarticletitle{What Will You Do for the Rest of the Day? An
  Approach to Continuous Trajectory Prediction}.
\newblock \bibinfo{journal}{\emph{Proc. ACM Interact. Mob. Wearable Ubiquitous
  Technol.}} \bibinfo{volume}{2}, \bibinfo{number}{4}, Article
  \bibinfo{articleno}{186} (\bibinfo{date}{Dec.} \bibinfo{year}{2018}),
  \bibinfo{numpages}{26}~pages.
\newblock
\urldef\tempurl%
\url{https://doi.org/10.1145/3287064}
\showDOI{\tempurl}


\bibitem[\protect\citeauthoryear{Saeed, Salim, Ozcelebi, and Lukkien}{Saeed
  et~al\mbox{.}}{2021}]%
        {2021}
\bibfield{author}{\bibinfo{person}{Aaqib Saeed}, \bibinfo{person}{Flora~D.
  Salim}, \bibinfo{person}{Tanir Ozcelebi}, {and} \bibinfo{person}{Johan
  Lukkien}.} \bibinfo{year}{2021}\natexlab{}.
\newblock \showarticletitle{Federated Self-Supervised Learning of Multisensor
  Representations for Embedded Intelligence}.
\newblock \bibinfo{journal}{\emph{IEEE Internet of Things Journal}}
  \bibinfo{volume}{8}, \bibinfo{number}{2} (\bibinfo{date}{Jan}
  \bibinfo{year}{2021}), \bibinfo{pages}{1030–1040}.
\newblock
\showISSN{2372-2541}
\urldef\tempurl%
\url{https://doi.org/10.1109/jiot.2020.3009358}
\showDOI{\tempurl}


\bibitem[\protect\citeauthoryear{Sarker, Colman, Kabir, and Han}{Sarker
  et~al\mbox{.}}{2018}]%
        {sarker2018individualized}
\bibfield{author}{\bibinfo{person}{Iqbal~H Sarker}, \bibinfo{person}{Alan
  Colman}, \bibinfo{person}{Muhammad~Ashad Kabir}, {and} \bibinfo{person}{Jun
  Han}.} \bibinfo{year}{2018}\natexlab{}.
\newblock \showarticletitle{Individualized time-series segmentation for mining
  mobile phone user behavior}.
\newblock \bibinfo{journal}{\emph{Comput. J.}} \bibinfo{volume}{61},
  \bibinfo{number}{3} (\bibinfo{year}{2018}), \bibinfo{pages}{349--368}.
\newblock


\bibitem[\protect\citeauthoryear{Sch{\"o}lkopf, Smola, Bach,
  et~al\mbox{.}}{Sch{\"o}lkopf et~al\mbox{.}}{2002}]%
        {scholkopf2002learning}
\bibfield{author}{\bibinfo{person}{Bernhard Sch{\"o}lkopf},
  \bibinfo{person}{Alexander~J Smola}, \bibinfo{person}{Francis Bach},
  {et~al\mbox{.}}} \bibinfo{year}{2002}\natexlab{}.
\newblock \bibinfo{booktitle}{\emph{Learning with kernels: support vector
  machines, regularization, optimization, and beyond}}.
\newblock


\bibitem[\protect\citeauthoryear{Sinn, Ghodsi, and Keller}{Sinn
  et~al\mbox{.}}{2012}]%
        {sinn2012detecting}
\bibfield{author}{\bibinfo{person}{Mathieu Sinn}, \bibinfo{person}{Ali Ghodsi},
  {and} \bibinfo{person}{Karsten Keller}.} \bibinfo{year}{2012}\natexlab{}.
\newblock \showarticletitle{Detecting change-points in time series by maximum
  mean discrepancy of ordinal pattern distributions}. In
  \bibinfo{booktitle}{\emph{Proceedings of the Twenty-Eighth Conference on
  Uncertainty in Artificial Intelligence}}. \bibinfo{pages}{786--794}.
\newblock


\bibitem[\protect\citeauthoryear{Stankovic}{Stankovic}{2014}]%
        {IoT000}
\bibfield{author}{\bibinfo{person}{J. Stankovic}.}
  \bibinfo{year}{2014}\natexlab{}.
\newblock \showarticletitle{Research Directions for the {Internet of Things}}.
\newblock \bibinfo{journal}{\emph{IEEE Internet of Things Journal}}
  \bibinfo{volume}{1}, \bibinfo{number}{1} (\bibinfo{date}{Feb.}
  \bibinfo{year}{2014}), \bibinfo{pages}{3 -- 9}.
\newblock


\bibitem[\protect\citeauthoryear{Sugiyama, Nakajima, Kashima, Buenau, and
  Kawanabe}{Sugiyama et~al\mbox{.}}{2008}]%
        {sugiyama2008direct}
\bibfield{author}{\bibinfo{person}{M. Sugiyama}, \bibinfo{person}{S. Nakajima},
  \bibinfo{person}{H. Kashima}, \bibinfo{person}{P.~V. Buenau}, {and}
  \bibinfo{person}{M. Kawanabe}.} \bibinfo{year}{2008}\natexlab{}.
\newblock \showarticletitle{Direct importance estimation with model selection
  and its application to covariate shift adaptation}. In
  \bibinfo{booktitle}{\emph{Advances in neural information processing
  systems}}. \bibinfo{pages}{1433--1440}.
\newblock


\bibitem[\protect\citeauthoryear{Sugiyama, Suzuki, and Kanamori}{Sugiyama
  et~al\mbox{.}}{2012}]%
        {sugiyama2012density}
\bibfield{author}{\bibinfo{person}{M. Sugiyama}, \bibinfo{person}{T. Suzuki},
  {and} \bibinfo{person}{T. Kanamori}.} \bibinfo{year}{2012}\natexlab{}.
\newblock \bibinfo{booktitle}{\emph{Density ratio estimation in machine
  learning}}.
\newblock \bibinfo{publisher}{Cambridge University Press}.
\newblock


\bibitem[\protect\citeauthoryear{{Takayasu}}{{Takayasu}}{2015}]%
        {7288606}
\bibfield{author}{\bibinfo{person}{H. {Takayasu}}.}
  \bibinfo{year}{2015}\natexlab{}.
\newblock \showarticletitle{Basic methods of change-point detection of
  financial fluctuations}. In \bibinfo{booktitle}{\emph{2015 International
  Conference on Noise and Fluctuations (ICNF)}}. \bibinfo{pages}{1--3}.
\newblock


\bibitem[\protect\citeauthoryear{Tang, Perez-Pozuelo, Spathis, Brage, Wareham,
  and Mascolo}{Tang et~al\mbox{.}}{2021}]%
        {10.1145/3448112}
\bibfield{author}{\bibinfo{person}{Chi~Ian Tang}, \bibinfo{person}{Ignacio
  Perez-Pozuelo}, \bibinfo{person}{Dimitris Spathis}, \bibinfo{person}{Soren
  Brage}, \bibinfo{person}{Nick Wareham}, {and} \bibinfo{person}{Cecilia
  Mascolo}.} \bibinfo{year}{2021}\natexlab{}.
\newblock \showarticletitle{SelfHAR: Improving Human Activity Recognition
  through Self-Training with Unlabeled Data}.
\newblock \bibinfo{journal}{\emph{Proc. ACM Interact. Mob. Wearable Ubiquitous
  Technol.}} \bibinfo{volume}{5}, \bibinfo{number}{1}, Article
  \bibinfo{articleno}{36} (\bibinfo{date}{March} \bibinfo{year}{2021}),
  \bibinfo{numpages}{30}~pages.
\newblock
\urldef\tempurl%
\url{https://doi.org/10.1145/3448112}
\showDOI{\tempurl}


\bibitem[\protect\citeauthoryear{Vincent, Larochelle, Lajoie, Bengio, and
  Manzagol}{Vincent et~al\mbox{.}}{2010}]%
        {vincent2010stacked}
\bibfield{author}{\bibinfo{person}{P. Vincent}, \bibinfo{person}{H.
  Larochelle}, \bibinfo{person}{I. Lajoie}, \bibinfo{person}{Y. Bengio}, {and}
  \bibinfo{person}{P.~A. Manzagol}.} \bibinfo{year}{2010}\natexlab{}.
\newblock \showarticletitle{Stacked denoising autoencoders: Learning useful
  representations in a deep network with a local denoising criterion}.
\newblock \bibinfo{journal}{\emph{Journal of machine learning research}}
  \bibinfo{volume}{11}, \bibinfo{number}{Dec} (\bibinfo{year}{2010}),
  \bibinfo{pages}{3371--3408}.
\newblock


\bibitem[\protect\citeauthoryear{Wang, Olugbade, Mathur, De~C.~Williams, Lane,
  and Bianchi-Berthouze}{Wang et~al\mbox{.}}{2019}]%
        {Wang2019}
\bibfield{author}{\bibinfo{person}{Chongyang Wang},
  \bibinfo{person}{Temitayo~A. Olugbade}, \bibinfo{person}{Akhil Mathur},
  \bibinfo{person}{Amanda~C. De~C.~Williams}, \bibinfo{person}{Nicholas~D.
  Lane}, {and} \bibinfo{person}{Nadia Bianchi-Berthouze}.}
  \bibinfo{year}{2019}\natexlab{}.
\newblock \showarticletitle{Recurrent Network Based Automatic Detection of
  Chronic Pain Protective Behavior Using MoCap and SEMG Data}. In
  \bibinfo{booktitle}{\emph{Proceedings of the 23rd International Symposium on
  Wearable Computers}} \emph{(\bibinfo{series}{ISWC ’19})}.
  \bibinfo{pages}{225–230}.
\newblock


\bibitem[\protect\citeauthoryear{Wang, Wu, Ji, Wang, and Liang}{Wang
  et~al\mbox{.}}{2011}]%
        {Wang11}
\bibfield{author}{\bibinfo{person}{Y. Wang}, \bibinfo{person}{C. Wu},
  \bibinfo{person}{Z. Ji}, \bibinfo{person}{B. Wang}, {and} \bibinfo{person}{Y.
  Liang}.} \bibinfo{year}{2011}\natexlab{}.
\newblock \showarticletitle{Non-Parametric Change-Point Method for Differential
  Gene Expression Detection}.
\newblock \bibinfo{journal}{\emph{PloS one}}  \bibinfo{volume}{6}
  (\bibinfo{date}{05} \bibinfo{year}{2011}), \bibinfo{pages}{e20060}.
\newblock


\bibitem[\protect\citeauthoryear{Wang and Zheng}{Wang and Zheng}{2018}]%
        {10.1145/3287071}
\bibfield{author}{\bibinfo{person}{Yanwen Wang} {and} \bibinfo{person}{Yuanqing
  Zheng}.} \bibinfo{year}{2018}\natexlab{}.
\newblock \showarticletitle{Modeling RFID Signal Reflection for Contact-Free
  Activity Recognition}.
\newblock \bibinfo{journal}{\emph{Proc. ACM Interact. Mob. Wearable Ubiquitous
  Technol.}} \bibinfo{volume}{2}, \bibinfo{number}{4}, Article
  \bibinfo{articleno}{193} (\bibinfo{date}{Dec.} \bibinfo{year}{2018}),
  \bibinfo{numpages}{22}~pages.
\newblock
\urldef\tempurl%
\url{https://doi.org/10.1145/3287071}
\showDOI{\tempurl}


\bibitem[\protect\citeauthoryear{Webb, Hyde, Cao, Nguyen, and Petitjean}{Webb
  et~al\mbox{.}}{2016}]%
        {1484-charconcept}
\bibfield{author}{\bibinfo{person}{G.~I. Webb}, \bibinfo{person}{R. Hyde},
  \bibinfo{person}{H. Cao}, \bibinfo{person}{H.~L. Nguyen}, {and}
  \bibinfo{person}{F. Petitjean}.} \bibinfo{year}{2016}\natexlab{}.
\newblock \showarticletitle{Characterizing Concept Drift}.
\newblock \bibinfo{journal}{\emph{Data Min. Knowl. Discov.}}
  \bibinfo{volume}{30}, \bibinfo{number}{4} (\bibinfo{year}{2016}),
  \bibinfo{pages}{964–994}.
\newblock


\bibitem[\protect\citeauthoryear{{Willsky} and {Jones}}{{Willsky} and
  {Jones}}{1976}]%
        {1101146}
\bibfield{author}{\bibinfo{person}{A. {Willsky}} {and} \bibinfo{person}{H.
  {Jones}}.} \bibinfo{year}{1976}\natexlab{}.
\newblock \showarticletitle{A generalized likelihood ratio approach to the
  detection and estimation of jumps in linear systems}.
\newblock \bibinfo{journal}{\emph{IEEE Trans. Automat. Control}}
  \bibinfo{volume}{21}, \bibinfo{number}{1} (\bibinfo{year}{1976}),
  \bibinfo{pages}{108--112}.
\newblock


\bibitem[\protect\citeauthoryear{Yamanishi, Takeuchi, Williams, and
  Milne}{Yamanishi et~al\mbox{.}}{2000}]%
        {Yamanishi00}
\bibfield{author}{\bibinfo{person}{K. Yamanishi}, \bibinfo{person}{J.
  Takeuchi}, \bibinfo{person}{G. Williams}, {and} \bibinfo{person}{P. Milne}.}
  \bibinfo{year}{2000}\natexlab{}.
\newblock \showarticletitle{On-Line Unsupervised Outlier Detection Using Finite
  Mixtures with Discounting Learning Algorithms}. In
  \bibinfo{booktitle}{\emph{6th International Conference on Knowledge Discovery
  and Data Mining}} \emph{(\bibinfo{series}{KDD ’00})}.
  \bibinfo{pages}{320–324}.
\newblock


\bibitem[\protect\citeauthoryear{Yamanishi and Takeuchi}{Yamanishi and
  Takeuchi}{2002}]%
        {Takuechi}
\bibfield{author}{\bibinfo{person}{Kenji Yamanishi} {and}
  \bibinfo{person}{Jun-ichi Takeuchi}.} \bibinfo{year}{2002}\natexlab{}.
\newblock \showarticletitle{A Unifying Framework for Detecting Outliers and
  Change Points from Non-Stationary Time Series Data}. In
  \bibinfo{booktitle}{\emph{Proceedings of the Eighth ACM SIGKDD International
  Conference on Knowledge Discovery and Data Mining}}
  \emph{(\bibinfo{series}{KDD ’02})}. \bibinfo{pages}{676–681}.
\newblock


\bibitem[\protect\citeauthoryear{Zhang, Hao, Yu, Chang, Lai, and Yang}{Zhang
  et~al\mbox{.}}{2020}]%
        {zhang2020explainable}
\bibfield{author}{\bibinfo{person}{Ruohong Zhang}, \bibinfo{person}{Yu Hao},
  \bibinfo{person}{Donghan Yu}, \bibinfo{person}{Wei-Cheng Chang},
  \bibinfo{person}{Guokun Lai}, {and} \bibinfo{person}{Yiming Yang}.}
  \bibinfo{year}{2020}\natexlab{}.
\newblock \bibinfo{title}{Explainable Unsupervised Change-point Detection via
  Graph Neural Networks}.
\newblock
\newblock
\showeprint[arxiv]{2004.11934}~[cs.LG]


\bibitem[\protect\citeauthoryear{Zhuang, Qi, Duan, Xi, Zhu, Zhu, Xiong, and
  He}{Zhuang et~al\mbox{.}}{2020}]%
        {zhuang2020comprehensive}
\bibfield{author}{\bibinfo{person}{Fuzhen Zhuang}, \bibinfo{person}{Zhiyuan
  Qi}, \bibinfo{person}{Keyu Duan}, \bibinfo{person}{Dongbo Xi},
  \bibinfo{person}{Yongchun Zhu}, \bibinfo{person}{Hengshu Zhu},
  \bibinfo{person}{Hui Xiong}, {and} \bibinfo{person}{Qing He}.}
  \bibinfo{year}{2020}\natexlab{}.
\newblock \bibinfo{title}{A Comprehensive Survey on Transfer Learning}.
\newblock
\newblock
\showeprint[arxiv]{1911.02685}~[cs.LG]


\bibitem[\protect\citeauthoryear{Zimmermann, Weigel, and Fischer}{Zimmermann
  et~al\mbox{.}}{2018}]%
        {Zimmermann}
\bibfield{author}{\bibinfo{person}{Lars Zimmermann}, \bibinfo{person}{Robert
  Weigel}, {and} \bibinfo{person}{Georg Fischer}.}
  \bibinfo{year}{2018}\natexlab{}.
\newblock \showarticletitle{Fusion of Nonintrusive Environmental Sensors for
  Occupancy Detection in Smart Homes}.
\newblock \bibinfo{journal}{\emph{IEEE Internet of Things Journal}}
  \bibinfo{volume}{5}, \bibinfo{number}{4} (\bibinfo{year}{2018}),
  \bibinfo{pages}{2343--2352}.
\newblock
\urldef\tempurl%
\url{https://doi.org/10.1109/JIOT.2017.2752134}
\showDOI{\tempurl}


\bibitem[\protect\citeauthoryear{Zou, Liang, Poor, and Shi}{Zou
  et~al\mbox{.}}{2014}]%
        {zou14}
\bibfield{author}{\bibinfo{person}{Shaofeng Zou}, \bibinfo{person}{Yingbin
  Liang}, \bibinfo{person}{H.~Vincent Poor}, {and}
  \bibinfo{person}{Xinghua~Mindy Shi}.} \bibinfo{year}{2014}\natexlab{}.
\newblock \showarticletitle{Nonparametric Detection of Anomalous Data via
  Kernel Mean Embedding}.
\newblock  (\bibinfo{date}{04} \bibinfo{year}{2014}).
\newblock


\end{thebibliography}

\end{document}